%% file: _thesis_.tex
\documentclass[a4paper,oneside,12pt,times,index,custommargin]{classes/PhDThesisPSnPDF}
\input{preamble/preamble}

\input{my_settings}
\input{math_commands.tex}

\begin{document}

\frontmatter
\input{Chapters/title.tex}
\maketitle


\begin{spacing}{1.1}
\input{Chapters/acknowledgments.tex}
\tableofcontents 
\input{Chapters/abstract.tex}
\listoftables
\listoffigures
\input{Chapters/listofsymbols.tex}
\end{spacing}

\mainmatter
\input{Chapters/introduction}

 \input{Chapters/ijcai2022}
 \input{Chapters/iclr2020}

 \input{Chapters/icml2021}

\input{Chapters/conclusion}

\begin{spacing}{1}
\bookmarksetup{startatroot}
\bibliography{References/intro.bib,References/iclr2020.bib,References/icml2021.bib,References/ijcai2022.bib}

\end{spacing}



%
%


\end{document}

%% file: my_settings.tex
\usepackage{CJKutf8}

\usepackage{bookmark}

\usepackage{natbib} 
\setcitestyle{authoryear,open={[},close={]}}
\usepackage{microtype} 

\usepackage{color}
\usepackage[usenames,dvipsnames]{xcolor}
\usepackage{soul}

\usepackage{enumitem}
\usepackage{graphicx,grffile,subcaption,wrapfig}
\usepackage{multirow,booktabs,longtable,tabulary}
\usepackage{hhline}

\newcommand{\proj}{\text{Proj}}
\newcommand{\vdelta}{\bm \delta}

\newcommand{\tit}[1]{\textit{#1}}
\newcommand{\tbf}[1]{\textbf{#1}}

\newcommand{\etal}{\emph{et al. }}

\newcommand\blfootnote[1]{
\begingroup
\renewcommand\thefootnote{}\footnote{#1}\addtocounter{footnote}{-1}
\endgroup}

\newcommand{\thl}[1]{\tbf{#1}} 
\newcommand{\tHL}[1]{\underline{#1}} 

%% file: math_commands.tex
\usepackage{amsmath,amsfonts, amssymb, mathtools, bm, bbm, nicefrac}
\newcommand{\mcal}[1]{\mathcal{#1}}

\usepackage{algorithm}
\usepackage[noend]{algpseudocode}

\newtheorem{theorem}{Theorem}[section]

\def\eqref#1{equation~\ref{#1}}

\def\1{\mathbbm{1}}

\newcommand{\E}{\mathbb{E}}

\newcommand{\R}{\mathbb{R}}

\newcommand{\KL}{D_{\mathrm{KL}}}
\newcommand{\Var}{\mathrm{Var}}

\newcommand{\Cov}{\mathrm{Cov}}

\DeclareMathOperator*{\argmax}{arg\,max}

\def\ra{{\textnormal{a}}}

\def\rx{{\textnormal{x}}}

\def\rz{{\textnormal{z}}}

\def\rva{{\mathbf{a}}}

\def\rvv{{\mathbf{v}}}

\def\rvx{{\mathbf{x}}}
\def\rvy{{\mathbf{y}}}

\def\rmA{{\mathbf{A}}}

\def\vmu{{\bm{\mu}}}
\def\vtheta{{\bm{\theta}}}
\def\va{{\bm{a}}}

\def\vg{{\bm{g}}}

\def\vm{{\bm{m}}}
\def\vn{{\bm{n}}}

\def\vp{{\bm{p}}}

\def\vv{{\bm{v}}}

\def\vx{{\bm{x}}}
\def\vy{{\bm{y}}}
\def\vz{{\bm{z}}}

\def\mA{{\bm{A}}}

\def\mH{{\bm{H}}}
\def\mI{{\bm{I}}}
\def\mJ{{\bm{J}}}

\def\mX{{\bm{X}}}

\def\mSigma{{\bm{\Sigma}}}

\DeclareMathAlphabet{\mathsfit}{\encodingdefault}{\sfdefault}{m}{sl}
\SetMathAlphabet{\mathsfit}{bold}{\encodingdefault}{\sfdefault}{bx}{n}

\def\sA{{\mathbb{A}}}
\def\sB{{\mathbb{B}}}

\def\sM{{\mathbb{M}}}

\def\sQ{{\mathbb{Q}}}
\def\sR{{\mathbb{R}}}
\def\sS{{\mathbb{S}}}

\def\sX{{\mathbb{X}}}
\def\sY{{\mathbb{Y}}}



%% file: Chapters/title.tex
\title{\LARGE \textsc{Towards Adversarial Robustness of Deep Vision Algorithms}}


\crest{\includegraphics[width=0.4\textwidth]{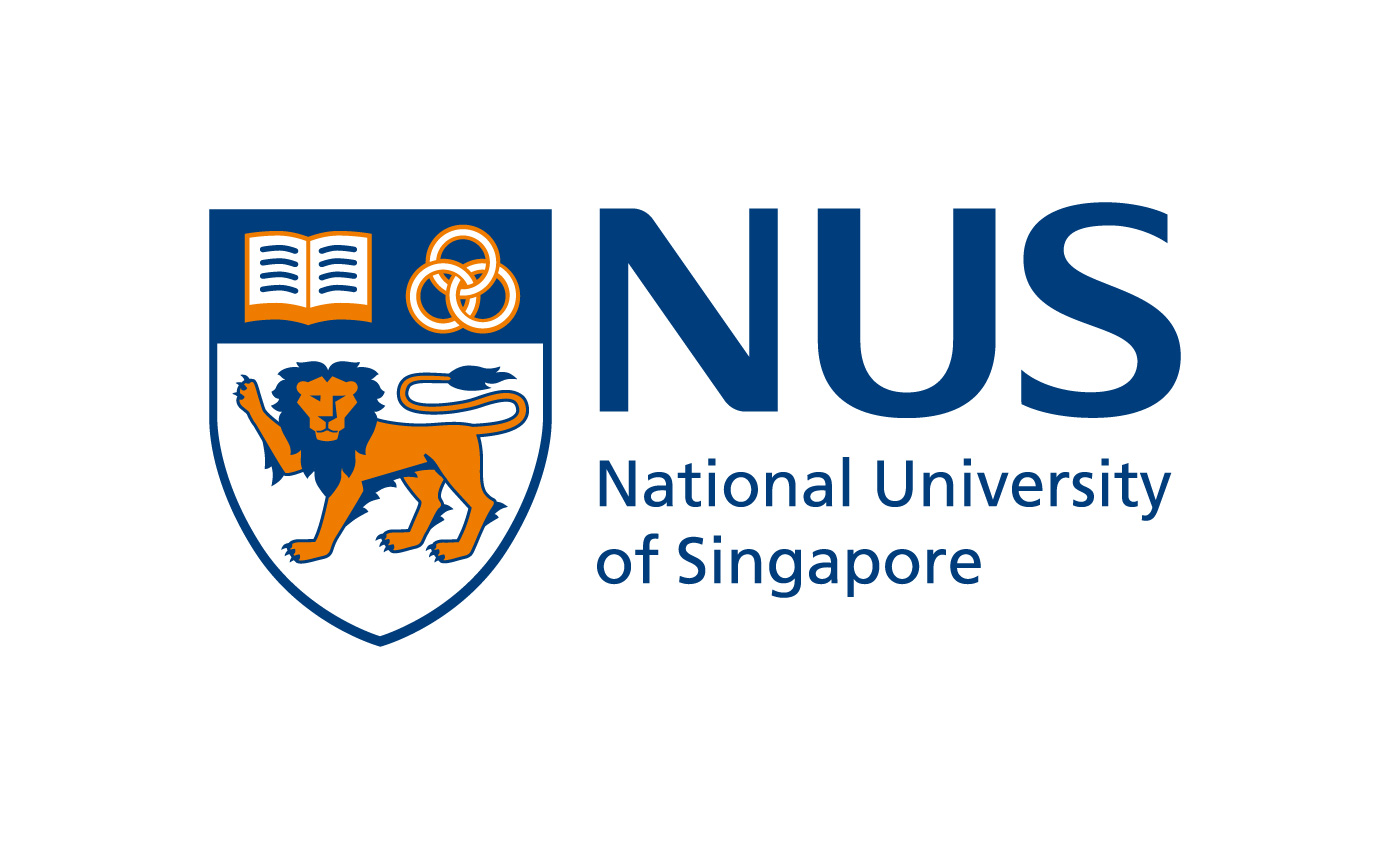}}


\author{
	\textbf{Hanshu Yan} \\
	{\large	( \textit{M.Sc., NUS; ~~B.Eng.\& B.Sc., BUAA} )} \\~\\
}

\Misc{
		\textbf{Supervisor}: \\
		Associate Professor Vincent Y. F. Tan\\
}




\dept{Department of Electrical and Computer Engineering}
\university{National University of Singapore}

%




\degreetitle{\textbf{Doctor of Philosophy}\\~\\}




%% file: Chapters/acknowledgments.tex
\begin{acknowledgements}      

This has been a happy, fortunate, and fruitful journey. It is also long and, sometimes, arduous. I am very grateful for the help of many individuals. It would have been much longer without their support.

I would like to thank my supervisors, Prof. Vincent Y. F. Tan and Dr. Jiashi Feng, for teaching me research thinking and supporting me in pursuing my research interests. They taught me to think deeply about the value and feasibility of potential problems. They scheduled weekly meetings to discuss ongoing projects and gave me suggestions on promising research directions. I have benefitted from their broad range of knowledge and deep insights into research problems. These help me establish my long-term research objectives. I also want to thank them for teaching me to write academic papers rigorously and present ideas cogently. Vincent always spares no effort to help me with English and mathematical writing. I am pleased that my writing and presentation skills have improved significantly during the past four years. Besides, I am grateful that they help me build connections with many brilliant researchers in academia and industry. Without the support and encouragement of my supervisors, this thesis would never have happened. They are excellent supervisors and good friends (\begin{CJK*}{UTF8}{gkai}良师益友\end{CJK*} in Chinese).

I would like to express my sincere gratitude to Dr. Jingfeng Zhang, Dr. Gang Niu, and Prof. Masashi Sugiyama from RIKEN-AIP and the University of Tokyo. I met with Jingfeng to discuss the robustness of deep neural networks when he was doing his Ph.D. in NUS. We were both interested in developing robust network architectures. After he joined the RIKEN-AIP, we carried out a series of collaborations to robustify deep vision algorithms. I truly appreciated that Prof. Sugiyama and Dr. Niu joined the discussion, helped examine the technical reasonableness, and edited research papers. Working with the team was a memorable and invaluable experience for my Ph.D. 

I would thank my friends in Vincent's group and Jiashi's group. They created a homely atmosphere where we support, inspire, and encourage each other. I particularly appreciate two individuals, Dr. Junhao Liew and Ms. Haiyun He. Junhao has always encouraged and comforted me like a big brother when the going gets tough. He also shared with me his experience and lessons in research unreservedly. Haiyun and I are in the same class. We helped each other in curriculum modules and collaborated on research work. The collaboration has been fruitful and enjoyable. I hope we can continue working together in the future. I also want to thank my lab mates, Jiawei Du, Mengjiun Chiou, Kuangqi Zhou, Yifan Zhang, Yanfei Dong, Dapeng Hu, Jian Liang, Kaixin Wang, Yujun Shi, Pan Zhou, and others, for the inspiring discussions and far too many things to mention here. 

I would like to thank many other students from different departments at NUS. It has been a wonderful experience to extend my research areas and apply deep learning techniques to scientific and engineering problems. 

Lastly, I am extremely indebted to my parents for their love, and the support in my pursuits and decisions. I am also very grateful to my girlfriend, Tan Ye, for her deep love, unwavering support, and company. Words cannot express my love and gratitude to all of them. 

\end{acknowledgements}

%% file: Chapters/abstract.tex
%
%
\clearpage
\chapter*{}
\addcontentsline{toc}{chapter}{Summary}

\thispagestyle{plain}
\vspace{-10em}
\begin{center}
	{\large \textbf{Towards Adversarial Robustness of Deep Vision Algorithms}}
	\vspace{1em}
	
	{Hanshu Yan}\blfootnote{Thesis Supervisor: Associate Professor Vincent Y. F. Tan}
	\vspace{1em}
	
	Submitted to the Department of Electrical and Computer Engineering in May 2022, in partial fulfillment of the requirements for the degree of Doctor of Philosophy
\end{center}

	\vspace{1em}
	\begin{flushleft}
		\textbf{Summary}
	\end{flushleft}
	Deep learning methods have achieved great successes in solving computer vision tasks and they have been widely utilized in artificially intelligent systems for image processing, analysis, and understanding. However, deep neural networks (DNNs) have been shown to be vulnerable to adversarial perturbations in input data. The security issues of deep neural networks have thus come to the fore. It is imperative to comprehensively study the adversarial robustness of deep vision algorithms. This thesis focuses on the robustness of deep classification models and deep image denoisers. 
	
	For image denoising, we systematically investigate the robustness of deep image denoisers. Specifically, we propose a novel attack method, observation-based zero-mean attack (ObsAtk), that considers the zero-mean assumption of natural noise, to generate adversarial perturbations of noisy input images. We develop an effective and theoretically-grounded PGD-based optimization technique to implement ObsAtk. With ObsAtk, we propose the hybrid adversarially training (HAT) to enhance the robustness of deep image denoisers. Extensive experiments demonstrate the effectiveness of HAT. Furthermore, we explore the connection between the adversarial robustness of denoisers and the adaptivity to unseen types of real-world noise. We find that deep denoisers that are HAT-trained only with synthetic noisy data can generalize well to unseen types of noise. The noise removal capabilities are even comparable to the denoisers trained with true real-world noise.

	For image classification, we explore novel robust architectures other than vanilla convolutional neural networks (CNNs). First, we study the robustness of neural ordinary differential equations (NODE). We empirically demonstrate that NODE-based classifiers show superior robustness against input perturbations in comparison to CNN-based classifiers. To further boost the robustness of NODE-based models, we introduce the time-invariant property to NODEs and impose a steady-state constraint to regularize the flow of ODEs on perturbed data. We demonstrate that the resultant model, dubbed as time-invariant steady neural ODE (TisODE),  is more robust than the vanilla NODE. 
	
	Second, we also investigate the robustness of vanilla CNNs from the perspective of channel-wise activations and propose a feature selection mechanism to enhance the robustness of vanilla CNNs. In particular, we compare the channel-wise activations of normally-trained classifiers when dealing with natural and adversarial data. We observe that adversarial data mislead the deep classifiers by over-activating negatively-relevant (NR) channels but under-activating positively-relevant (PR) ones. We also compare the channel-wise activations of normally-trained models to the adversarially-trained ones and observe that adversarial training robustifies models by promoting the under-activated PR channels and suppressing the over-activated NR ones. Thus, we hypothesize that scaling the activations of channels corresponding to their relevances to the true categories improves the robustness. To examine this hypothesis, we develop a novel channel manipulation technique, namely channel-wise importance-based feature selection (CIFS), that can scale channels' activations by generating non-negative multipliers based on their relevances. Extensive experimental results verify the hypothesis and the superior robustness of CIFS-modified CNNs.

	In summary, this thesis systematically studies the robustness of deep vision algorithms, including robustness evaluation (ObsAtk), robustness improvement (HAT, TisODE, and CIFS), and the connection between adversarial robustness and generalization capability to new domains. 
	\vfill

%% file: Chapters/listofsymbols.tex
\chapter{List of Symbols}

\begin{longtable}{p{.3\textwidth} p{0.7\textwidth}}

\multicolumn{2}{c}{ \textbf { Numbers and Arrays}} \\
	$\displaystyle a$ & A scalar (integer or real)\\
	$\displaystyle \va$ & A vector\\
	$\displaystyle \mA$ & A matrix\\
	$\displaystyle \mI_n$ & Identity matrix with $n$ rows and $n$ columns\\
	$\displaystyle \mI$ & Identity matrix with dimensionality implied by context\\
	$\displaystyle \text{diag}(\va)$ & A square, diagonal matrix with diagonal given by $\va$\\
	$\displaystyle \ra$ & A scalar random variable\\
	$\displaystyle \rva$ & A vector-valued random variable\\
	$\displaystyle \rmA$ & A matrix-valued random variable\\

~\\
\multicolumn{2}{c}{ \textbf { Sets}} \\
	$\displaystyle \sX$ & A set\\
	$\displaystyle \R$ & The set of real numbers \\
	$\displaystyle \{0, 1\}$ & The set containing 0 and 1 \\
	$\displaystyle \{0, 1, \dots, n \}$ & The set of all integers between $0$ and $n$\\
	$\displaystyle [a, b]$ & The real interval including $a$ and $b$\\
	$\displaystyle (a, b]$ & The real interval excluding $a$ but including $b$\\


~\\
\multicolumn{2}{c}{\textbf { Calculus} }\\
	$\displaystyle\frac{d y} {d x}$ & Derivative of $y$ with respect to $x$\\ [2ex]
	$\displaystyle \frac{\partial y} {\partial x} $ & Partial derivative of $y$ with respect to $x$ \\
	$\displaystyle \nabla_\vx y $ & Gradient of $y$ with respect to $\vx$ \\
	$\displaystyle \nabla_\mX y $ & Matrix derivatives of $y$ with respect to $\mX$ \\
	$\displaystyle \frac{\partial f}{\partial \vx} $ & Jacobian matrix $\mJ \in \R^{m\times n}$ of $f: \R^n \rightarrow \R^m$\\
	$\displaystyle \nabla_\vx^2 f(\vx)\text{ or }\mH( f)(\vx)$ & The Hessian matrix of $f$ at input point $\vx$\\
	$\displaystyle \int f(\vx) d\vx $ & Definite integral over the entire domain of $\vx$ \\
	$\displaystyle \int_\sS f(\vx) d\vx$ & Definite integral with respect to $\vx$ over the set $\sS$ \\

~\\
\multicolumn{2}{c}{\textbf { Probability and Information Theory} }\\
	$\displaystyle \ra \sim P$ & Random variable $\ra$ has distribution $P$\\
	$\displaystyle  \E_{\rx\sim P} [ f(x) ]\text{ or } \E f(x)$ & Expectation of $f(x)$ with respect to $P(\rx)$ \\
	$\displaystyle \Var(f(x)) $ &  Variance of $f(x)$ under $P(\rx)$ \\
	$\displaystyle \Cov(f(x),g(x)) $ & Covariance of $f(x)$ and $g(x)$ under $P(\rx)$\\
	$\displaystyle H(\rx) $ & Shannon entropy of the random variable $\rx$\\
	$\displaystyle \KL ( P \Vert Q ) $ & Kullback--Leibler divergence of $P$ and $Q$ \\
	$\displaystyle \mathcal{N} ( \vx ; \vmu , \mSigma)$ & Gaussian distribution %
	over $\vx$ with mean $\vmu$ and covariance $\mSigma$ \\

~\\
\multicolumn{2}{c}{\textbf { Functions} } \\
	$\displaystyle f: \sA \rightarrow \sB$ & The function $f$ with domain $\sA$ and range $\sB$\\
	$\displaystyle f \circ g $ & Composition of the functions $f$ and $g$ \\
	$\displaystyle f(\vx ; \vtheta) $ & A function of $\vx$ parametrized by $\vtheta$.
	(Sometimes we write $f(\vx)$ and omit the argument $\vtheta$ to lighten notation) \\
	$\displaystyle \log x$ & Natural logarithm of $x$ \\
	$\displaystyle \sigma(x)$ & Logistic sigmoid, $\displaystyle \frac{1} {1 + \exp(-x)}$ \\
	$\displaystyle \zeta(x)$ & Softplus, $\log(1 + \exp(x))$ \\
	$\displaystyle || \vx ||_p $ & $\ell_p$ norm of $\vx$ \\
	$\displaystyle \1_\mathrm{condition}$ & is 1 if the condition is true, 0 otherwise\\

\end{longtable}

%% file: Chapters/introduction.tex
\chapter{Introduction}
Intelligent systems have revolutionized various domains, including industry, research, and our daily lives \citep{lecun2015deep,bedi2019deep,veres2019deep,wang2019review}. For example, medical diagnostic systems facilitate health screening and assist doctors to diagnose diseases \citep{shen2017deep}; unmanned underwater vehicles broaden the horizons of scientists and automatically perform subsea operations and sample collections \citep{moniruzzaman2017deep}; smartphones have reinvented daily lives and made it easy to exchange information and connect with people \citep{abdelhamed_high-quality_2018,ma2019image}. Behind the success of intelligent systems, computer vision and other technology play vital roles in processing and analyzing information. Computer vision aims to design algorithms that can enhance and comprehend digital images effectively and efficiently \citep{krizhevsky2012imagenet,he2017mask,he_deep_2016}. For example, denoising and deblurring algorithms can improve the visual quality of medical images that are degraded by unknown noise and device movement \citep{yan2019unsupervised,amini2021medical}; classification algorithms help to recognize the categories of samples of marine life \citep{rathi2017underwater}; generative vision models are able to synthesize fancy objects that do not yet exist in the real world, to create engaging virtual environments \citep{goodfellow2014generative,kingma2013auto}.

In recent years, deep learning has achieved great successes in computer vision tasks \citep{he_deep_2016,zhao2019object,minaee2021image}. For example, deep classification networks have accomplished astonishing performances of real-world image recognition on several benchmark datasets \citep{liu2021swin}; the performances are comparable to human-level recognition capability. Deep image denoising models can remove intense RGB noise and reconstruct high-quality images from extremely low-light shots \citep{zhang_beyond_2017,pang_recorrupted--recorrupted_2021}. Due to these impressive results, deep learning-based algorithms have been widely applied to real-world applications, such as digital cameras \citep{liang2021cameranet}, medical imaging systems \citep{shen2017deep}, facial recognition systems \citep{wang2021deep}, and autonomous vehicles \citep{grigorescu2020survey}. 

Although deep learning algorithms perform very well on various benchmark datasets, they have been shown to be vulnerable and unstable when confronted with uncertainties in data. For example, Gaussian noise corruption in input images \citep{hendrycks2018benchmarking} degrades the classification performance. Furthermore, Goodfellow \citep{goodfellow_explaining_2015} showed that well-crafted imperceptible adversarial perturbations in input data can completely mislead a well-trained classification model and result in unreasonable predictions. Various sources of uncertainty are ubiquitous and inevitable in real-world applications, including inherent noise and observation errors. In high-stake applications like fraud detection \citep{raghavan2019fraud}, medical diagnosis \citep{shen2017deep}, and natural disaster prediction \citep{kang2016deep}, such high sensitivity to perturbations will lead to unacceptable economic losses and other catastrophic consequences. To build reliable and trustworthy intelligent systems, it is imperative to systematically study the robustness of deep vision algorithms.

This thesis focuses on the problem of evaluating and enhancing the robustness of deep vision algorithms against perturbations of input data. In particular, to examine the sensitivity of deep models against any arbitrary type of input noise, we develop adversarial attack methods to compute the worst-case perturbation with a certain limited budget. The resistance against adversarial perturbations is termed \tbf{\tit{adversarial robustness}}. Besides, to enhance the adversarial robustness of deep models, we devise solutions from the perspectives of network architectures, learning objectives, and training frameworks. Finally, we also explore the connection between adversarial robustness and generalization to new domains and empirically demonstrate that adversarial robustness benefits the models in terms of the ability to adapt to unseen types of data.

\section{Background}

Noise in observed signals is ubiquitous in real-world applications due to intrinsic noise, measuring errors, numerical errors induced in signal processing, etc. Reliable intelligent systems ought to exclude the interference of noise and extract salient task-related information from the noisy observations. To wit, they should not only effectively process and analyze the authentic ground truth but also maintain acceptable performances on corrupted data. The first step toward robust systems is developing methods to evaluate the robustness of deep models. In Section \ref{chp:intro_adv_robustness}, we introduce existing works that proposed attack methods to craft adversarial noise for the robustness evaluation. We also discuss several metrics for measuring the robustness level by using the attack methods. Based on these robustness metrics, we then introduce several state-of-the-art techniques for robustness improvement in Section \ref{chp:intro_robustify}. 

\subsection{Adversarial Robustness}
\label{chp:intro_adv_robustness}
Consider a deep neural network (DNN) $\{f_{\theta}:\theta\in \Theta\}$ mapping an input $\vx$ to a target $\vy$. The model is trained to minimize a certain loss function that is measured by particular distance measure $d(\cdot,\cdot)$ between the output $f_{\theta}(\vx)$ and the target $\vy$. In critical applications, the DNN should resist small perturbations on the input data and map the perturbed input to a result that is close to the target. The notion of \tit{robustness} has been proposed to measure the resistance of DNNs against slight changes of the input \citep{szegedy_intriguing_2014,goodfellow_explaining_2015}. The robustness is characterized by the distance $d(f_{\theta}(\vx'),\vy)$ between the prediction $f_{\theta}(\vx')$ and the target $\vy$, where the perturbed input $\vx'$ is located within a small $\rho$-neighborhood of the original input $\vx$.

\paragraph{Adversarial Examples} It is possible to use various types of random perturbations, such as Gaussian noise and uniform noise, to evaluate robustness. However, we, in reality, are incognizant of the true statistics of real-world noise. DNNs that are robust against random Gaussian noise may be very vulnerable to other types of perturbations. Thus, to comprehensively examine the robustness of deep models, we consider the worst-case perturbations of input data, namely $\vx'$ is synthesized by maximizing the distance between its output and the target $\vy$:
\begin{equation}
	\vx' = \argmax_{\vx':\|\vx'-\vx\|\leq \rho} d(f_{\theta}(\vx'),\vy).
	\label{eq:adv_attacks}
\end{equation}

The distance $d(f_{\theta}(\vy'),\vx)$ is an indication of the robustness of $f_{\theta}$ around $\vy$: a small distance implies strong robustness and vice versa. In terms of image classification, the $\rho$-neighborhood is usually defined by the $\ell_{\infty}$-norm and the distance $d(\cdot, \cdot)$ is measured by the cross-entropy loss \citep{madry_towards_2018} or a margin loss \citep{carlini_towards_2017}. For image restoration, such as image denoising and super-resolution, the distance between images is usually measured by the $\ell_2$-norm \citep{zhang_beyond_2017}. In addition to the constraint of the limited budget, there may be other constraints that are used in the optimization procedure and depend on the forms of the adversarial perturbations. For example, attackers may add the $\ell_0$-norm to restrict the number of pixels that are allowed to be changed \citep{croce2019sparse}; attachers may multiply the perturbation with a mask of certain shapes to generate adversarial perturbations with certain semantic meanings, such as adversarial T-shirts. Anyone wearing such T-shirts can protect themselves from malicious person detectors \citep{xu2020adversarial}. 

\paragraph{Adversarial Attack Methods} In general, the worst-case perturbation $\vx'$ can be approximated via backpropagation-based methods, such as Fast Gradient Signed Method (FGSM) \citep{goodfellow_explaining_2015}, Iterative-FGSM \citep{kurakin_adversarial_2017}, and Projected Gradient Descent (PGD) attack \citep{madry_towards_2018}, which approximately solve \eqref{eq:adv_attacks} via gradient descent methods. However, some deep models may involve randomization modules or non-differentiable operations \citep{xie_mitigating_2018}. In this case, backpropagation on its own may not be able to approximate the worst-case perturbations sufficiently accurately. To solve this problem, \citet{athalye_obfuscated_2018} demonstrated that attackers could circumvent the gradient obfuscation problem through the Expectation over Transformation (EoT) technique for randomization methods or the Backward Pass Differentiable Approximate (BPDA) technique for non-differentiable transformations. EoT updates perturbations with the average gradients that are computed by performing multiples times of forward and backward passes; BPDA approximates the gradients of non-differentiable modules by substituting them with differentiable modules.

Considering the case that attackers are not allowed to access the inner information of neural networks, researchers have proposed several black-box and transfer-based attack methods \citep{guo2019simple,liu2016delving}. Black-box methods usually iteratively update the adversarial perturbations by querying the target models for a limited number of times; transfer-based methods, instead of diagnosing the target models, attack substitute models to craft the adversarial perturbations. In comparison, gradient-based (white-box) attack methods, if they are allowed to access the gradients of deep models, are usually much stronger than other types of attacks \citep{tramer_adaptive_2020} because they are cognizant of the whole information of deep models. To further boost the strength and the adaptivity of attacks, \citet{croce_reliable_2020} proposed to use an ensemble of four white-box and black-box attacks, namely AutoAttack, to reliably evaluate the robustness of deep models.

\paragraph{Robustness Metrics}
Robustness is measured by the performance of deep models on perturbed input data. The metrics of the robustness of deep models depend on the tasks performed. In semantic-level classification tasks, we usually use the accuracy (or $0$-$1$ loss), $\1{\left[f_{\theta}(x)=y_j\right]}$, to measure the performance. Thus, the robustness of classification models is quantified by the average accuracy of adversarial examples over a test set $\sS$:
$$\overline{\text{Acc}} = \frac{1}{|\sS|} \sum_{(\vx, \vy)\in \sS} \1{\left[f_{\theta}(\vx')=\vy\right]}.$$
In pixel-level vision tasks, such as image reconstruction and image segmentation, each pixel has a certain ground-truth value. The metrics are usually defined based on the norm of the difference between pixel-value vectors. For example, in image denoising and deblurring, we use the peak signal-to-noise ratio (PNSR) to quantify the reconstruction quality, where PSNR is defined as:
$$\text{PSNR}(\vx, \vy) = 10 \cdot \log_{10}(\frac{\text{MAX}^2}{\text{MSE}} ),$$ 
where MAX represents the maximum possible value of pixels (e.g., 255 for 8-bit unsigned-integer images and 1 for floating-point images) and $\text{MSE}$ represents the mean-squared-loss between vector $\vx$ and $\vy$. Thus, we characterize the robustness via the metric of average PSNR over a test set $\sS$:
$$\overline{\text{PSNR}} = \frac{1}{|\sS|}\sum_{(\vx, \vy)\in \sS} \text{PSNR}(\vx', \vy).$$

\subsection{Robustifying Deep Neural Networks}
\label{chp:intro_robustify}
Deep learning models have been shown to be vulnerable to adversarial attacks under normal training (NT) \citep{tramer_adaptive_2020,yan_robustness_2020}. For example, a well-trained image classification model can achieve above 99\% accuracy on authentic images from the MNIST test set, but the accuracy will drop to around 0\% under some strong attacks like multiple-step PGD and C\&W attacks \citep{madry_towards_2018, carlini_towards_2017}. To robustify DNNs, researchers have explored a variety of techniques from the perspectives of model training and network architecture design, among others. 

\paragraph{Robust Training}
Robust training methods aim to devise novel training objectives to impose robustness regularization to the resultant models. The original idea of robust training is attributed to \citet{goodfellow_explaining_2015}, where the authors propose to use the FGSM adversarial examples for model training. However, models trained with FGSM adversarial examples cannot effectively defend other attacks like iteratively-FGSM \citep{kurakin_adversarial_2017} and C\&W attacks \citep{carlini_towards_2017}. The formal statement of PGD-adversarial training (AT) is proposed in \citet{madry_towards_2018}, where AT is formulated as the following min-max optimization problem,
\begin{equation}
	\min_{\theta\in \Theta} \max_{\vx':\|\vx'-\vx\|\leq \rho } d(f_{\theta}(\vx'),\vy).
	\label{eq:pgd-at}
\end{equation}
AT aims to guarantee the performance of models on the worst-case perturbed data. In practice, the PGD-AT method generates adversarial examples on the fly for each mini-batch via PGD attacks. Then, PGD adversarial examples are used to calculate training losses for updating model parameters. The effectiveness of AT has been verified by extensive empirical and theoretical studies \citep{yan_cifs_2021,gao_convergence_2019}. For further improvement, many variants of AT have been proposed in terms of its robustness enhancement \citep{cai_curriculum_2018,wang_improving_2020}, generalization to non-adversarial data \citep{zhang_theoretically_2019,zhang_attacks_2020}, and computational efficiency \citep{shafahi_adversarial_2019,wong_fast_2019}. \citet{wang_improving_2020} proposed the misclassification-aware-adversarial-training (MART) to further robustify DNNs, where MART simultaneously applies the misclassified natural data and the adversarial data for model training. To mitigate the trade-off between the different accuracies on adversarial data and authentic non-perturbed data, \citet{zhang2019theoretically} proposed the TRADES method that minimizes the classification loss on authentic data and the local sensitivity of deep model around training data at the same time. Additionally, due to the high extra computational overhead incurred by generating strong adversarial examples, \citet{shafahi_adversarial_2019} proposed Free-AT to accelerate AT by reusing the gradient information computed when updating model parameters. AT is a generally applicable framework that can be used in different tasks, including computer vision and natural language processing \citep{zhu2019freelb,gan2020large}, AT-based methods thus have become the default option for enhancing the robustness of DNNs.

In addition, some other works introduced various types of regularization for training models, such as layer-wise feature matching \citep{sankaranarayanan_regularizing_2018, liao_defense_2018, kannan_adversarial_2018}, low-rank representations \citep{sanyal_robustness_2020, mustafa_adversarial_2019}, attention map alignment \citep{xu_interpreting_2019}, and Lipschitz regularity \citep{Virmaux_lipschitz_2018, cisse_parseval_2017}. These types of regularization can work in conjunction with AT and benefit the models' robustness.

\paragraph{Robust Architecture Design} Apart from robust training strategies, researchers also have explored robust network architectures. A line of works propose to utilize pre-processing modules to remove adversarial noise from images or introduce randomness to ruin the adversarial patterns \citep{samangouei2018defense,yang2019me,xie_mitigating_2018}, so that adversarial perturbations will not affect the pre-trained models. Although these types of methods can defend pure gradient backpropagation attacks, such as I-FGSM, they can be easily circumvented by adaptive attacks like BPDA or EoT \citep{athalye_obfuscated_2018}. Another line of works attempt to modify the basic components of DNNs. For example, \citet{guo_sparse_2018} demonstrated that the robustness of DNNs can be improved by appropriately designed higher model sparsity. \citet{xie_feature_2019} performed feature denoising at the end of convolutional layers to ameliorate the adversarial effects on feature maps. \citet{zoran_towards_2020} used the spatial attention mechanism to highlight important regions in feature maps. Existing empirical works mostly focus on the spatial modification of feature maps in convolutional neural networks (CNNs). The channel-wise selection of convolutional layers and other families of models, such as neural ordinary differential equations \citep{chen2018neural}, energy-based models \citep{du2019implicit}, and vision transformers \citep{liu2021swin}, have been underexplored.


\section{Thesis Focus and Main Contributions}

This thesis systematically studies the adversarial robustness of deep vision algorithms, from the perspective of robustness evaluation, robustness enhancement, and the connection between robustness and the adaptation to unseen data. We solve the robustness evaluation and enhancement problems by developing a novel PGD-based optimization method to perform attacking, exploring robust network architectures, and effective adversarial training strategies.

\begin{itemize}
	\item Since image denoising algorithms have been extensively used to enhance image quality in computer vision systems, it is necessary to examine the robustness of deep image noises against adversarial corruptions. We propose the observation-based zero-mean attack (ObsAtk), which considers the zero-mean assumption of natural noise, for robustness evaluation. By using ObsAtk, we develop a novel adversarial training strategy, hybrid adversarial training (HAT), to enhance the robustness of deep image denoisers. Besides, we reveal that deep image denoisers that are adversarially trained with only synthetic data can generalize or adapt well to unseen types of real-world noise. In this case, one can train an effective denoiser without the need to collect a large amount of real-world noisy data for training. This work will be presented at the International Joint Conference on Artificial Intelligence (IJCAI), 2022 \citep{yan2022towards}.
	\item Neural ordinary differential equations (NODE) \citep{chen2018neural}, a new family of models, approximate nonlinear mappings by using continuous-time ODEs. They have attracted much attention in various research domains recently due to their desirable properties, such as invertibility and time-continuity. We study the robustness property of NODEs and found that NODE-based image classifiers enjoy superior robustness against random noise and adversarial perturbations in comparison to CNN-based ones. To further boost the robustness of NODEs, we propose time-invariant steady neural ODE (TisODE), which regularizes the flow of neural ODEs via the time-invariant property and a steady-state constraint. We empirically demonstrate that TisODE outperforms vanilla NODEs and can work in conjunction with other robustification techniques. This work appeared at the International Conference on Learning Representations (ICLR), 2020 \citep{yan_robustness_2020}.
	\item In CNN classifiers, channels in deep layers can extract semantic features, and the final predictions are made by aggregating information from various channels. In this case, abnormal activations of channels in deep layers may result in incorrect predictions. To understand the effects of adversarial data on each channel, we compare the activations of channels caused by authentic and adversarial data. We find that natural data tend to over-activate positively-relevant (PR) channels and under-activate negatively-relevant (NR) channels, but adversarial data do the opposite. Here, the relevance of each channel is defined as the gradient of the prediction logit associated with the true label with respect to the activation level. To investigate the effects of adversarial training on channels, we also compare the channel activations between normally-trained and adversarially-trained models. We observe that AT robustifies CNNs by aligning the channel-wise activations of adversarial data with those of their natural counterparts. Given these observations, we hypothesize that suppressing NR channels and promoting PR ones based on their relevances can further enhance the robustness of CNNs. To examine this hypothesis, we propose the Channel-wise Importance-based Feature Selection (CIFS). The CIFS mechanism manipulates channels’ activations of certain layers by generating non-negative multipliers to these channels based on their relevances. Extensive experiments on benchmark datasets clearly verify the hypothesis and CIFS’s effectiveness in robustifying CNNs. This work appeared at the International Conference on Machine Learning (ICML), 2021 \citep{yan_cifs_2021}.
\end{itemize}

\section{Organization of Thesis}
In Chapter \ref{chp:ijcai2022}, we propose the ObsAtk to evaluate the robustness of deep image denoisers and develop the HAT to enhance the robustness of deep denoisers by using ObsAtk. 
In Chapter \ref{chp:iclr2020}, we demonstrate that NODE-based classifiers are more robust against input perturbations in comparison to CNN-based ones and propose TisODE to further boost the robustness of NODEs. 
In Chapter \ref{chp:icml2021}, we propose the CIFS mechanism to modify CNN-based classifiers and show that CIFS clearly robustifies CNNs against adversarial corruptions. 
Finally, Chapter \ref{chp:conclusion} concludes this thesis and suggests several promising topics for future research.

%% file: Chapters/ijcai2022.tex
\chapter{Towards Adversarially Robust Deep Image Denoising}
\label{chp:ijcai2022}
\nocite{yan2022towards}

This work systematically investigates the adversarial robustness of deep image denoisers (DIDs), i.e, how well DIDs can recover the ground truth from noisy observations degraded by adversarial perturbations. Firstly, to evaluate DIDs' robustness,  we propose a novel adversarial attack, namely Observation-based Zero-mean Attack (ObsAtk), to craft adversarial zero-mean perturbations on given noisy images. We find that existing DIDs are vulnerable to the adversarial noise generated by ObsAtk. Secondly, to robustify DIDs, we propose an adversarial training strategy, hybrid adversarial training (HAT), that jointly trains DIDs with adversarial and non-adversarial noisy data to ensure that the reconstruction quality is high and the denoisers around non-adversarial data are locally smooth. The resultant DIDs can effectively remove various types of synthetic and adversarial noise. We also uncover that the robustness of DIDs benefits their generalization capability on unseen real-world noise. Indeed, HAT-trained DIDs can recover high-quality clean images from real-world noise even without training on real noisy data. Extensive experiments on benchmark datasets, including Set68, PolyU, and SIDD, corroborate the effectiveness of ObsAtk and HAT.

\section{Introduction}
Image denoising, which aims to reconstruct clean images from their noisy observations, is a vital part of the image processing systems. The noisy observations are usually modeled as the addition between ground-truth images and zero-mean noise maps \citep{dabov_image_2007,zhang_beyond_2017}. Recently, deep learning-based methods have made significant advancements in denoising tasks \citep{zhang_beyond_2017,anwar_real_2019} and have been applied in many areas including medical imaging \citep{gondara_medical_2016} and photography \citep{abdelhamed_high-quality_2018}. Despite the success of deep denoisers in recovering high-quality images from a certain type of noisy images, we still lack knowledge about their robustness against adversarial perturbations, which may cause severe safety hazards in high-stake applications like medical diagnosis. To address this problem, the first step should be developing attack methods dedicated for denoising to evaluate the robustness of denoisers. In contrast to the attacks for classification \citep{goodfellow_explaining_2015,madry_towards_2018}, attacks for denoising should consider not only the adversarial budget but also some assumptions of natural noise, such as zero-mean, because certain perturbations, such as adding a constant value, do not necessarily result in visual degradation. Although \citet{choi_deep_2021,choi_evaluating_2019} studied the vulnerability for various deep image processing models, they directly applied the attack from classification. To the best of our knowledge, no attacks are truly dedicated for the denoising task till now.

To this end, we propose the observation-based zero-mean attack (ObsAtk), which crafts a worst-case zero-mean perturbation for a noisy observation by maximizing the distance between the output and the ground-truth. To ensure that the perturbation satisfies the adversarial budget and the zero-mean constraints, we utilize the classical projected-gradient-descent (PGD) \citep{madry_towards_2018} method for optimization, and develop a two-step operation to project the perturbation back into the feasible region. Specifically, in each iteration, we first project the perturbation onto the zero-mean hyperplane. Then, we linearly rescale the perturbation to adjust its norm to be less or equal to the adversarial budget.  We examine the effectiveness of ObsAtk on several benchmark datasets and find that deep image denoisers are indeed susceptible to ObsAtk: the denoisers cannot remove adversarial noise completely and even yield atypical artifacts, as shown in Figure \ref{fig:adv-noise-advout}.

To robustify deep denoisers against adversarial perturbations, we propose an effective adversarial training strategy, namely hybrid adversarial training (HAT), to train denoisers by using adversarially noisy images and non-adversarial noisy images together. The loss function of HAT consists of two terms. The first term ensures the reconstruction performance from common non-adversarial noisy images, and the second term ensures the reconstructions between non-adversarial and adversarial images to be close to each other. Thus, we can obtain denoisers that perform well on both non-adversarial noisy images and their adversarial perturbed versions. Extensive experiments on benchmark datasets verify the effectiveness of HAT. 

Moreover, we reveal that adversarial robustness benefits the generalization capability to unseen types of noise, i.e., HAT can train denoisers for real-world noise removal only with synthetic noise sampled from common distributions like Gaussians. That is because ObsAtk searches for the worst-case perturbations around different levels of noisy images, and training with adversarial data ensures the denoising performance on various types of noise. In contrast, other reasonable methods for real-world denoising \citep{guo_toward_2019,lehtinen_noise2noise_2018} mostly require a large number of real-world noisy data for the training, which are unfortunately not available in some applications like medical radiology. We conduct experiments on several real-world datasets. Numerical and visual results demonstrate the effectiveness of HAT for real-world noise removal.

In summary, there are three main contributions in this work: 1) We propose a novel attack, ObsAtk, to generate adversarial examples for noisy observations, which facilitates the evaluation of the robustness of deep image denoisers. 2) We propose an effective adversarial training strategy, HAT, for robustifying deep image denoisers. 3) We build a connection between adversarial robustness and the generalization to unseen noise, and show that HAT   serves as a promising framework for training generalizable deep image denoisers.

\section{Notation and Background}

\paragraph{Deep image denoising} During image capturing, unknown types of noise may be induced by physical sensors, data compression, and transmission. Noisy observations are usually modeled as the addition between the ground-truth images and certain zero-mean noise \citep{dabov_image_2007,zhang_poisson_gaussian_2019}, i.e., $\rvy = \rvx + \rvv$ with $\E_{Q}\big[ \sum_{i=1}^{m} \rvv_{[i]} \big] = 0$, where $\rvv_{[i]}$ is the $i^{\text{th}}$ element of $\rvv$. The random vector $\rvx \in \sR^m$ with  distribution $P$ denotes a random clean image and the noise $\rvv \in \sR^m$ with a distribution $Q$ satisfies the zero-mean constraint. 
Denoising techniques aim to recover clean images from their noisy observations \citep{zhang_beyond_2017,dabov_image_2007}. Suppose we are given a training set  $\sS=\{(\vy_j, \vx_j)\}^{N}_{j=1}$ of noisy and clean image pairs sampled from distributions $Q$ and $P$ respectively, we can train a DNN to effectively remove the noise induced by distribution $Q$ from the noisy observations. A series of DNNs have been developed for denoising in recent years, including DnCNN~\citep{zhang_beyond_2017}, FFDNet~\citep{zhang_ffdnet_2018}, and RIDNet~\citep{anwar_real_2019}.

In real-world applications \citep{abdelhamed_high-quality_2018,xu_multi-channel_2017}, the noise distribution $Q$ is usually unknown due to the complexity of the image capturing procedures; besides, collecting a large number of image pairs (clean/noisy or noisy/noisy) for training sometimes may be unrealistic in safety-critical domains such as medical radiology \citep{zhang_poisson_gaussian_2019}. To overcome these, researchers developed denoising techniques by approximating real noise with common distributions like Gaussian or Poisson \citep{dabov_image_2007,zhang_poisson_gaussian_2019}. 
To train denoisers that can deal with different levels of noise, where the noise level is measured by the energy-density $\|\vv\|^2_2/m$ of noise, the training set may consist of noisy images sampled from a variety of noise distributions \citep{zhang_beyond_2017}, whose expected energy-densities range from zero to certain budget $\gamma^2$ (the expected $\ell_2$-norms range from zero to $\gamma \sqrt{m}$).
For example, $\sS^{\gamma}=\{(\vy_j, \vx_j)\}^{N}_{j=1}$  where $\vy_j=\vx_j+\vv_j$ and $\vx_j$ and $\vv_j$ are sampled from $P$ and $Q$ respectively and where $Q$ is randomly selected from a set of Gaussian distributions 
$\sQ^\gamma = \{ \mcal N(\bm{0}, \sigma^2 \mathbf{I})|   \sigma\in [0, \gamma] \}$. 
The denoiser $f_{\theta}^{\gamma}(\cdot)$ trained with $\sS^{\gamma}$ is termed as an $\gamma$-denoiser.

\paragraph{On robustness of deep image denoisers}
In practice, data storage and transmission may induce imperceptible perturbations on the original data so that the perturbed noise may be statistically slightly different from the noise sampled from the specific original distribution.
Although an $\gamma$-denoiser can successfully remove noise sampled from $Q \in \sQ^{\gamma}$, the performance of noise removal on the perturbed data is not guaranteed. Thus, we propose a novel attack method, ObsAtk, to assess the adversarial robustness of DIDs in Section \ref{sec:obsattack}. To robustify DIDs, we propose an adversarial training strategy, HAT, in Section \ref{sec:hat}. HAT-trained DIDs can effectively denoise adversarial perturbed noisy images and preserve good performance on non-adversarial data. 

Besides the adversarial robustness issue, it has been shown that $\gamma$-denoisers trained with $\sS^{\gamma}$ cannot generalize well to unseen real-world noise \citep{lehtinen_noise2noise_2018,batson_noise2self_2019}. 
Several methods have been proposed for real-world noise removal, but most of them require a large number of real noisy data for training, e.g., CBDNet (clean/noisy pairs) \citep{guo_toward_2019} and Noise2Noise (noisy pairs) \citep{lehtinen_noise2noise_2018}, which is sometimes impractical. In Section \ref{sec:hat_for_unseen}, we show that HAT-trained DIDs can generalize well to unseen real noise without the need of utilizing real noisy images for training.

\section{ObsAtk for Robustness Evaluation}
\label{sec:obsattack}
In this section, we propose a novel adversarial attack, Observation-based Zero-mean Attack (ObsAtk), to evaluate the robustness of DIDs. We also conduct experiments on benchmark datasets to demonstrate that normally-trained DIDs are vulnerable to adversarial perturbations.

\subsection{Observation-based Zero-mean Attack}
\begin{figure}[t!]
	\centering
	\includegraphics[width=.6\linewidth]{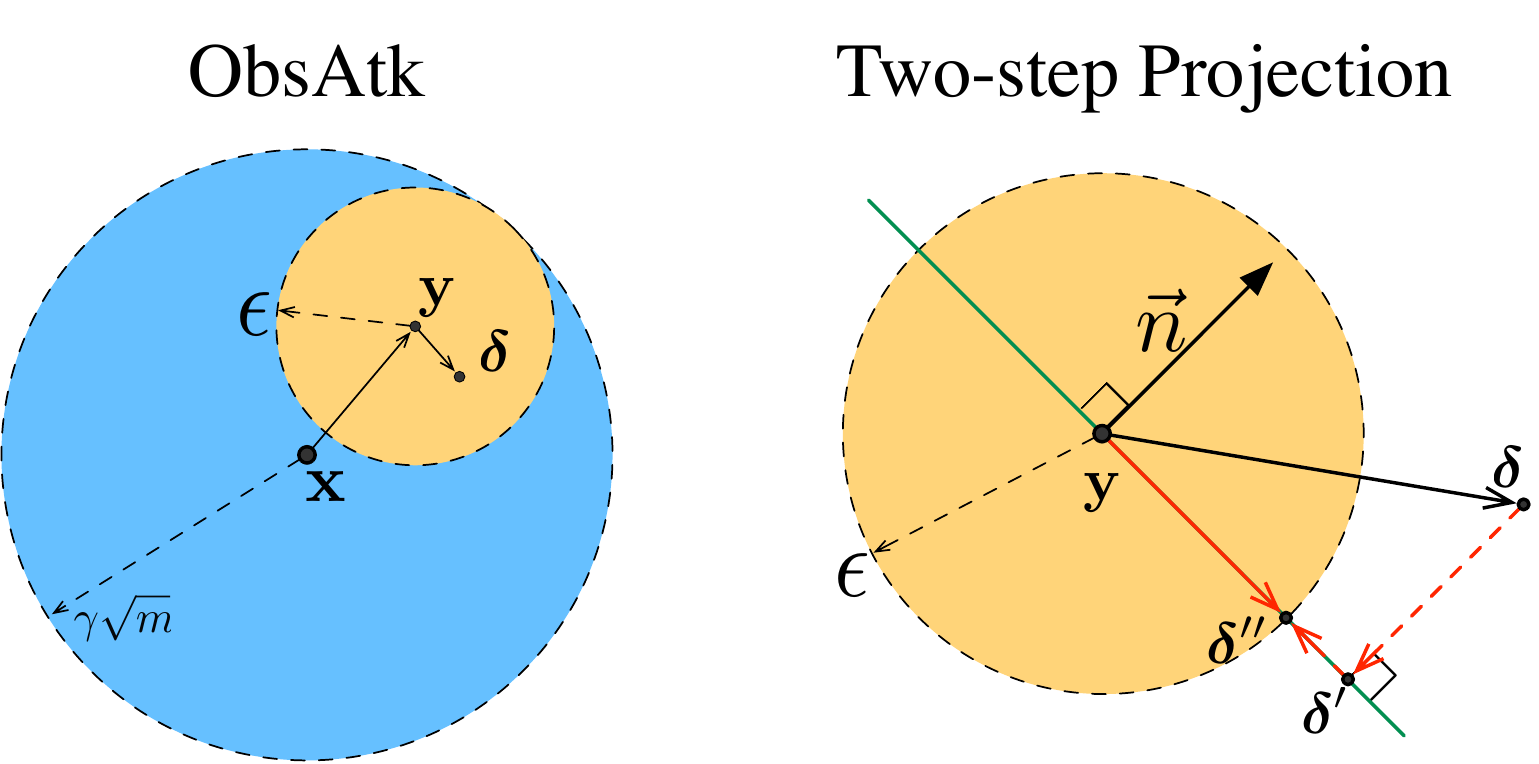}
	\caption[Illustration of ObsAtk.]{Illustration of ObsAtk. Left: We perturb a noisy observation $\vy$ of the ground-truth $\vx$ with an adversarial budget $\epsilon$ in the $\ell_2$-norm. For an $\gamma$-denoiser, we choose a proper value of $\epsilon$ to ensure the norm of the total noise is bounded by $\gamma\sqrt{m}$, where $m$ denotes the image size. Right: The perturbation $\vdelta$ is projected via the two-step operation onto the region defined by the zero-mean and $\epsilon$-ball constraints.}
	\label{fig:obsattack}
\end{figure}
An $\gamma$-denoiser $f^{\gamma}_{\theta}(\cdot)$ can generate a high-quality reconstruction $f^{\gamma}_{\theta}(\vy)$ close to the ground-truth $\vx$ from a noisy observation $\vy=\vx+\vv$. To evaluate the robustness of $f^{\gamma}_{\theta}(\cdot)$ with respect to a perturbation on $\vy$, we develop an attack to search for the worst perturbation $\bm{\delta}^*$ that degrades the recovered image $f^{\gamma}_{\theta}(\vy+\bm{\delta}^*)$ as much as possible. Formally, we need to solve the problem stated in \eqref{eq:obsattack}. The optimization problem is subject to \tit{two} constraints: The first constraint requires the norm of $\bm{\delta}$ to be bounded by a small adversarial budget $\epsilon$.
The second constraint restricts the mean $M(\bm{\delta})$ of all elements in $\bm{\delta}$ to be zero.
This corresponds to the zero-mean assumption of noise in real-world applications because a small mean-shift does not necessarily result in visual noise. For example, a mean-shift in gray-scale images implies a slight change of brightness. Since the zero-mean perturbation is added to a noisy observation $\vy$, we term the proposed attack as Observation-based Zero-mean Attack (ObsAtk).
\begin{subequations}
\begin{align}
	\bm{\delta}^* & = \argmax_{\bm{\delta} \in \sR^m} \| f_{\theta}^{\gamma}(\vy+\bm{\delta}) - \vx\|_2^2 , \\
	\text{s.t.~~} &   \|\bm{\delta}\|_2 \leq \epsilon, \;\;\text{~}\;\; M(\bm{\delta})=\frac{1}{m}\sum_{i=1}^{m} \bm{\delta}_{[i]}=0.\label{eq:obsattack-constraints}
\end{align}
\label{eq:obsattack}
\end{subequations}

We solve the constrained optimization problem Eq.~\eqref{eq:obsattack} by using the classical projected-gradient-descent (PGD) method. PGD-like methods update optimization variables iteratively via gradient descent and ensure the constraints to be satisfied by projecting parameters back to the feasible region at the end of each iteration. To deal with the $\ell_2$-norm and zero-mean constraints, we develop a two-step operation in Eq.~\eqref{eq:two-step-projection}, that first projects the perturbation $\bm{\delta}$ back to the zero-mean hyperplane and then projects the result onto the $\epsilon$-neighborhood.
\begin{subequations}
\begin{align}
	\bm{\delta}' &= \bm{\delta} - \frac{\bm{\delta}^\top \vn}{\|{\vn}\|^2_2} {\vn},\quad\mbox{where}\quad \vn = [1,1, \ldots, 1]^\top, \label{eq:zero-mean-proj} \\
	\bm{\delta}'' &= \min\Big(\frac{\epsilon}{\|\bm{\delta}'\|_2}, 1\Big) \text{~} \bm{\delta}'. \label{eq:proj-norm}
\end{align}
\label{eq:two-step-projection}
\end{subequations}

In each iteration, as shown in Figure~\ref{fig:obsattack}, the first step involves  projecting the perturbation $\bm{\delta}$ onto the zero-mean hyperplane. The zero-mean hyperplane consists of all the vectors $\vz$ whose mean of all elements equals zero, i.e., $\vn^{\top} \vz=0$, where $\vn$ is the length-$d$ all ones vector. Thus, $\vn$ is a normal of the zero-mean plane. We can project any vector onto the zero-mean plane via \eqref{eq:zero-mean-proj}. The vector $\bm{\delta}$ is first projected along the direction of $\vn$, then its projection $\bm{\delta}'$ onto the zero-mean plane equals itself minus its projection onto $\vn$.
The second step involves further projecting $\bm{\delta}'$ back to the $\epsilon$-ball via linear scaling. If  $\bm{\delta}'$ is already within the $\epsilon$-ball, we keep $\bm{\delta}'$ unchanged. Otherwise, the final projection $\bm{\delta}''$ is obtained by scaling $\bm{\delta}'$ with a factor ${\epsilon}/{\|\bm{\delta}'\|_2}$. For any two sets $A$ and $B$, although the projection onto $A\cap B$ is, in general, not equal to the result obtained by first projecting onto $A$, then onto $B$, surprisingly, the following holds for the two sets  in \eqref{eq:obsattack-constraints}.

\begin{algorithm}[t!]
	\caption{ObsAtk}
	\label{alg:obsattack}
	\begin{algorithmic}[1]
		\Require Denoiser $f_{\theta}(\cdot)$, ground-truth $\vx$, noisy observation $\vy$, adversarial budget $\epsilon$, \#iterations $T$, step-size $\eta$, minimum pixel value $p_{\text{min}}$, maximum pixel value $p_{\text{max}}$
		\Ensure Adversarial perturbation $\bm{\delta}$
		\State $\bm{\delta} \leftarrow \mathbf{0}$
		\For{$t=1$ to $T$}
		\State $\bm{\delta} \leftarrow \bm{\delta} + \eta\nabla_{\bm{\delta}}\| f_{\theta}^{\gamma}(\vy+\bm{\delta}) - \vx\|_2^2$; 
		\State $\bm{\delta} \leftarrow \bm{\delta} - (\bm{\delta}^\top \vn /{\|{\vn}\|^2_2}) {\vn}$ where $\vn$ is in \eqref{eq:zero-mean-proj}
		\State $\bm{\delta} \leftarrow \min(\nicefrac{\epsilon}{\|\bm{\delta}\|_2}, 1) \bm{\delta}$;
		\EndFor
		\State $\bm{\delta} \leftarrow \text{Clip}(\vy+\bm{\delta}, p_{\text{min}}, p_{\text{max}}) - \vy$
	\end{algorithmic}
\end{algorithm}

\begin{theorem}[Informal]
 Given any vector $\bm{\delta} \in \sR^m$, the projection of $\bm{\delta} $ via the two-step operation in \eqref{eq:two-step-projection} satisfies the two constraints in \eqref{eq:obsattack-constraints}, and the two-step projection is equivalent to the exact projection onto the set defined by \eqref{eq:obsattack-constraints}.
	\label{thm:two-step-proj}
\end{theorem}
The formal statement and the proof of Theorem \ref{thm:two-step-proj} are provided in Appendix \ref{sec:thm1-proof}. The complete procedure of ObsAtk is summarized in Algorithm \ref{alg:obsattack}.

\def \SubFigWidth {0.2} 
\def \SubImgWidth {1}
\begin{figure}[t!]
    \centering
    \begin{subfigure}{\SubFigWidth\linewidth}
        \centering
        \includegraphics[width=\SubImgWidth \linewidth]{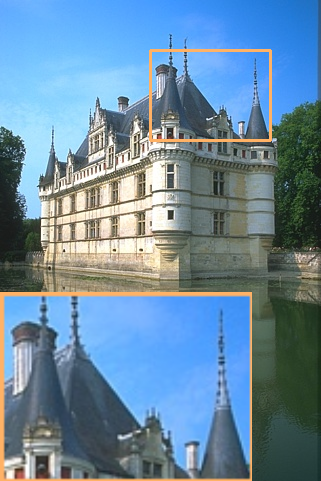}
        \caption{ $\vx$}
    \end{subfigure}
    \begin{subfigure}{\SubFigWidth\linewidth}
        \centering
        \includegraphics[width=\SubImgWidth \linewidth]{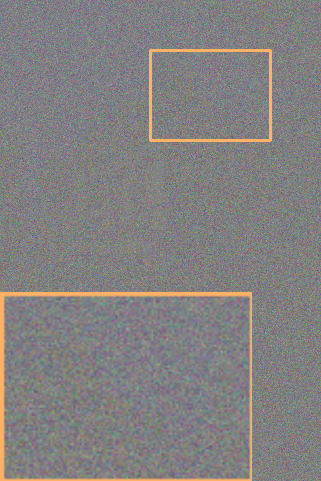}
        \caption{ $\vv$}
    \end{subfigure}
    \begin{subfigure}{\SubFigWidth\linewidth}
        \centering
        \includegraphics[width=\SubImgWidth \linewidth]{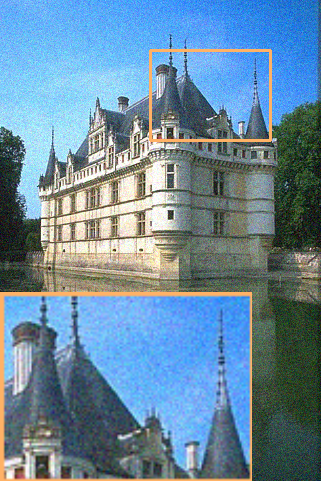}
        \caption{ $\vy$}
    \end{subfigure}
    \begin{subfigure}{\SubFigWidth\linewidth}
        \centering
        \includegraphics[width=\SubImgWidth \linewidth]{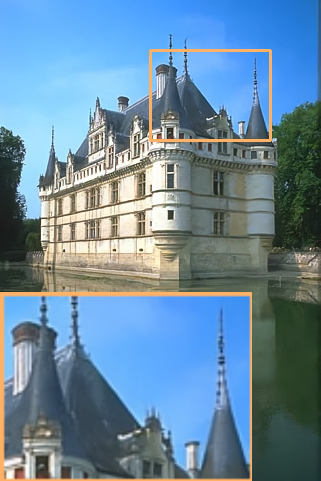}
        \caption{ $f^{\gamma}_{\theta}(\vy)$}
    \end{subfigure}

    \begin{subfigure}{\SubFigWidth\linewidth}
	\centering
		\begin{minipage}{1in}
		\hfill\vspace{1.5in}
		\end{minipage}
	\end{subfigure}
    \begin{subfigure}{\SubFigWidth\linewidth}
        \centering
        \includegraphics[width=\SubImgWidth \linewidth]{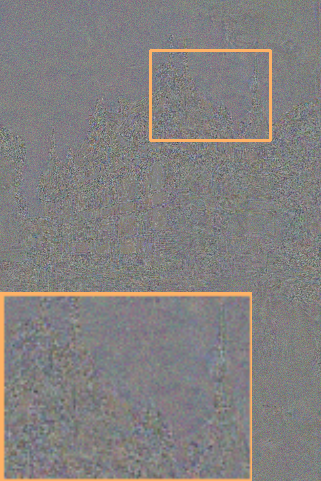}
        \caption{ $\vv+\bm{\delta}$}
    \end{subfigure}
    \begin{subfigure}{\SubFigWidth\linewidth}
        \centering
        \includegraphics[width=\SubImgWidth \linewidth]{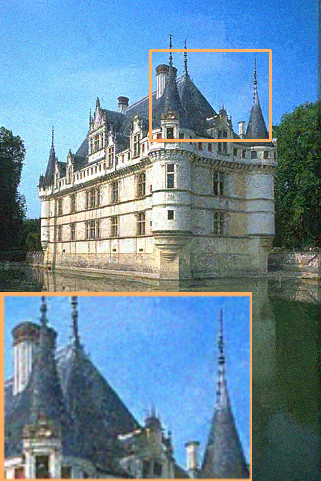}
        \caption{ $\vy+\bm{\delta}$}
    \end{subfigure}
    \begin{subfigure}{\SubFigWidth\linewidth}
        \centering
        \includegraphics[width=\SubImgWidth \linewidth]{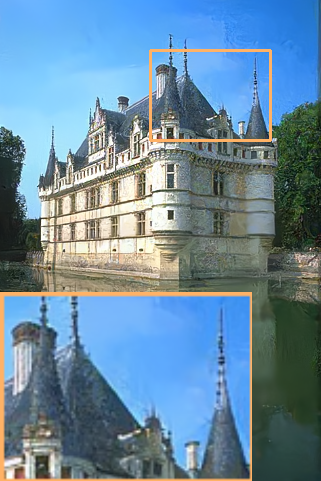}
        \caption{ $f^{\gamma}_{\theta}(\vy+\bm{\delta})$}
        \label{fig:adv-noise-advout}
    \end{subfigure}
	\caption[Vulnerability of normally-trained deep denoisers under adversarial attacks.]{
		Given a normally-trained denoiser $f^{\gamma}_{\theta}(\cdot)$, from left to right are the ground-truth image $\vx$, Gaussian noise $\vv$, the Gaussian noisy image $\vy=\vx+\vv$, the reconstruction  $f^{\gamma}_{\theta}(\vy)$ from $\vy$, adversarial noise $\vv+\bm{\delta}$, the adversarially noisy image $\vy+\bm{\delta}$, and the reconstruction $f^{\gamma}_{\theta}(\vy+\bm{\delta})$ from $\vy+\bm{\delta}$. Comparing (a), (d) and (g), we observe that $f^{\gamma}_{\theta}(\cdot)$ can effectively remove Gaussian noise but its performance is degraded when dealing with the adversarial noise (noise remains on the roof and strange contours appear in the sky).
		} 
	\label{fig:adv-noise}
\end{figure}

\subsection{Robustness Evaluation via ObsAtk}

We use ObsAtk to evaluate the adversarial robustness of $\gamma$-denoisers on several gray-scale and RGB benchmark datasets, including Set12, Set68, BSD68, and Kodak24. For gray-scale image denoising, we use Train400 to train a DnCNN-B \citep{zhang_beyond_2017} model, which consists of 20 convolutional layers. We follow the training setting in Zhang \etal \citet{zhang_beyond_2017} and randomly crop $128\times 3000$ patches in size of $50\times50$. Noisy and clean image pairs are constructed by injecting different levels of white Gaussian noise into clean patches. The noise levels $\sigma$ are uniformly randomly selected from $[0, \gamma]$ with $\gamma=\nicefrac{25}{255}$. For RGB image denoising, we use BSD432 (BSD500 excluding images in BSD68) to train a DnCNN-C model with the same number of layers as DnCNN-B and but set the input and output channels to be three. Other settings follow those of the training of DnCNN-B. 

We evaluate the denoising capability of the $\gamma$-denoiser on Gaussian noisy images and their adversarially perturbed versions. The image quality of reconstruction is measured via the peak-signal-noise ratio (PSNR) metric. A large PSNR between reconstruction and ground-truth implies a good performance of denoising. We denote the energy-density of the noise in test images as $\hat \gamma^2$ 
and consider three levels of noise, i.e., $\hat \gamma=\nicefrac{25}{255}$, $\nicefrac{15}{255}$, and $\nicefrac{10}{255}$.
For Gaussian noise removal, we add white Gaussian noise with $\sigma=\hat \gamma$ to clean images. 
For Uniform noise removal, we generate noise from $\mcal U(-\sqrt{3}\hat \gamma,\sqrt{3}\hat \gamma)$.
For denoising adversarial noisy images, the norm budgets of adversarial perturbation are set to be $\epsilon=\nicefrac{5}{255}\cdot \sqrt{m}$ and $\nicefrac{7}{255}\cdot\sqrt{m}$ respectively, where $m$ equals the size of test images.
We perturb noisy observations whose noise are generated from 
$\mcal N(0,\hat \gamma-\nicefrac{\epsilon}{\sqrt{m}})$, 
so that the $\ell_2$-norms of total noise in adversarial images are still bounded by $\hat \gamma \cdot \sqrt{m}$ and the energy-density thus are bounded by $\hat \gamma^2$ . We use Atk-$\nicefrac{\epsilon}{\sqrt{m}}$ to denote the adversarially perturbed noisy images in the size of $m$ with adversarial budget $\epsilon$. The number of iterations of PGD in ObsAtk is set to be five. 

\begin{table}[h]
    \centering
    \caption[The average PSNR (in dB) results of DnCNN denoisers on the gray-scale and RGB datasets. ]{The average PSNR (in dB) results of DnCNN denoisers on the gray-scale and RGB datasets. 
    Four types of noise are used for evaluation, viz. Gaussian $\mcal N$ and Uniform $\mcal U$ random noise, and ObsAtk with two different adversarial budgets. The energy-density of noise is bounded by $\hat \gamma^2$.}
    \scalebox{0.9}{
    \begin{tabular}{cccccc}
    \toprule
    {Dataset} & $\hat \gamma$ & $\mcal N$ & $\mcal U$ & Atk-\nicefrac{5}{255} & Atk-\nicefrac{7}{255} \\
    \hline
	\multirow{3}{*}{\small Set68}
	& \nicefrac{25}{255} &   29.16/0.02 & 29.15/0.01 & 24.26/0.12 & 23.12/0.10  \\
    & \nicefrac{15}{255} &   31.68/0.00 & 31.68/0/00 & 26.66/0.04 & 26.08/0.02  \\
    \midrule
	\multirow{3}{*}{\small Set12}
	& \nicefrac{25}{255} &   30.39/0.01 & 30.41/0.01 & 24.32/0.18 & 22.96/0.13  \\
    & \nicefrac{15}{255} &   32.78/0.00 & 32.81/0.00 & 26.91/0.05 & 26.25/0.01  \\
	\midrule
	\multirow{3}{*}{\small BSD68}
	& \nicefrac{25}{255} &   31.25/0.11 & 31.17/0.11 & 27.44/0.08 & 26.08/0.06  \\
    & \nicefrac{15}{255} &   33.98/0.11 & 33.93/0.12 & 29.31/0.08 & 27.84/0.04  \\
    \midrule
	\multirow{3}{*}{\small Kodak24}
	& \nicefrac{25}{255} &   32.20/0.13 & 32.13/0.14 & 27.87/0.08 & 26.37/0.07  \\
    & \nicefrac{15}{255} &   34.77/0.13 & 34.73/0.14 & 29.55/0.07 & 28.00/0.04  \\
    \bottomrule
    \end{tabular}
    }
    \label{tab:attack-gray}
\end{table}{}

From Tables \ref{tab:attack-gray}, we observe that ObsAtk clearly degrades the reconstruction performance of DIDs. In comparison to Gaussian or Uniform noisy images with the same noise levels, the recovered results from adversarial images are much worse in the sense of the PSNR. For example, when removing $\hat \gamma=\nicefrac{15}{255}$ noisy images in Set12, the average PSNR of reconstructions from Gaussian noise can achieve 32.78 dB, whereas the PSNR drops to 26.25 dB when dealing with Atk-$\nicefrac{7}{255}$ adversarial images. We observe the consistent phenomenon that a normally-trained denoiser $f^{\gamma}_{\theta}(\cdot)$ cannot effectively remove adversarial noise from visual results in Figure~\ref{fig:adv-noise}. 

\section{Robust and Generalizable Denoising via HAT}
\label{sec:hat}
The previous section shows that existing deep denoisers are vulnerable to adversarial perturbations. 
To improve the adversarial robustness of deep denoisers, we propose an adversarial training method, hybrid adversarial training (HAT), that uses original noisy images and their adversarial versions for training. 
Furthermore, we build a connection between the adversarial robustness of deep denoisers and their generalization capability to unseen types of noise. We show that HAT-trained denoisers can effectively remove real-world noise without the need to leverage the real-world noisy data.
\subsection{Hybrid Adversarial Training}
\label{sec:hat-method}
AT has been proved to be a successful and universally applicable technique for robustifying deep neural networks. Most variants of AT are developed for the classification task specifically, such as TRADES \citep{zhang_theoretically_2019} and GAIRAT \citep{zhang_geometry-aware_2020}. Here, we propose an AT strategy, HAT, for robust image denoising:
\begin{align}
	\min_{\theta\in \Theta} \E_{\rvx\sim P} & \E_{Q\sim \mcal U(\sQ^\gamma)}  \E_{\rvv\sim Q}  \frac{1}{2} \Big(\frac{1}{1+\alpha}\|f_{\theta}^{\gamma}(\rvy)-\rvx \|_2^2 \nonumber \\
	& + \frac{\alpha}{1+\alpha}\|f_{\theta}^{\gamma}(\rvy)-f_{\theta}^{\gamma}(\rvy') \|_2^2 \Big), 
	\label{eq:hybrid-at}
\end{align}
where $\rvy = \rvx+\rvv$ and $\rvy'=\rvy+\bm{\delta}^*$. Note that $\bm{\delta}^*$ is the adversarial perturbation obtained by solving ObsAtk in Eq.~\eqref{eq:obsattack}.

As shown in Eq.~\eqref{eq:hybrid-at}, the loss function consists of two terms. The first term measures the distance between ground-truth images $\vx$ and reconstructions $f_{\theta}^{\gamma}(\vy)$ from non-adversarial noisy images $\vy$, where $\vy$ contains noise $\vv$ sampled from a certain common distribution $Q$, such as Gaussian. This term encourages a good reconstruction performance of $f_{\theta}^{\gamma}$ from common distributions. The second term is the distance between $f_{\theta}^{\gamma}(\vy)$ and the reconstruction $f_{\theta}^{\gamma}(\vy')$ from the adversarially perturbed version $\vy'$ of $\vy$. This term ensures that the reconstructions from any two noisy observations within a small neighborhood of $\vy$ have similar image qualities. Minimizing these two terms at the same time controls the worst-case reconstruction performance $\|f_{\theta}^{\gamma}(\vy')-\vx\|$. 

The coefficient $\alpha$ balances the trade-off between reconstruction from common noise and the local continuity of $f_{\theta}^{\gamma}$. When $\alpha$ equals zero, HAT degenerates to normal training on common noise. The obtained denoisers fail to resist adversarial perturbations as shown in Section \ref{sec:obsattack}. When $\alpha$ is very large, the optimization gradually ignores the first term and completely aims for local smoothness. This may yield a trivial solution that $f_{\theta}^{\gamma}$ always outputs a constant vector for any input. A proper value of $\alpha$ thus ensures a denoiser that performs well for common noise and the worst-case adversarial perturbations simultaneously. We perform an ablation study on the effect of $\alpha$ for the robustness enhancement and unseen noise removal in Appendix \ref{sec:apdx_ablation}.

To train a denoiser applicable to different levels of noise with an energy-density bounded by $\gamma^2$, we randomly select a noise distribution $Q$ from a family of common distributions $\sQ^{\gamma}$. $\sQ^{\gamma}$ includes a variety of zero-mean distributions whose variance are bounded by $\gamma^2$. For example, we define 
$\sQ^{\gamma}_{\mcal N} = \{ \mcal N(\bm{0}, \sigma^2 \mathbf{I}))|\sigma \sim\mcal U(0, \gamma) \}$ for the experiments in the remaining of this paper.

\subsection{Robustness Enhancement via HAT}
\label{sec:hat_for_robustness}

\def \SubFigWidth {0.25} 
\def \SubImgWidth {.95}
\begin{figure}[t!]
    \centering
    \begin{subfigure}{\SubFigWidth\linewidth}
        \centering
        \includegraphics[width=\SubImgWidth \linewidth]{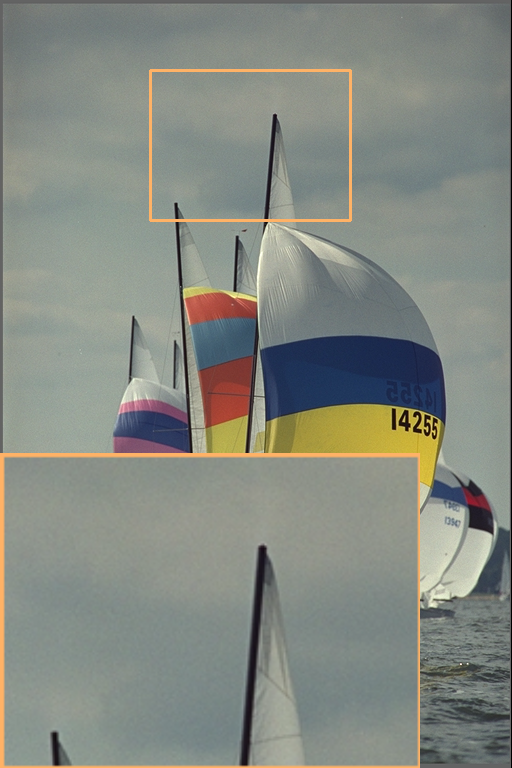}
        \caption{ Ground-truth}
    \end{subfigure}
    \begin{subfigure}{\SubFigWidth\linewidth}
        \centering
        \includegraphics[width=\SubImgWidth \linewidth]{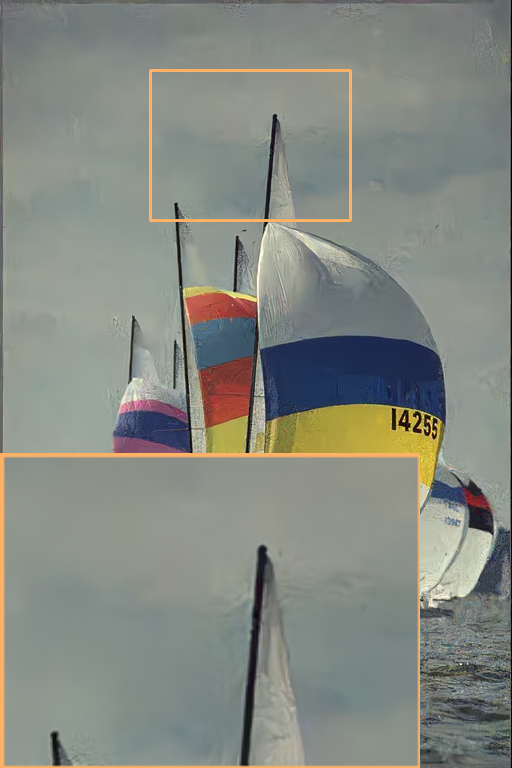}
        \caption{ NT}
    \end{subfigure}
    \begin{subfigure}{\SubFigWidth\linewidth}
        \centering
        \includegraphics[width=\SubImgWidth \linewidth]{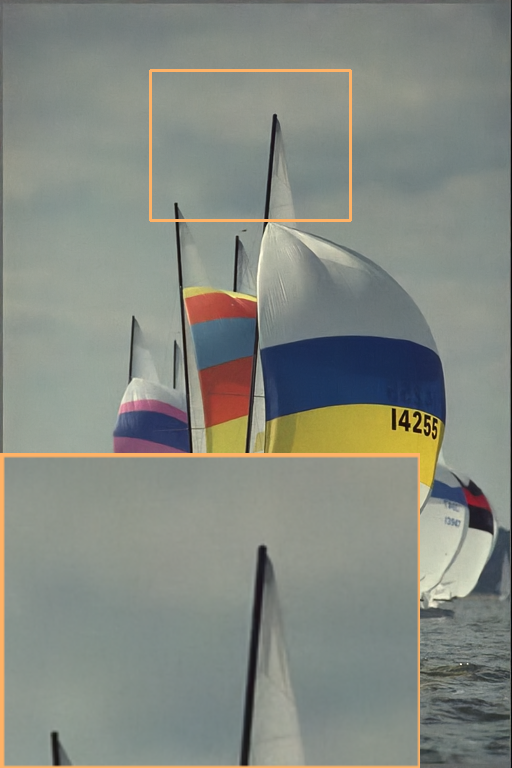}
        \caption{ HAT}
    \end{subfigure}
	\caption[Robustness comparison between normally and HAT-trained denoisers.]{
		From left to right are the ground-truth, the reconstruction of a normally-trained denoiser against attack, and the reconstruction of a HAT-trained denoiser against attack.}
	\label{fig:hat-defense}
\end{figure}

We follow the same settings as those in Section \ref{sec:obsattack} for training and evaluating $\gamma$-deep denoisers. The highest level of noise used for training is set to be $\gamma=\nicefrac{25}{255}$. Noise is sampled from a set of Gaussian distributions $\sQ^{\gamma}_{\mcal N}$.
We train deep denoisers with the HAT strategy and set $\alpha$ to be $1$, and use one-step Atk-$\nicefrac{5}{255}$ to generate adversarially noisy images for training. We compare HAT with normal training (NT) and the vanilla adversarial training (vAT) used in Choi \etal \citet{choi_deep_2021} that trains denoisers only with adversarial data. The results on Set68 and BSD68 are provided in this section. More results on Set12 and Kodak24 (in Tables \ref{tab:hat-set12} and \ref{tab:hat-kodak24}) are provided in Appendix \ref{sec:apdx_robustness_enhancement}. 

\begin{table}[h!]
    \centering
    \caption[Robustness comparison between normally and adversarially-trained DnCNN-B denoisers on the gray-scale Set68 dataset.]{The average PSNR (in dB) results of DnCNN-B denoisers on the gray-scale Set68 dataset. NT and HAT are compared in terms of the noise removal of Gaussian noise and adversarial noise. We repeat the training for three times and report the mean and standard deviation (mean/std).}
    \scalebox{0.9}{
    \begin{tabular}{ccccccc}
    \toprule
    Method & $\hat \gamma$ & $\mcal N$ & Atk-\nicefrac{3}{255} & Atk-\nicefrac{5}{255} & Atk-\nicefrac{7}{255} \\
    \hline
	\multirow{3}{*}{NT}
	& \nicefrac{25}{255} &   \thl{29.16}/0.02 & 26.20/0.07 & 24.26/0.12 & 23.12/0.10  \\
    & \nicefrac{15}{255} &   \thl{31.68}/0.00 & 27.98/0.05 & 26.66/0.04 & 26.08/0.02  \\
	\midrule
	\multirow{3}{*}{vAT}
	& \nicefrac{25}{255} &   29.05/0.07 & 27.02/0.15 & 25.51/0.32 & 24.34/0.34 \\
    & \nicefrac{15}{255} &   31.53/0.09 & 28.74/0.16 & 27.43/0.19 & 26.68/0.15 \\
	\midrule
	\multirow{3}{*}{HAT}
	& \nicefrac{25}{255} &   {28.88}/0.04 & \thl{27.48}/0.10 & \thl{26.40}/0.16 & \thl{25.32}/0.17 \\
    & \nicefrac{15}{255} &   {31.36}/0.03 & \thl{29.52}/0.01 & \thl{28.34}/0.03 & \thl{27.34}/0.03 \\
    \bottomrule
    \end{tabular}
    }
    \label{tab:hat-set68}
\end{table}{}

From Tables \ref{tab:hat-set68} and \ref{tab:hat-bsd68}, we observe that HAT obviously improves the reconstruction performance from adversarial noise in comparison to normal training. For example, on the Set68 dataset (Table \ref{tab:hat-set68}), when dealing with $\nicefrac{15}{255}$-level noise, the normally-trained denoiser achieves 31.68 dB for Gaussian noise removal, but the PSNR drops to 26.08 dB against Atk-\nicefrac{7}{255}. In contrast, the HAT-trained denoiser achieves a PSNR of 27.34 dB (1.26 dB higher) against Atk-\nicefrac{7}{255} and maintains a PSNR of 31.36 dB for Gaussian noise removal. In Figure~\ref{fig:hat-defense}, we can see that when dealing with adversarially noisy images, the HAT-trained denoiser can recover high-quality images while the normally-trained denoiser preserves noise patterns in the output. Besides, we observe that, similar to image classification tasks \citep{zhang_theoretically_2019},  AT-based methods (HAT and vAT) robustify deep denoisers at the expense of the performance on non-adversarial data (Gaussian denoising). Nevertheless, the degraded reconstructions are still reasonably good  in terms of the PSNR. 

\begin{table}[h!]
    \centering
    \caption[Robustness comparison between normally and adversarially-trained DnCNN-B denoisers on the RGB BSD68  dataset.]{The average PSNR (in dB) results of DnCNN-C denoisers on the RGB BSD68 dataset.}
    \scalebox{0.9}{
    \begin{tabular}{ccccccc}
    \toprule
    Method & $\hat \gamma$ & $\mcal N$ & Atk-\nicefrac{3}{255} & Atk-\nicefrac{5}{255} & Atk-\nicefrac{7}{255} \\
    \hline
	\multirow{3}{*}{NT}
	& \nicefrac{25}{255} &   \thl{31.25}/0.11 & 28.93/0.08 & 27.44/0.08 & 26.08/0.06  \\
    & \nicefrac{15}{255} &   \thl{33.98}/0.11 & 31.09/0.10 & 29.31/0.08 & 27.84/0.04  \\
	\midrule
	\multirow{3}{*}{vAT}
	& \nicefrac{25}{255} &   30.64/0.02 & 28.81/0.03 & 27.67/0.01 & 26.64/0.03 \\
    & \nicefrac{15}{255} &   33.45/0.06 & 31.10/0.05 & 29.79/0.02 & 28.63/0.08 \\
    \midrule
	\multirow{3}{*}{HAT}
	& \nicefrac{25}{255} &   {30.98}/0.03 & \thl{29.18}/0.03  & \thl{28.02}/0.02 & \thl{26.93}/0.04 \\
    & \nicefrac{15}{255} &   {33.67}/0.04 & \thl{31.38}/0.04  & \thl{30.03}/0.02 & \thl{28.80}/0.01 \\
    \bottomrule
    \end{tabular}
    }
    \label{tab:hat-bsd68}
\end{table}{}

\begin{table}[t!]
    \centering
    \caption[Comparison of different methods for denoising real-world noisy images in terms of PSNR (dB).]{Comparison of different methods for denoising real-world noisy images in terms of PSNR (dB). We repeat the experiments of each denoising method for three times and report the  mean/standard deviation of PSNR values.
    }
    \scalebox{0.8}{
    \begin{tabular}{cccccccc}
    \toprule
    Dataset &  BM3D & DIP & N2S(1) & NT & vAT & HAT & N2C\\
    \hline
    PolyU 	        & 37.40/0.00 & 36.08/0.01 & 35.37/0.15 & 35.86/0.01 & 36.77/0.00 & \thl{37.82}/0.04 & -- / --    \\
    CC 		        & 35.19/0.00 & 34.64/0.06 & 34.33/0.14 & 33.56/0.01 & 34.49/0.10 & \thl{36.26}/0.06 & -- / --  \\
    SIDD 	        & 25.65/0.00 & 26.89/0.02 & 26.51/0.03 & 27.20/0.70 & 27.08/0.28 & \thl{33.44}/0.02 & \tHL{33.50}/0.03  \\
    \bottomrule
    \end{tabular}
    }
    \label{tab:real-denoising}
\end{table}

\subsection{Robustness Benefits Generalization to Unseen Noise}
\label{sec:hat_for_unseen}

It has been shown that denoisers that are normally trained on common synthetic noise fail to remove real-world noise induced by standard imaging procedures \citep{xu_multi-channel_2017,abdelhamed_high-quality_2018}. 
To train denoisers that can handle real-world noise, researchers have proposed several methods which can be roughly divided into two categories, namely \tit{dataset-based} denoising methods and \tit{single-image-based} denoising methods. High-performance dataset-based methods require a set of real noisy data for training, e.g., CBDNet requiring pairs of clean and noisy images \citep{guo_toward_2019} and Noise2Noise requiring multiple noisy observations of every single image \citep{lehtinen_noise2noise_2018}. 
However, a large number of paired data are not available in some applications, such as medical radiology and high-speed photography. 
To address this, single-image-based methods are proposed to remove noise by exploiting the correlation between signals across pixels and the independence between noise. This category of methods, such as DIP \citep{ulyanov_deep_2018} and N2S \citep{batson_noise2self_2019}, are adapted to various types of signal-independent noise, but they optimize the deep denoiser on each test image. The test-time optimization is extremely time-consuming, e.g., N2S needs to update a denoiser for \tit{thousands of iterations} to achieve good reconstruction performance. 

Here, we point out that HAT is a promising framework to train a generalizable deep denoiser \tit{only with synthetic noise}. The resultant denoiser can be directly applied to perform  denoising for unseen noisy images in real-time. During training, HAT first samples noise from common distributions (Gaussian) with noise levels from low to high. ObsAtk then explores the $\epsilon$-neighborhood for each noisy image to search for a particular type of noise that degrades the denoiser the most. By ensuring the denoising performance of the worst-case noise, the resultant denoiser can deal with other unknown types of noise within the $\epsilon$-neighborhood as well. To train a robust denoiser that generalizes well to real-world noise, we need to choose a proper adversarial budget $\epsilon$. When $\epsilon$ is very small and close to zero, the HAT reduces to normal training. When $\epsilon$ is very much larger than the norm of basic noise $\vv$, the adversarially noisy image may be visually unnatural because the adversarial perturbation $\bm{\delta}$ only satisfies the zero-mean constraint and is not guaranteed to be spatially uniformly distributed as other types of natural noise being. In practice, we set the value of $\epsilon$ of ObsAtk to be $\nicefrac{5}{255}\cdot \sqrt{m}$, where $m$ denotes the size of image patches. The value of $\alpha$ of HAT is kept unchanged as $2$.

\paragraph{Experimental Settings} We evaluate the generalization capability of HAT on several real-world noisy datasets, including PolyU \citep{xu_real-world_2018}, CC \citep{xu_multi-channel_2017}, and SIDD \citep{abdelhamed_high-quality_2018}. 
PolyU, CC, and SIDD contain RGB images of common scenes in daily life. These images are captured by different brands of digital cameras and smartphones, and they contain various levels of noise by adjusting the ISO values. For the PolyU and CC, we use the clean images in BSD500 for training an adversarially robust $\gamma$-denoiser with $\gamma=\nicefrac{25}{255}$. We sample Gaussian noise from a set of distributions $\sQ^{\gamma}_{\mcal N}$ and add the noise to clean images to craft noisy observations. HAT trains the denoiser jointly with Gaussian noisy images and their adversarial versions.
For the SIDD, we use clean images in the SIDD-small set for training and test the denoisers on the SIDD-val set. The highest level of noise used for HAT is set to be $\gamma=\nicefrac{50}{255}$. In each case, we only use clean images for training denoisers without the need of real noisy images

\paragraph{Results} We compare HAT-trained denoisers with the NT and vAT-trained ones. From Table \ref{tab:real-denoising}, we observe that HAT performs much better than both competitors.
For example, on the SIDD-val dataset, the HAT-trained denoiser achieves an average PSNR value of 33.44 dB that is 6.24 dB higher than the NT-trained one. We also compare HAT-trained denoisers with single-image-based methods, including DIP, N2S, and the classical BM3D \citep{dabov_image_2007}. For DIP and N2S,\footnote{The officially released codes of DIP and N2S are used here. 
} the numbers of iterations for each image are set to be 2,000 and 1,000, respectively. N2S works in two modes, namely single-image-based denoising and dataset-based denoising. Here, we use N2S in the single-image-based mode, denoted as N2S(1), due to the assumption that no real noisy data are available for training. We observe that HAT-trained denoisers consistently outperform these baselines. Visual comparisons are provided in Appendix \ref{sec:apdx_vis_real}. Besides, since the SIDD-small provides a set of real noisy and ground-truth pairs, we train a denoiser, denoted as Noise2Clean (N2C), with these paired data and use the N2C denoiser as the oracle for comparison. We observe that HAT-trained denoisers are comparable to the N2C one for denoising images in SIDD-val (a PSNR of 33.44dB vs 33.50dB).

\section{Chapter Summary}

Normally-trained deep denoisers are vulnerable to adversarial attacks. HAT can effectively robustify deep denoisers and boost their generalization capability to unseen real-world noise. In the future, we will extend the adversarial-training framework to other image restoration tasks, such as deblurring. We aim to develop a generic AT-based robust optimization framework to train deep models that can recover clean images from unseen types of degradation.


\section{Appendices}

\subsection{Two-step Projection}
\label{sec:thm1-proof}
\setcounter{theorem}{0}
\begin{theorem} 
	For any arbitrary vector $\vdelta \in \sR^m$, its projection onto the region defined by the intersection of the  norm-bounded and zero-mean constraints  is equivalent to the projection first onto the zero-mean hyperplane followed by the projection onto the $\epsilon$-ball ($\epsilon>0$), i.e., 
	\begin{equation}
		\proj_{A\cap B}(\vdelta) = \proj_B(\proj_A(\vdelta)),
		\label{eq:thm}
	\end{equation}
	where
	\begin{subequations}
		\begin{align}
			A & = \big\{\vz\in \sR^m  \,\big|\, \vn^{\top}\vz = 0 \big\}, \\
			B & =\big\{\vz\in \sR^m  \,\big|\, \|\vz\|_2^2 \leq \epsilon^2\big\},
		\end{align}
	\end{subequations}
	and  $\vn = [1,1, \ldots, 1]^\top$.
\end{theorem}

\begin{figure}[h!]
	\centering
	\includegraphics[width=.7\linewidth]{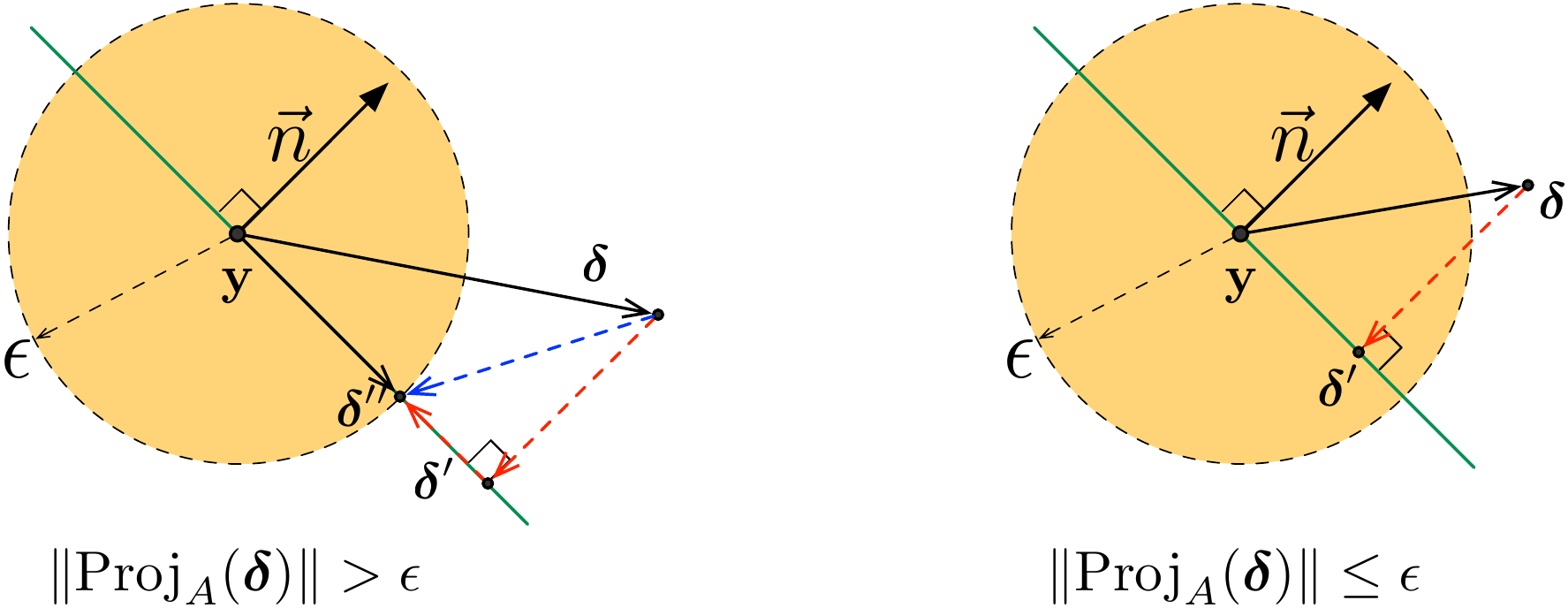}
	\caption[Illustration of Theorem \ref{sec:thm1-proof}.]{Illustration of Theorem \ref{sec:thm1-proof}. In the case of $\|\proj_A(\vdelta)\| > \epsilon$, the red-dot lines show that the perturbation $\vdelta$ is projected onto the region defined by the zero-mean and $\epsilon$-ball constraints sequentially. The blue-dot line shows the exact projection of $\vdelta$ on to $A\cap B$.}
\end{figure}

\paragraph{Proof} 
Let us consider the RHS of Eq. \eqref{eq:thm} first. It is easy to derive the projections onto $A$ and $B$ seperately: 
\begin{subequations}
	\begin{align}
		\proj_A(\vdelta) & = \vdelta - \frac{\vn^\top \vdelta }{\|{\vn}\|^2_2} {\vn}, \label{eq:apdx-proj-zero-mean} \\
		\proj_B(\vdelta) & = \min\Big(\frac{\epsilon}{\|\vdelta\|_2}, 1\Big) \text{~} \vdelta. \label{eq:apdx-proj-norm}
	\end{align}
	\label{eq:apdx-two-step-projection}
\end{subequations}
Thus, we have
\begin{equation}
	\proj_B(\proj_A(\vdelta)) =
	\begin{cases}
		\proj_A(\vdelta), 
		& \text{if $\|\proj_A(\vdelta)\| \leq \epsilon$};\\
		\frac{\epsilon}{\|\proj_A(\vdelta)\|_2} \proj_A(\vdelta), 
		& \text{if $\|\proj_A(\vdelta)\| > \epsilon$.}
	\end{cases}
	\label{eq:two-step-projection-sol}   
\end{equation}

Now let us consider the LHS of Eq.~\eqref{eq:thm}. The projection onto $A\cap B$ can be formulated as the solution of the following convex optimization problem:
\begin{equation}
	\min_{\vz} \frac{1}{2}\|\vdelta -\vz\|^2_2, \quad
	\text{s.t.} \quad \vn^{\top} \vz = 0, \quad
	\|\vz\|^2_2 \leq \epsilon^2,
	\label{eq:proj-jointly-opt}
\end{equation}
where $\vz \in \sR^m$. We can write the Lagrangian, $L:\sR^m \times \sR \times \sR \to \sR$, associated with the problem \eqref{eq:proj-jointly-opt} as 
\begin{equation}
	L(\vz, \lambda, \nu) = \frac{1}{2}\|\vdelta -\vz\|^2_2 + \lambda (\|\vz^*\|^2_2 - \epsilon^2) + \nu \vn^{\top} \vz.
\end{equation}
Since there exists an $\vz \in \sR^m$, e.g., $\vz=[0,\ldots,0]^{\top} \in \sR^m$, such that $\vn^{\top}\vz=0$ and $\|\vz\|^2_2 < \epsilon^2$, the problem \eqref{eq:proj-jointly-opt} is strictly feasible, i.e., it satisfies the Slater's condition \cite{boyd2004convex}. Besides, the objective and the constraints are all differentiable, thus the KKT conditions in Eq.~\eqref{eq:proj-jointly-opt-kkt} provide necessary and sufficient conditions for optimality.
\begin{subequations}
	\begin{align}
		\|\vz^*\|^2_2 - \epsilon^2 \leq 0, \\
		\vn^{\top}\vz^* = 0, \\
		\lambda \geq 0,\\
		\lambda (\|\vz^*\|^2_2 - \epsilon^2) = 0, \\
		\frac{\partial L}{\partial \vz} = (1+2\lambda) \vz^* - \vdelta + \nu \vn= 0.
	\end{align}
	\label{eq:proj-jointly-opt-kkt}
\end{subequations}

We obtain the optimal solution by considering the following two cases separately, i.e., $\|\proj_A(\vdelta)\| \leq \epsilon$ and $\|\proj_A(\vdelta)\| > \epsilon$. 

~\\
\textbf{Case-(1)}: $\|\proj_A(\vdelta)\| > \epsilon$. \\
If $\lambda>0$, then Eq. \eqref{eq:proj-jointly-opt-kkt} reduces to the following equation:
\begin{subequations}
	\begin{align}
		\vn^{\top}\rz^* = 0, \\
		\|\vz^*\|^2_2 - \epsilon^2 = 0, \\
		(1+2\lambda) \vz^* - \vdelta + \nu \vn= 0.
	\end{align}
\end{subequations}
We can easily solve these equations and obtain that
\begin{subequations}
	\begin{align}
		(1+2\lambda) &= \proj_A(\vdelta) / \epsilon, \\
		\nu &= \vn^{\top}\vdelta / m, \\
		\vz^* &= \frac{\epsilon}{\|\proj_A(\vdelta)\|_2} \proj_A(\vdelta). 
	\end{align}
\end{subequations}
If $\lambda=0$, then Eq. \eqref{eq:proj-jointly-opt-kkt} reduces to the following  set of equations:
\begin{subequations}
	\begin{align}
		\|\vz^*\|^2_2 - \epsilon^2 &\leq 0, \\
		\vn^{\top}\rz^* &= 0 \label{eqn:bottom2a}\\
		\vz^* - \vdelta + \nu \vn&= 0.  \label{eqn:bottom2b}
	\end{align}
\end{subequations}
According to \eqref{eqn:bottom2a} and \eqref{eqn:bottom2b}, we obtain that $\vz^* = \vdelta - \frac{\vn^{\top}\vdelta}{m}\vn = \proj_A(\vdelta) $ with a norm strictly larger than $\epsilon$, which contradicts the constraint $\|\vz^*\|^2_2 \leq \epsilon^2$. Thus, for the case of  $\|\proj_A(\vdelta)\| > \epsilon$, we have that $\vz^* = \frac{\epsilon}{\|\proj_A(\vdelta)\|_2} \proj_A(\vdelta)$ which is equal to $\proj_B(\proj_A(\vdelta))$ in Eq. \eqref{eq:two-step-projection-sol}.

~\\
\textbf{Case-(2)}: $\|\proj_A(\vdelta)\| \leq \epsilon$. \\ 
Since $\|\proj_A(\vdelta)\| \leq \epsilon$ and $\|\proj_A(\vdelta)\| \in A$, we have $\proj_A(\vdelta) \in A\cap B$. For any other point $\vz' \in A\cap B$ and $\vz' \neq \proj_A(\vdelta)$, we have $\|\vdelta-\vz'\| > \|\vdelta-\proj_A(\vdelta)\| $, where the strict inequality holds because $A$ is the set of points from a hyperplane. Thus, $\vz' $ is not the $ \proj_{A\cap B}(\vdelta)$. Therefore, $\proj_{A\cap B}(\vdelta) = \proj_A(\vdelta) = \proj_B(\proj_A(\vdelta))$.

~\\
In summary, we show that $\proj_{A\cap B}(\vdelta) = \proj_B(\proj_A(\vdelta))$ for any arbitrary $\vdelta \in \sR^m$.

\subsection{Experiments of Robustness Enhancement on Set12 and Kodak24}
We compare the robustness of deep denoisers trained via three strategies, i.e., NT, vAT and HAT. The results on Set 12 and Kodak24 are provided in Table \ref{tab:hat-set12} and Table \ref{tab:hat-kodak24} respectively.
We observe that HAT can effectively robustify deep denoisers. The reconstruction quality of HAT-trained denoisers from adversarially noisy images is clearly better than that of the NT and vAT-trained ones.

\label{sec:apdx_robustness_enhancement}
\begin{table}[h!]
	\centering
	\caption{The average PSNR (in dB) results of DnCNN-B denoisers on the gray-scale Set12 dataset. }
	\scalebox{.9}{
		\begin{tabular}{ccccccc}
			\toprule
			Training & $\hat \gamma$ & $\mcal N$ & Atk-\nicefrac{3}{255} & Atk-\nicefrac{5}{255} & Atk-\nicefrac{7}{255} \\
			\hline
			\multirow{3}{*}{NT}
			& \nicefrac{25}{255} &   \thl{30.39}/0.01 & 26.51/0.14 & 24.32/0.18 & 22.96/0.13  \\
			& \nicefrac{15}{255} &   \thl{32.78}/0.00 & 28.50/0.08 & 26.91/0.05 & 26.25/0.01  \\
			\midrule
			\multirow{3}{*}{vAT}
			& \nicefrac{25}{255} &   30.25/0.08 & 27.56/0.06 & 25.82/0.04 & 24.33/0.04 \\
			& \nicefrac{15}{255} &   32.63/0.09 & 29.37/0.17 & 27.83/0.15 & 26.91/0.08 \\
			\midrule
			\multirow{3}{*}{HAT}
			& \nicefrac{25}{255} &   30.01/0.06 & \thl{27.96}/0.15 & \thl{26.46}/0.20 & \thl{25.13}/0.19 \\
			& \nicefrac{15}{255} &   32.47/0.04 & \thl{29.95}/0.03 & \thl{28.45}/0.04 & \thl{27.20}/0.03 \\
			\bottomrule
		\end{tabular}
	}
	\label{tab:hat-set12}
\end{table}{}

\begin{table}[h!]
	\centering
	\caption{The average PSNR (in dB) results of DnCNN-C denoisers on the RGB Kodak24 dataset.}
	\scalebox{.9}{
		\begin{tabular}{ccccccc}
			\toprule
			Training & $\hat \gamma$ & $\mcal N$ & Atk-\nicefrac{3}{255} & Atk-\nicefrac{5}{255} & Atk-\nicefrac{7}{255} \\
			\hline
			\multirow{3}{*}{NT}
			& \nicefrac{25}{255} &   \thl{32.20}/0.13 & 29.57/0.09 & 27.87/0.08 & 26.37/0.07  \\
			& \nicefrac{15}{255} &   \thl{34.77}/0.13 & 31.54/0.11 & 29.55/0.07 & 28.00/0.04  \\
			\midrule
			\multirow{3}{*}{vAT}
			& \nicefrac{25}{255} &   31.44/0.01 & 29.41/0.05 & 28.13/0.06 & 26.98/0.02 \\
			& \nicefrac{15}{255} &   34.14/0.08 & 31.53/0.11 & 30.06/0.08 & 28.78/0.06 \\
			\midrule
			\multirow{3}{*}{HAT}
			& \nicefrac{25}{255} &   31.83/0.04 & \thl{29.85}/0.02  & \thl{28.56}/0.02 & \thl{27.34}/0.05 \\
			& \nicefrac{15}{255} &   34.36/0.06 & \thl{31.84}/0.05  & \thl{30.37}/0.02 & \thl{29.05}/0.01 \\
			\bottomrule
		\end{tabular}
	}
	\label{tab:hat-kodak24}
\end{table}{}

\vfill
\subsection{Ablation study}
\label{sec:apdx_ablation}

\subsubsection{Effect of $\alpha$ on Robustness Enhancement and Generalization to Real-world noise}

Here, we evaluate the effect of $\alpha$ in HAT on the adversarial robustness and the generalization capability to real-world noise. We train deep denoisers on the RGB BSD500 (except 68 images for test) dataset. The obtained denoisers are tested on the BSD68 dataset for Gaussian and adversarial noise removal. The generalization capability is evaluated on two datasets of real-world noisy images, i.e., PolyU and CC. Experimental settings follow those in Section \ref{sec:hat_for_robustness}.

\begin{figure}[h!]
	\centering
	\includegraphics[width=.7\linewidth]{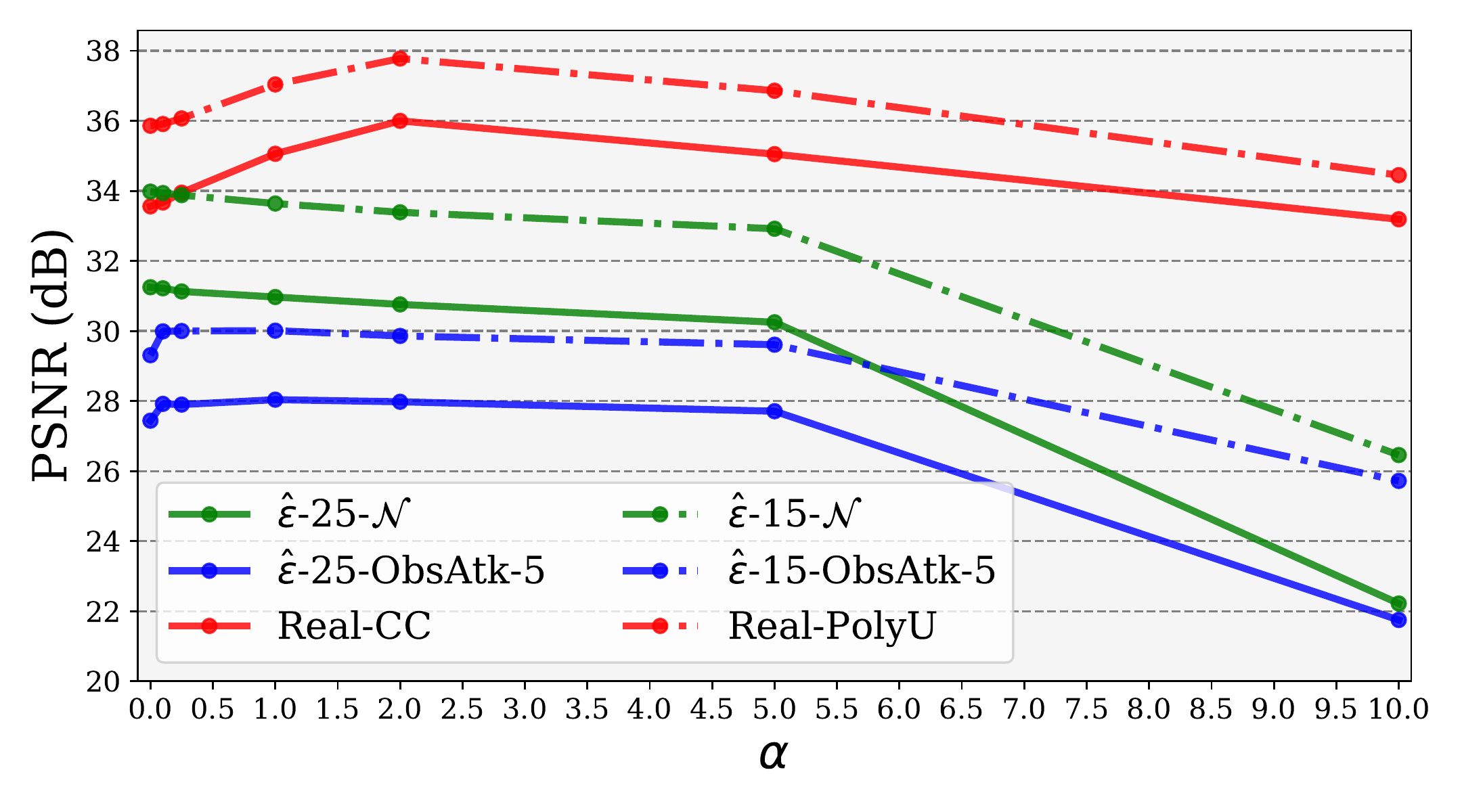}
	\caption[Ablation study on the effect of $\alpha$ in HAT.]{Ablation study on the effect of $\alpha$ in HAT. \textcolor{Green}{Green lines} show the denoising results on non-adversarial noise sampled from common distributions. The legend $\hat \gamma$-w-$\mcal N$ denotes the Gaussian noise ($\sigma=\nicefrac{w}{255}$) with a energy-density bounded by $\hat \gamma^2=\nicefrac{w^2}{255^2}$. \textcolor{Blue}{Blue lines} show that denoising results on adversarially perturbed noisy images. $\hat \gamma$-w-ObsAtk-5 denotes the adversarial noise crafted by ObsAtk-5 with a energy-density bounded by $\hat \gamma^2$. \textcolor{Red}{Red lines} show the denoising results on real-world noisy images.}
	\label{fig:apdx-ablation-alpha}
\end{figure}

Figure \ref{fig:apdx-ablation-alpha} corroborates the analysis in Section \ref{sec:hat-method} that the coefficient $\alpha$ balances the trade-off between reconstruction from common noise and the adversarial robustness. We also find that the generalization capability to real-world noise is correlated to the adversarial robustness. Specifically, good adversarial robustness usually implies good generalization to real-world noise. In Figure \ref{fig:apdx-ablation-alpha}, the best robustness and the best performance on real-world noise appear around $\alpha=1$ or $2$. When $\alpha$ is too large or too small, the robustness and generalization worsen. For the noise sampled from Gaussian distributions, increasing $\alpha$ degrades the denoising performance. In summary, we set $\alpha$ to $1$ or $2$ to achieve a good balance between the denoising performance on common noise and the adversarial robustness as well as real-world generalization.

\begin{figure}[h!]
	\centering
	\includegraphics[width=.7\linewidth]{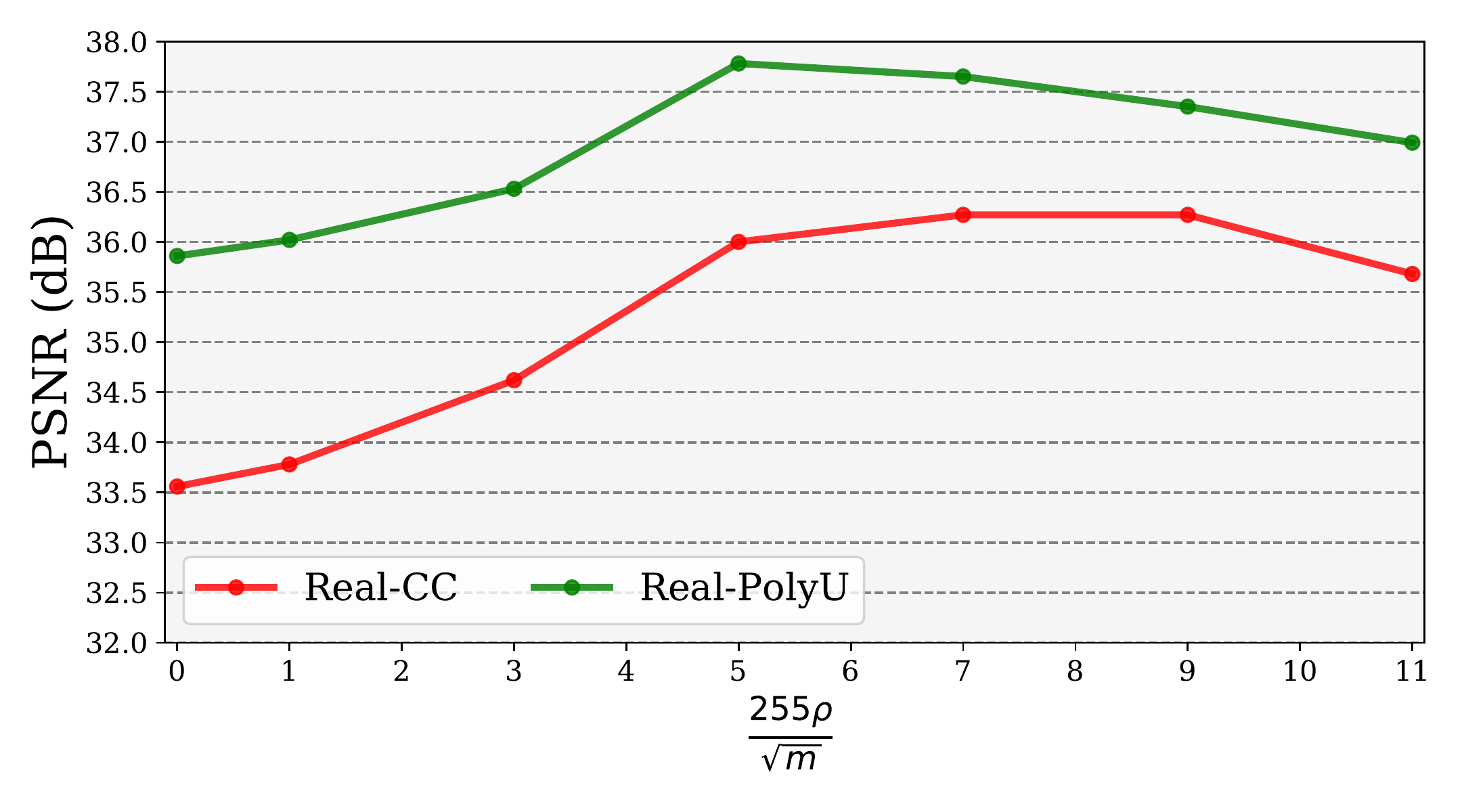}
	\caption{Ablation study on the effect of $\epsilon$ in HAT.}
	\label{fig:apdx-ablation-rho}
\end{figure}

\def \SubFigWidth {0.11} 
\def \SubImgWidth {.99}
\begin{figure}[h!]
	\centering
	\begin{subfigure}{\SubFigWidth\linewidth}
		\centering
		\includegraphics[width=\SubImgWidth \linewidth]{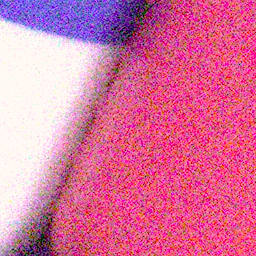}
	\end{subfigure}
	\begin{subfigure}{\SubFigWidth\linewidth}
		\centering
		\includegraphics[width=\SubImgWidth \linewidth]{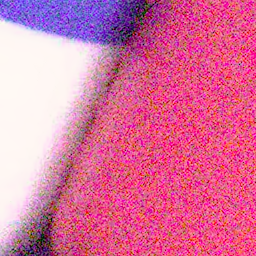}
	\end{subfigure}
	\begin{subfigure}{\SubFigWidth\linewidth}
		\centering
		\includegraphics[width=\SubImgWidth \linewidth]{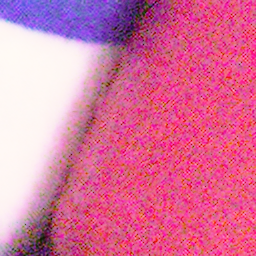}
	\end{subfigure}
	\begin{subfigure}{\SubFigWidth\linewidth}
		\centering
		\includegraphics[width=\SubImgWidth \linewidth]{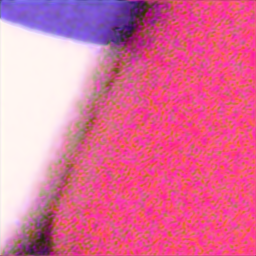}
	\end{subfigure}
	\begin{subfigure}{\SubFigWidth\linewidth}
		\centering
		\includegraphics[width=\SubImgWidth \linewidth]{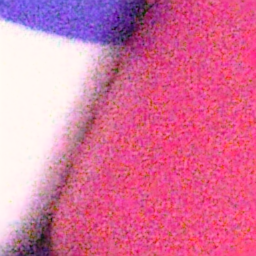}
	\end{subfigure}
	\begin{subfigure}{\SubFigWidth\linewidth}
		\centering
		\includegraphics[width=\SubImgWidth \linewidth]{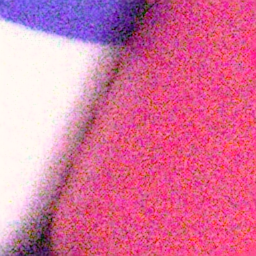}
	\end{subfigure}
	\begin{subfigure}{\SubFigWidth\linewidth}
		\centering
		\includegraphics[width=\SubImgWidth \linewidth]{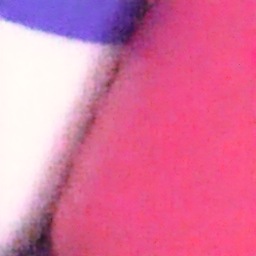}
	\end{subfigure}
	\begin{subfigure}{\SubFigWidth\linewidth}
		\centering
		\includegraphics[width=\SubImgWidth \linewidth]{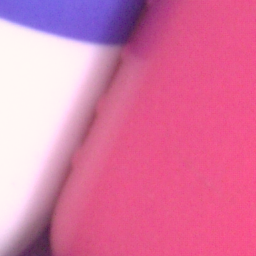}
	\end{subfigure}
	\vspace{.5em}
	
	\begin{subfigure}{\SubFigWidth\linewidth}
		\centering
		\includegraphics[width=\SubImgWidth \linewidth]{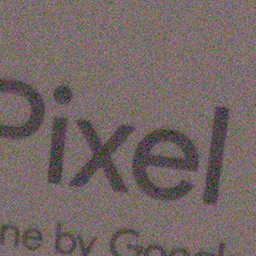}
	\end{subfigure}
	\begin{subfigure}{\SubFigWidth\linewidth}
		\centering
		\includegraphics[width=\SubImgWidth \linewidth]{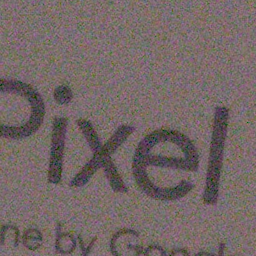}
	\end{subfigure}
	\begin{subfigure}{\SubFigWidth\linewidth}
		\centering
		\includegraphics[width=\SubImgWidth \linewidth]{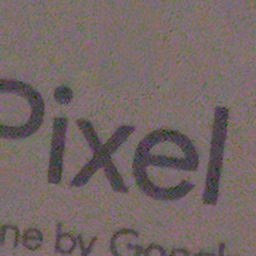}
	\end{subfigure}
	\begin{subfigure}{\SubFigWidth\linewidth}
		\centering
		\includegraphics[width=\SubImgWidth \linewidth]{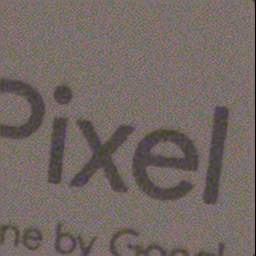}
	\end{subfigure}
	\begin{subfigure}{\SubFigWidth\linewidth}
		\centering
		\includegraphics[width=\SubImgWidth \linewidth]{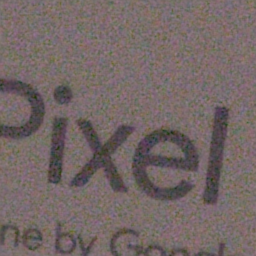}
	\end{subfigure}
	\begin{subfigure}{\SubFigWidth\linewidth}
		\centering
		\includegraphics[width=\SubImgWidth \linewidth]{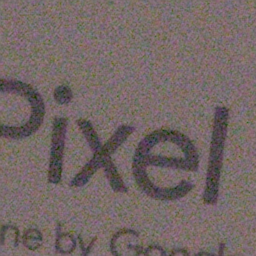}
	\end{subfigure}
	\begin{subfigure}{\SubFigWidth\linewidth}
		\centering
		\includegraphics[width=\SubImgWidth \linewidth]{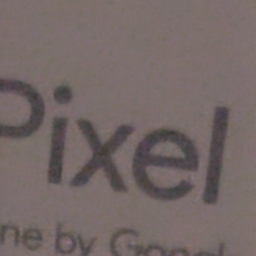}
	\end{subfigure}
	\begin{subfigure}{\SubFigWidth\linewidth}
		\centering
		\includegraphics[width=\SubImgWidth \linewidth]{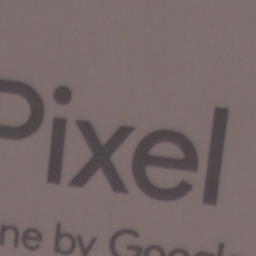}
	\end{subfigure}
	\vspace{.5em}
	
	\begin{subfigure}{\SubFigWidth\linewidth}
		\centering
		\includegraphics[width=\SubImgWidth \linewidth]{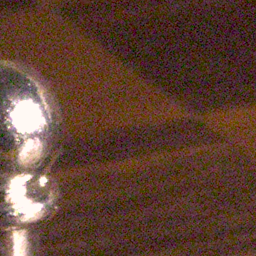}
	\end{subfigure}
	\begin{subfigure}{\SubFigWidth\linewidth}
		\centering
		\includegraphics[width=\SubImgWidth \linewidth]{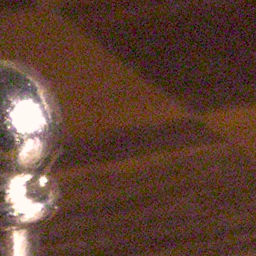}
	\end{subfigure}
	\begin{subfigure}{\SubFigWidth\linewidth}
		\centering
		\includegraphics[width=\SubImgWidth \linewidth]{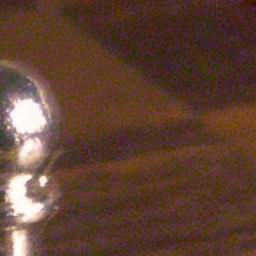}
	\end{subfigure}
	\begin{subfigure}{\SubFigWidth\linewidth}
		\centering
		\includegraphics[width=\SubImgWidth \linewidth]{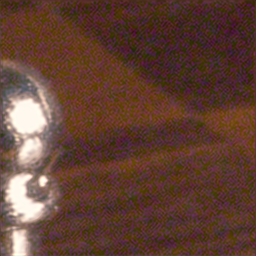}
	\end{subfigure}
	\begin{subfigure}{\SubFigWidth\linewidth}
		\centering
		\includegraphics[width=\SubImgWidth \linewidth]{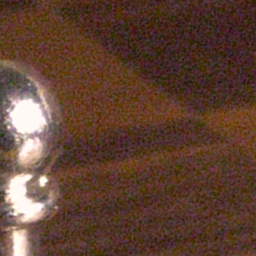}
	\end{subfigure}
	\begin{subfigure}{\SubFigWidth\linewidth}
		\centering
		\includegraphics[width=\SubImgWidth \linewidth]{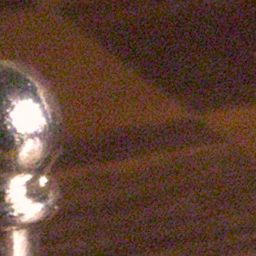}
	\end{subfigure}
	\begin{subfigure}{\SubFigWidth\linewidth}
		\centering
		\includegraphics[width=\SubImgWidth \linewidth]{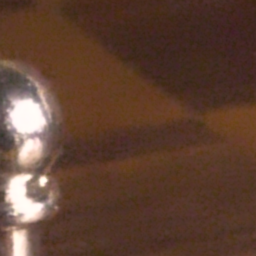}
	\end{subfigure}
	\begin{subfigure}{\SubFigWidth\linewidth}
		\centering
		\includegraphics[width=\SubImgWidth \linewidth]{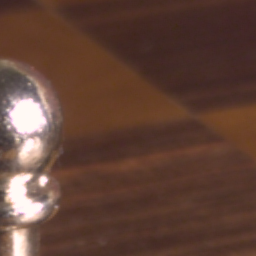}
	\end{subfigure}
	\vspace{.5em}
	
	\begin{subfigure}{\SubFigWidth\linewidth}
		\centering
		\includegraphics[width=\SubImgWidth \linewidth]{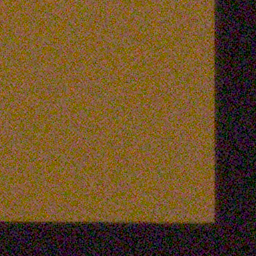}
		\caption{\scriptsize Noisy}
	\end{subfigure}
	\begin{subfigure}{\SubFigWidth\linewidth}
		\centering
		\includegraphics[width=\SubImgWidth \linewidth]{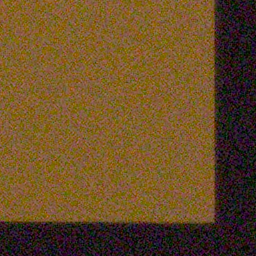}
		\caption{\scriptsize BM3D}
	\end{subfigure}
	\begin{subfigure}{\SubFigWidth\linewidth}
		\centering
		\includegraphics[width=\SubImgWidth \linewidth]{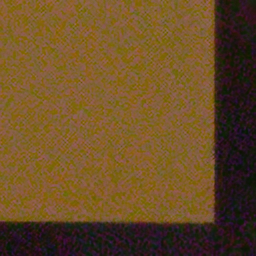}
		\caption{\scriptsize DIP}
	\end{subfigure}
	\begin{subfigure}{\SubFigWidth\linewidth}
		\centering
		\includegraphics[width=\SubImgWidth \linewidth]{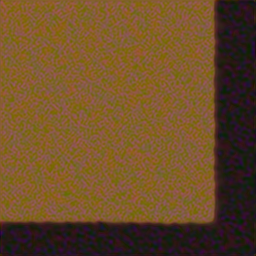}
		\caption{\scriptsize N2S}
	\end{subfigure}
	\begin{subfigure}{\SubFigWidth\linewidth}
		\centering
		\includegraphics[width=\SubImgWidth \linewidth]{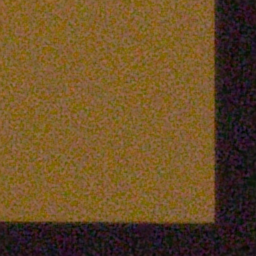}
		\caption{\scriptsize NT}
	\end{subfigure}
	\begin{subfigure}{\SubFigWidth\linewidth}
		\centering
		\includegraphics[width=\SubImgWidth \linewidth]{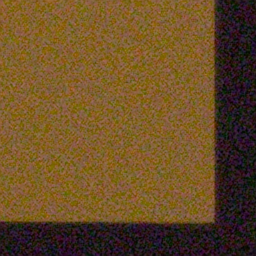}
		\caption{\scriptsize vAT}
	\end{subfigure}
	\begin{subfigure}{\SubFigWidth\linewidth}
		\centering
		\includegraphics[width=\SubImgWidth \linewidth]{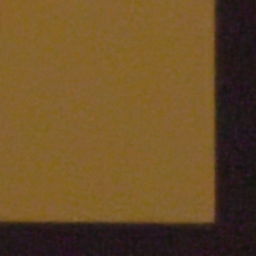}
		\caption{\scriptsize HAT}
	\end{subfigure}
	\begin{subfigure}{\SubFigWidth\linewidth}
		\centering
		\includegraphics[width=\SubImgWidth \linewidth]{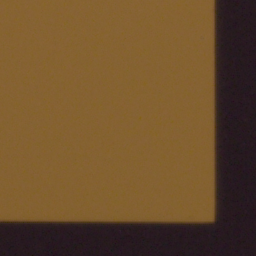}
		\caption{\scriptsize G.T.}
	\end{subfigure}
	\vspace{.1em}
	\caption[Comparison of different denoisers for denoising SIDD-val set.]{
		Comparison of different denoisers for denoising SIDD-val set. From left to right are the input noisy images, reconstructions of different denoisers, and the ground-truth (G.T.) images We can see that the HAT-trained denoiser performs the best in comparison to other baseline methods.
	}
	\label{fig:hat-sidd}
\end{figure}

\subsubsection{Effect of $\epsilon$ on Generalization to Real-world Noise}

Here, we evaluate the effect of $\epsilon$ used in HAT on the generalization capability to real-world noise. We train deep denoisers on the RGB BSD500 (except 68 images for test) dataset and evaluate the generalization capability on two real-world datasets, namely PolyU and CC. The $\alpha$ is set to be $2$. The adversarial budget $\epsilon$ of ObsAtk-$\nicefrac{\epsilon}{\sqrt{m}}$, that generates adversarially noisy images for HAT, is set to be values from $[0, \sqrt{m}, 3\sqrt{m},\ldots, 11\sqrt{m}]$ for comparison, where $m$ denotes the size of images. Other experimental settings follow those in Section \ref{sec:hat_for_robustness}.

Figure \ref{fig:apdx-ablation-rho} corroborates the analysis in Section \ref{sec:hat_for_unseen}.
When $\epsilon$ is very small and close to zero, the HAT reduces to normal training. The resultant denoisers cannot effectively remove real-world noise. When $\epsilon$ is very much larger than the norm of basic noise $\vv$, the statistics of adversarial noise may be very unnatural because the adversarial perturbation $\bm{\delta}$ might concentrate on a certain region, like edges or texture, and not be spatially uniformly distributed as other types of natural noise being. We can see that, when $\epsilon > \nicefrac{7}{255} \sqrt{m}$, the denoising performance on real-world datasets starts to decrease. In practice, we set the value of $\epsilon$ of ObsAtk to be $\nicefrac{5}{255}\cdot \sqrt{m}$ to train generalizable denoisers.

\subsection{Visual results of real-world noise removal}
\label{sec:apdx_vis_real}

We show the denoising results on SIDD-val set in Figure \ref{fig:hat-sidd}. We observe that HAT-trained denoiser can effectively remove the real-world noise while the normally-trained one retains much noise in the reconstructions. Besides, the HAT-trained denoiser outperforms other baseline methods and produces much cleaner results. 

%% file: Chapters/iclr2020.tex
\chapter{On Robustness of Neural Ordinary Differential Equations}
\label{chp:iclr2020}
\nocite{yan_robustness_2020}
 Neural ordinary differential equations (ODEs) have been attracting increasing attention in various research domains recently.  There have been some works studying optimization issues and  approximation capabilities of neural ODEs, but their robustness is still yet unclear. In this work, we fill this important gap by exploring robustness properties of neural ODEs both empirically and theoretically. We first present an empirical study on the robustness of the neural ODE-based networks (ODEnets) by exposing them to inputs with various types of perturbations and subsequently investigating the changes of the corresponding outputs. In contrast to conventional convolutional neural networks (CNNs), we find that the ODEnets are more robust against both random Gaussian perturbations and adversarial attack examples. We then provide an insightful understanding of this phenomenon by exploiting a certain desirable property of the flow of a continuous-time ODE, namely that integral curves are non-intersecting. Our work suggests that, due to their intrinsic robustness, it is promising to use neural ODEs as a basic block for building robust deep network models. To further enhance the robustness of vanilla neural ODEs, we propose the time-invariant steady neural ODE (TisODE), which regularizes the flow on perturbed data via the time-invariant property and the imposition of a steady-state constraint. We show that the TisODE method outperforms vanilla neural ODEs and also can work in conjunction with other state-of-the-art architectural methods to build more robust deep networks.

\section{Introduction}

Neural ordinary differential equations \citep{chen2018neural} form a family of models that approximate nonlinear mappings by using continuous-time ODEs. Due to their desirable properties, such as invertibility and parameter efficiency, neural ODEs have attracted increasing attention recently \citep{dupont2019augmented, liu2019neural}. For example, \citet{grathwohl2018ffjord} proposed a neural ODE-based generative model---the FFJORD---to solve inverse problems; \citet{quaglino2019accelerating} used a higher-order approximation of the states in a neural ODE, and proposed the SNet to accelerate computation. Along with the wider deployment of neural ODEs, robustness issues come to the fore. However, the robustness of neural ODEs is still yet unclear. In particular, it is unclear how robust neural ODEs are in comparison to the widely-used CNNs. Robustness properties of CNNs  have been studied extensively. In this work, we present the first systematic study on exploring the robustness properties of neural ODEs.

\begin{figure}[t!]
    \centering
    \includegraphics[width=0.25\textwidth]{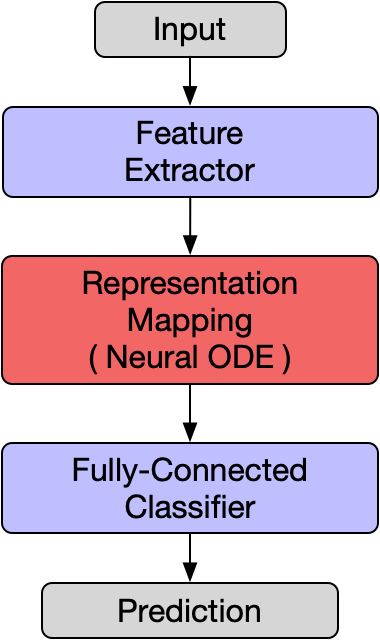}
    \caption[The architecture of an ODEnet.]{The architecture of an ODEnet. The neural ODE block serves as a dimension-preserving nonlinear mapping.}
    \label{fig:arch-ODEnet}
\end{figure}

To do so, we consider the task of image classification. We expect that results would be similar for other machine learning tasks such as  regression. Neural ODEs are dimension-preserving mappings, but a classification model transforms  a high-dimensional input---such as an image---into an output whose dimension is equal to the number of classes. Thus, we consider the neural ODE-based classification network (ODEnet) whose architecture is  shown in Figure \ref{fig:arch-ODEnet}. An ODEnet consists of three components: the feature extractor (FE) consists of convolutional layers which maps an input datum to a multi-channel feature map, a neural ODE that serves as the nonlinear representation mapping (RM), and the fully-connected classifier (FCC)  that  generates a prediction vector based on the output of the RM. 

The robustness of a classification model can be evaluated through the lens of its performance on perturbed images. To   comprehensively investigate the robustness of neural ODEs, we perturb original images with  commonly-used perturbations, namely, random Gaussian noise~\citep{szegedy2013intriguing} and harmful adversarial examples~\citep{goodfellow_explaining_2015, madry_towards_2018}.   We conduct experiments in two common settings---training the model only on authentic non-perturbed images and training the model on authentic images as well as the Gaussian perturbed ones.   We observe that ODEnets are more robust compared to CNN models against all types of perturbations in both settings. We then provide an insightful understanding of such intriguing robustness of neural ODEs by exploiting a certain property of the flow \citep{dupont2019augmented}, namely that integral curves that start at distinct initial states are non-intersecting. The flow of a continuous-time ODE is defined as the family of solutions/paths traversed by the state,  starting from different initial points, and an integral curve is a specific solution for a given initial point. The non-intersecting property indicates that an integral curve starting from some point is constrained by the integral curves starting from that point's neighborhood. Thus, in an ODEnet, if a correctly classified datum is slightly perturbed, the integral curve associated to its   perturbed version would not change too much from the original one. Consequently, the perturbed datum could still be correctly classified. Thus, there exists intrinsic robustness regularization in ODEnets, which is absent from CNNs.

Motivated by this property of the neural ODE flow, we attempt to explore a more robust neural ODE architecture by introducing stronger regularization on the flow. We thus propose a \emph{\textbf{T}ime-\textbf{I}nvariant \textbf{S}teady neural \textbf{ODE}} (TisODE). The TisODE removes the time dependence of the dynamics in an ODE and imposes a steady-state constraint on the integral curves. Removing the time dependence of the derivative results in the time-invariant property of the ODE. To wit, given a solution $\vz_1(t)$, another solution $\widetilde{\vz}_1(t)$, with an initial state $\widetilde{\vz}_1(0)=\vz_1(T')$ for some $T'>0$, can be regarded as the $-T'$- shift version of $\vz_1(t)$. Such a time-invariant property would make bounding the difference between output states convenient. 
To elaborate, let the  output of a neural ODE correspond to states at time $T>0$. By the time-invariant property, the difference between outputs, $\|\widetilde{\vz}_1(T)-\vz_1(T)\|$,  equals to $\|\vz_1(T+T')-\vz_1(T)\|$. To control this  distance, a steady-state regularization term is introduced to the overall objective to constrain the change of a state after time exceeds $T$. With  the time-invariant property and the steady-state term, we show that TisODE even is more robust. We do so by evaluating the robustness of TisODE-based classifiers against various types of perturbations and observe that such models are more robust than vanilla ODE-based models.

In addition, some other effective architectural solutions have also been recently proposed to improve the robustness of CNNs. For example,~\citet{xie_mitigating_2018} randomly resizes or pads zeros into test images to destroy the specific structure of adversarial perturbations. Besides, the model proposed by \citet{xie_feature_2019} contains feature denoising filters to remove the feature-level patterns of adversarial examples. We conduct experiments to show that our proposed TisODE can work seamlessly and in conjunction with these methods to further boost the robustness of deep models. Thus, the proposed TisODE can be used as a generally applicable and effective component for improving the robustness of deep models.

In summary, our contributions are as follows. Firstly,  we are the first to provide a systematic empirical study on the robustness of neural ODEs and find that the neural ODE-based models are more robust compared to conventional CNN models. This finding inspires new applications of neural ODEs in improving robustness of deep models, a problem that concerns many deep learning theorists and practitioners alike.  Secondly, we propose the TisODE method, which is simple yet effective in significantly boosting the robustness of neural ODEs. Moreover, the proposed TisODE can also be used in conjunction with other state-of-the-art robust architectures. Thus, TisODE can serve as a drop-in module to improve the robustness of deep models effectively.

\section{Preliminaries on neural ODE}

It has been shown that a residual block \citep{he_deep_2016} can be interpreted as the discrete approximation of an ODE  by setting the discretization step to be one. When the discretization step approaches zero, it yields a family of neural networks, which are called neural ODEs~\citep{chen2018neural}. Formally, in a neural ODE, the relation between input and output is characterized by the following set of equations: 
\begin{equation}
        \displaystyle\frac{\mathrm{d}\vz(t)}{\mathrm{d}t}=f_{\theta}(\vz(t), t), \quad \vz(0)= \vz_{\text{in}},    \quad     \vz_{\text{out}} = \vz(T),
    \label{eq:NODE}
\end{equation}
where $f_{\theta}:\sR^d \times [0,\infty)\rightarrow \sR^d$ denotes the trainable layers that are parameterized by weights $\theta$ and $\vz: [0,\infty) \rightarrow \sR^d$ represents the $d$-dimensional state of the neural ODE. 
We assume that $f_{\theta}$ is continuous in $t$ and globally Lipschitz continuous in $\vz$.
In this case, the input $\vz_{\text{in}}$ of the neural ODE corresponds to the state at $t=0$, and the output $\vz_{\text{out}}$ is associated to the state at some $T\in(0,\infty)$. Because $f_{\theta}$ governs how the state changes with respect to time $t$, we also use $f_{\theta}$ to denote the {\em dynamics} of the neural ODE.

Given input $\vz_{\text{in}}$,  the output $\vz_{\text{out}}$ can be computed by solving the ODE in (\ref{eq:NODE}). If $T$ is fixed, the output $\vz_{\text{out}}$ only depends on the input $\vz_{\text{in}}$ and the dynamics $f_{\theta}$, which also corresponds to the weighted layers in the neural ODE. Therefore, the neural ODE can be represented as the $d$-dimensional function $\phi_T(\cdot,\cdot)$ of the input $\vz_{\mathrm{in}}$ and the dynamics $f_\theta$, i.e., $$\vz_{\text{out}}=\vz(T)  = \vz(0) + \int_{0}^{T}f_{\theta}(\vz(t),t)\, \mathrm{d}t= \phi_T(\vz_{\text{in}}, f_{\theta}).$$
The terminal time $T$ of the output state $\vz(T)$ is set to be $1$ in practice. Several methods have been proposed for training neural ODEs, such as the adjoint sensitivity method~\citep{chen2018neural}, SNet~\citep{quaglino2019accelerating}, and the auto-differentiation technique~\citep{paszke2017automatic}. In this work, we use the most straightforward technique, i.e., updating the weights $\theta$ with the auto-differentiation technique in the PyTorch framework. 

\section{An empirical study on the robustness of ODEnets}
\label{sec:ode-vs-cnn}
Robustness of deep models has gained increased attention, as it is imperative  that deep models employed in critical applications, such as healthcare,  are robust. The robustness of a model is measured by the sensitivity of the prediction with respect to small perturbations on the inputs. In this study, we consider three commonly-used perturbation schemes, namely random Gaussian perturbations,  FGSM~\citep{goodfellow_explaining_2015} adversarial examples, and   PGD~\citep{madry_towards_2018} adversarial examples. These perturbation schemes reflect noise and adversarial robustness properties of the investigated models respectively. We evaluate the robustness via the classification accuracies on perturbed images, in which the original non-perturbed versions of these images are all correctly classified. 

For a fair comparison with conventional CNN models, we made sure that the number of parameters of an ODEnet is close to that of its counterpart CNN model. Specifically, the ODEnet shares the same network architecture with the CNN model for the FE and FCC parts. The only difference is that, for the RM part,  the input of the ODE-based RM is concatenated with one more channel which represents the time $t$, while the RM in a CNN model has a skip connection and serves as a residual block. During the training phase, all the hyperparameters are kept the same, including training epochs, learning rate schedules, and weight decay coefficients. Each model is trained   \emph{three} times with different random seeds, and we report the average performance (classification accuracy) together with the standard deviation.

\subsection{Experimental settings}
\textbf{Dataset:} We conduct experiments to compare the robustness of ODEnets with CNN models on three datasets, i.e., the MNIST~\citep{lecun1998gradient}, the SVHN \citep{netzer2011reading},  and a subset of the ImageNet datset~\citep{deng2009imagenet}. We call the subset ImgNet10 since it is collected from 10 synsets of ImageNet: dog, bird, car, fish, monkey, turtle, lizard, bridge, cow, and crab. We selected 3,000 training images and 300 test images from each synset and resized all images to $128 \times 128$.  

\textbf{Architectures:} On the MNIST dataset, both the ODEnet and the CNN model consists of four convolutional layers and one fully-connected layer. The total number of parameters of the two models is around 140k. On the SVHN dataset, the networks are similar to those for the MNIST; we only changed the input channels of the first convolutional layer to three. On the ImgNet10 dataset, there are nine convolutional layers and one fully-connected layer for both the ODEnet and the CNN model. The numbers of parameters is approximately 280k. In practice, the neural ODE can be solved with different numerical solvers such as the Euler method and the Runge-Kutta methods \citep{chen2018neural}. Here, we use the easily-implemented Euler method in the experiments. To balance the computation and the continuity of the flow, we solve the ODE initial value problem in equation~(\ref{eq:NODE}) by the Euler method with step size $0.1$. Our implementation builds on the open-source neural ODE codes.\footnote{\url{https://github.com/rtqichen/torchdiffeq}.} Details on the network architectures are included in the Appendix.

\textbf{Training:} The experiments are conducted using two settings on each dataset---training models only with original non-perturbed images and training models on original images together with their perturbed versions. In both settings, we added a weight decay term into the training objective to regularize the norm of the weights, since this can help control the model's representation capacity and improve the robustness of a neural network~\citep{sokolic2017robust}. In the second setting, images perturbed with random Gaussian noise are used to fine-tune the models, because augmenting the dataset with small perturbations can possibly improve the robustness of models and synthesizing Gaussian noise does not incur excessive computation time. 

\subsection{Robustness of ODEnets trained only on non-perturbed images}
\label{sec:none-pert}
The first question we are interested in is how robust ODEnets are against perturbations if the model is only trained on original non-perturbed images. We train CNNs and ODEnets to perform classification on three datasets and set the weight decay parameters for all models to be 0.0005. We make sure that both the well-trained ODEnets and CNN models have satisfactory performances on original non-perturbed images, i.e., around 99.5\% for MNIST, 95.0\% for the SVHN,  and 80.0\% for ImgNet10. Here, to calculate the robust accuracy, we first identify and select the samples whose clean versions are correctly classified by the obtained classifiers. Then, we compute the classification accuracy of the adversarial examples on the set of the selected samples.

Since Gaussian noise is  ubiquitous in modeling image degradation, we first evaluated the robustness of the models in the presence of zero-mean random Gaussian perturbations.  It has also been shown that a deep model is vulnerable to harmful adversarial examples, such as the FGSM~\citep{goodfellow_explaining_2015}. We are also interested in how robust ODEnets are in the presence of adversarial examples. The standard deviation $\sigma$ of Gaussian noise and the $l_{\infty}$-norm $\epsilon$ of the FGSM attack for each dataset are shown in Table \ref{tab:wd}. 

\begin{table}[ht]
    \centering
    \small
    \caption[Robustness comparison between CNNs and ODEnets.]{Robustness comparison of different models. We report their mean classification accuracies (\%) and standard deviations (mean $/$ std) on perturbed images from the MNIST, the SVHN, and the ImgNet10 datasets.  Two types of perturbations are used---zero-mean Gaussian noise  and FGSM adversarial attack. The results show that ODEnets are much more robust in comparison to CNN models.}
    \scalebox{0.9}{
    \begin{tabular}{c|ccc|ccc}
    \toprule
      &\multicolumn{3}{c|}{ Gaussian noise } & \multicolumn{3}{c}{ Adversarial attack} \\
    \midrule
    MNIST & $\sigma=\nicefrac{50}{255}$ & $\sigma=\nicefrac{75}{255}$ & $\sigma=\nicefrac{100}{255}$ & FGSM-0.15 & FGSM-0.3 & FGSM-0.5 \\
    \hline
    CNN & 98.1$/$0.7 & 85.8$/$4.3 & 56.4$/$5.6 & 63.4$/$2.3 & 24.0$/$8.9 & 8.3$/$3.2 \\
    ODEnet & \textbf{98.7}$/$0.6 & \textbf{90.6}$/$5.4 & \textbf{73.2}$/$8.6 & \textbf{83.5}$/$0.9 & \textbf{42.1}$/$2.4 & \textbf{14.3}$/$2.1\\
    \midrule
    \midrule
    SVHN & $\sigma=\nicefrac{15}{255}$ & $\sigma=\nicefrac{25}{255}$ & $\sigma=\nicefrac{35}{255}$ & FGSM-3/255 & FGSM-5/255 & FGSM-8/255 \\
    \hline
    CNN &  90.0$/$1.2 & 76.3$/$2.7 & 60.9$/$3.9 & 29.2$/$2.9 &13.7$/$1.9 & 5.4$/$1.5\\
    ODEnet & \textbf{95.7}$/$0.7 & \textbf{88.1}$/$1.5 & \textbf{78.2}$/$2.1 & \textbf{58.2}$/$2.3 & \textbf{43.0}$/$1.3 & \textbf{30.9}$/$1.4\\
    \midrule
    \midrule
    ImgNet10 & $\sigma=\nicefrac{10}{255}$ & $\sigma=\nicefrac{15}{255}$ & $\sigma=\nicefrac{25}{255}$ & FGSM-5/255& FGSM-8/255& FGSM-16/255 \\
    \hline
    CNN & 80.1$/$1.8 & 63.3$/$2.0 & 40.8$/$2.7 & 28.5$/$0.5 & 18.1$/$0.7 & 9.4$/$1.2\\
    ODEnet  & \textbf{81.9}$/$2.0 & \textbf{67.5}$/$2.0 & \textbf{48.7}$/$2.6 & \textbf{36.2}$/$1.0 & \textbf{27.2}$/$1.1 & \textbf{14.4}$/$1.7\\
    \bottomrule
    \end{tabular}
    }
    \label{tab:wd}
\end{table}{}

From the results in Table~\ref{tab:wd}, we observe that the ODEnets demonstrate  superior robustness compared to CNNs for  all types of perturbations. On the MNIST dataset, in the presence of Gaussian perturbations with a large $\sigma$ of \nicefrac{100}{255},  the ODEnet produces much higher accuracy on perturbed images compared to the CNN model (73.2\% vs.\ 56.4\%). For the FGSM-0.3 adversarial examples, the accuracy of ONEnet is around twice as high as that of the CNN model.
On the SVHN dataset, ODEnets significantly outperform CNN models, e.g., for the FGSM-5/255 examples, the accuracy of the ODEnet is 43.0\%, which is much higher than that of the CNN model (13.7\%).
On the ImgNet10, for both cases of $\sigma= \nicefrac{25}{255}$ and FGSM-8/255,  ODEnet outperforms CNNs by a large margin of around 9\%.
 
\subsection{Robustness of ODEnets trained on original images together with Gaussian perturbations}
\label{sec:gaus-pert}
Training a model on original images together with their perturbed versions can improve the robustness of the model. As mentioned previously, Gaussian noise is commonly assumed to be present in  real-world images. Synthesizing Gaussian noise is also fast and easy. Thus, we add random Gaussian noise into the original images to generate their perturbed versions. ODEnets and CNN models are both trained on original images together with their perturbed versions. The standard deviation of the added Gaussian noise is randomly chosen from $\{ \nicefrac{50}{255}, \nicefrac{75}{255}, \nicefrac{100}{255}\}$ on the MNIST dataset, $\{ \nicefrac{15}{255}, \nicefrac{25}{255}, \nicefrac{35}{255}\}$ on the SVHN dataset, and $\{ \nicefrac{10}{255}, \nicefrac{15}{255}, \nicefrac{25}{255}\}$ on the ImgNet10. All other hyperparameters are kept the same as above.

\begin{table}[ht]
    \centering
    \caption[Robustness comparison of different models trained with Gaussian noisy images.]{Robustness comparison of different models. We report their mean classification accuracies (\%) and standard deviations (mean $/$ std) on perturbed images from the MNIST, the SVHN, and the ImgNet10 datsets.  Three types of perturbations are used---zero-mean Gaussian noise, FGSM adversarial attack, and PGD adversarial attack. The results show that ODEnets are more robust compared to CNN models.}
    \scalebox{0.9}{
    \begin{tabular}{c|c|cccc}
    \toprule
     & \multicolumn{1}{c|}{ Gaussian noise } & \multicolumn{4}{c}{ Adversarial attack} \\
    \midrule
    MNIST & $\sigma= \nicefrac{100}{255}$ & FGSM-0.3 & FGSM-0.5 & PGD-0.2 & PGD-0.3 \\
    \hline
    CNN & 98.7$/$0.1 & 54.2$/$1.1 & 15.8$/$1.3 & 32.9$/$3.7 & 0.0$/$0.0\\
    ODEnet & \textbf{99.4}$/$0.1 & \textbf{71.5}$/$1.1 & \textbf{19.9}$/$1.2 & \textbf{64.7}$/$1.8 & \textbf{13.0}$/$0.2\\
    \midrule
    \midrule
    SVHN & $\sigma= \nicefrac{35}{255}$ & FGSM-5/255 & FGSM-8/255 & PGD-3/255 & PGD-5/255 \\
    \hline
    CNN & 90.6$/$0.2 & 25.3$/$0.6 & 12.3$/$0.7 & 32.4$/$0.4 & 14.0$/$0.5\\
    ODEnet & \textbf{95.1}$/$0.1 & \textbf{49.4}$/$1.0 & \textbf{34.7}$/$0.5 & \textbf{50.9}$/$1.3 & \textbf{27.2}$/$1.4\\
    \midrule
    \midrule
    ImgNet10 & $\sigma= \nicefrac{25}{255}$ & FGSM-5/255 & FGSM-8/255 & PGD-3/255 & PGD-5/255 \\
    \hline
    CNN & 92.6$/$0.6 & 40.9$/$1.8 & 26.7$/$1.7 & 28.6$/$1.5 & 11.2$/$1.2\\
    ODEnet & 92.6$/$0.5 & \textbf{42.0}$/$0.4 & \textbf{29.0}$/$1.0 & \textbf{29.8}$/$0.4 & \textbf{12.3}$/$0.6 \\
    \bottomrule
    \end{tabular}
    }
    \label{tab:gaus}
\end{table}

The robustness of the models is evaluated under Gaussian perturbations, FGSM adversarial examples, and PGD \citep{madry_towards_2018} adversarial examples. The latter is a stronger attacker compared to the FGSM. The $l_{\infty}$-norm $\epsilon$ of the PGD attack for each dataset is shown in Table \ref{tab:gaus}. Based on the results, we observe that ODEnets consistently outperform CNN models on both two datasets. On the MNIST dataset, the ODEnet outperforms the CNN against all types of perturbations. In particular,  for the PGD-0.2 adversarial examples, the accuracy of the ODEnet (64.7\%) is much higher than that of the CNN (32.9\%). Besides, for the PGD-0.3 attack, the CNN is completely misled by the adversarial examples, but the ODEnet can still classify perturbed images with an accuracy of 13.0\%. 
On the SVHN dataset, ODEnets also show superior robustness in comparison to CNN models. For all the adversarial examples,   ODEnets outperform CNN models by a margin of at least 10 percentage points.
On the ImgNet10 dataset, the ODEnet also performs better than CNN models against all forms of adversarial examples.

\subsection{Insights on the robustness of ODEnets}
\label{sec:understanding}

\begin{figure}
    \centering
    \includegraphics[width=0.6\textwidth]{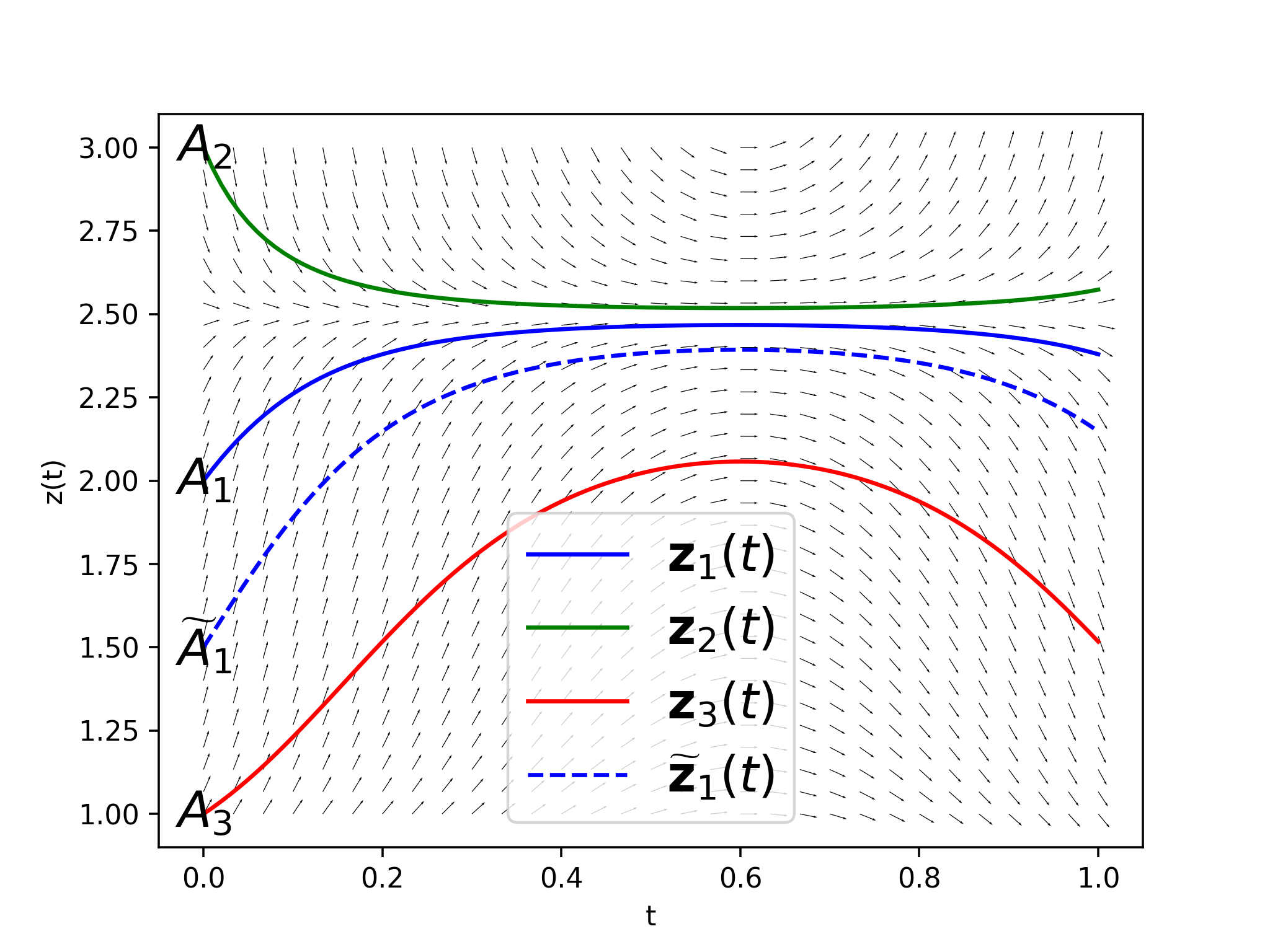}
    \caption[No integral curves intersect.]{No integral curves intersect. 
    The integral curve starting from $\widetilde{A_1}$ is always sandwiched between two integral curves starting from $A_1$ and $A_3$.}
    \label{fig:none-intersection}
\end{figure}

From the results in Sections \ref{sec:none-pert} and   \ref{sec:gaus-pert}, we find ODEnets are more robust compared to CNN models. Here, we attempt to provide an intuitive understanding of the robustness of the neural ODE. In an ODEnet, given some datum, the FE extracts an informative feature map from the datum. The neural ODE, serving as the RM, takes as input the feature map and performs a nonlinear mapping. In practice, we use the weight decay technique during training which regularizes the norm of weights in the FE part, so that the change of feature map in terms of a small perturbation on the input can be controlled. We aim to show that, in the neural ODE, a small change on the feature map will not lead to a large deviation from the original output associated with the feature map. 

\begin{theorem}[ODE integral curves never intersect.]
    \label{thm:none-intersection}
    Let $\vz_1(t)$ and $\vz_2(t)$ be two solutions of the ODE in (\ref{eq:NODE}) with different initial conditions, i.e. $\vz_1(0)\neq \vz_2(0)$. In (\ref{eq:NODE}), $f_{\theta}$ is continuous in $t$ and globally Lipschitz continuous in $\vz$.
    Then,  it holds that  $\vz_1(t)\neq \vz_2(t)$ for all $t\in[0, \infty)$. 
\end{theorem}

To illustrate Theorem \ref{thm:none-intersection} \citep{coddington1955theory,edwards1993elementary}, considering a simple 1-dimensional system in which the state is a scalar. As shown in Figure~\ref{fig:none-intersection}, equation (\ref{eq:NODE}) has a solution $z_1(t)$ starting from $A_1=(0, z_1(0))$, where $z_1(0)$ is the feature of some datum. Equation (\ref{eq:NODE}) also has another two solutions $z_2(t)$ and $z_3(t)$, whose starting points $A_2=(0, z_2(0))$ and $A_3=(0, z_3(0))$, both of which are close to $A_1$. Suppose $A_1$ is   between $A_2$ and $A_3$. By Theorem \ref{thm:none-intersection}, we know that the integral curve $z_1(t)$ is always sandwiched between the integral curves $z_2(t)$ and $z_3(t)$. 

Now, let $\epsilon<\min\{|z_2(0)-z_1(0)|, |z_3(0)-z_1(0)|\}$. Consider a solution $\widetilde{z}_1(t)$  of equation (\ref{eq:NODE}). The integral curve $\widetilde{z}_1(t)$  starts from a point $\widetilde{A}_1=(0,\widetilde{z}_1(0))$. The point $\widetilde{A}_1$ is in the $\epsilon$-neighborhood of $A_1$ with $|\widetilde{z}_1(0)-z_1(0)| < \epsilon$. By Theorem \ref{thm:none-intersection}, we know that $|\widetilde{z}_1(T)-z_1(T)| \leq |z_3(T)-z_2(T)|$. In other words, if any perturbation smaller than $\epsilon$ is added to the scalar $z_1(0)$ in $A_1$, the deviation from the original output $z_1(T)$ is bounded by the distance between $z_2(T)$ and $z_3(T)$. In contrast, in a CNN model, there is no such bound on the deviation from the original output. Thus,  we opine that due to this non-intersecting property, ODEnets are intrinsically robust.

\section{TisODE: boosting the robustness of  neural ODEs}

In the previous section, we presented an empirical study on the robustness of ODEnets  and observed that ODEnets are  more robust compared to CNN models. In this section, we explore how to boost the robustness of the vanilla neural ODE model further. This motivates the proposal of \emph{time-invariant steady neural ODEs} (TisODEs). 

\subsection{Time-invariant steady neural ODEs}

\begin{figure}
    \centering
    \includegraphics[width=0.6\textwidth]{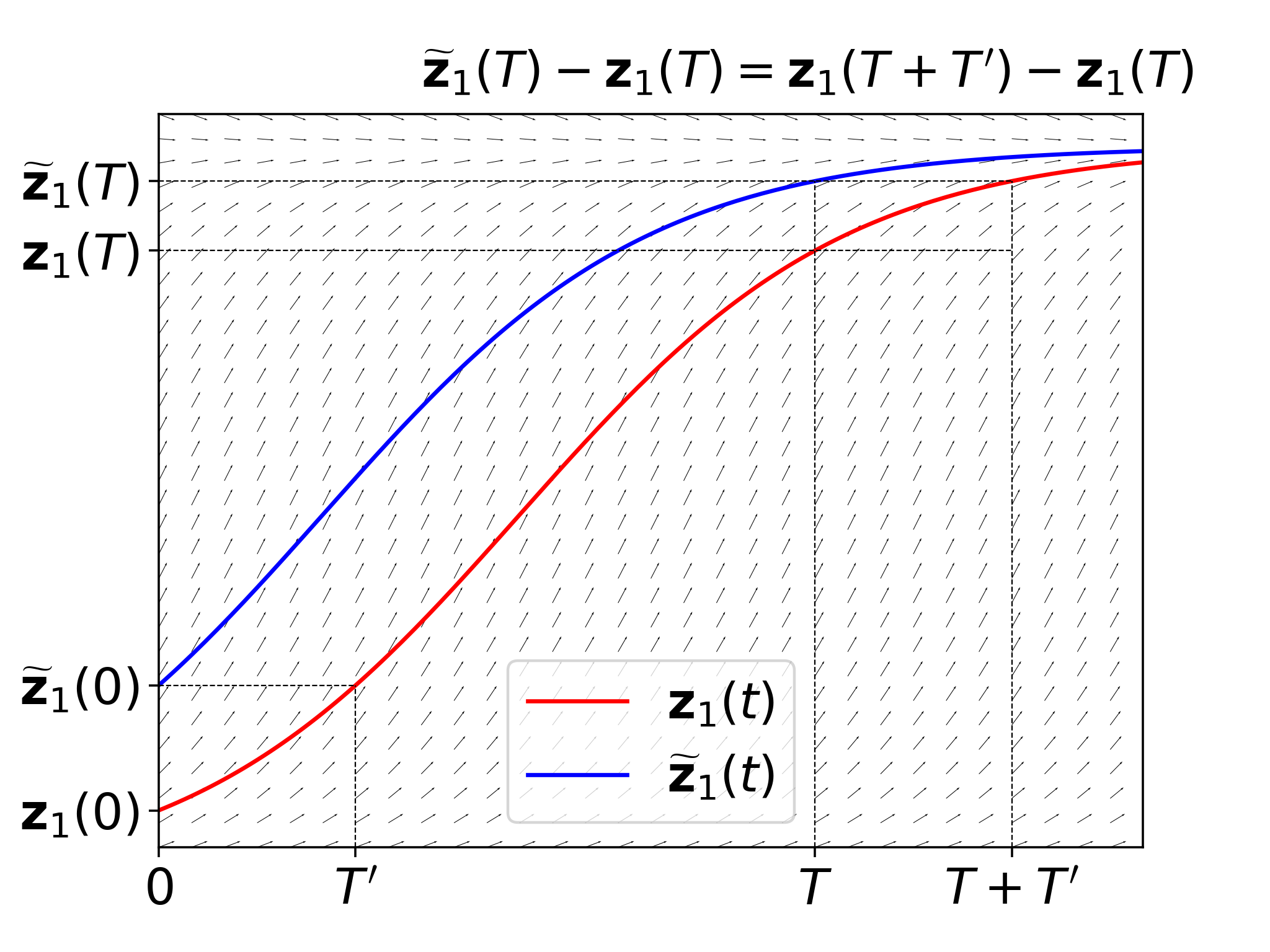}
    \caption[An illustration of the time-invariant property of ODEs.]{An illustration of the time-invariant property of ODEs. We can see that the curve $\widetilde{\vz}_1(t)$ is exactly the horizontal translation of $\vz_1(t)$ on the interval $[T',\infty)$.}
    \label{fig:TisODE}
\end{figure}

From the discussion in Section \ref{sec:understanding}, the key to improving the robustness of neural ODEs is to control the difference between neighboring integral curves. By Grownall's inequality \citep{howard1998gronwall} (see Theorem \ref{thm:gronwal} in the Appendix), we know that the difference between two terminal states is bounded by the difference between initial states multiplied by the exponential of the dynamics' Lipschitz constant. However, it is very difficult to bound the Lipschitz constant of the dynamics directly. Alternatively, we propose to achieve the goal of controlling the output deviation by following two steps: (i) removing the time dependence of the dynamics and (ii) imposing a certain steady-state constraint.

In the neural ODE characterized by equation~(\ref{eq:NODE}), the dynamics $f_{\theta}(\vz(t),t)$ depends on both the state $\vz(t)$ at time $t$ and the time $t$ itself. In contrast, if the neural ODE is modified to be time-invariant, the time dependence of the dynamics is removed. Consequently,  the dynamics depends {\em only} on the state $\vz$. So, we can rewrite the dynamics function as $f_{\theta}(\vz)$, and the neural ODE is characterized as
       \begin{equation}
        \begin{cases}
            \displaystyle\frac{\mathrm{d}\vz(t)}{\mathrm{d}t}=f_{\theta}(\vz(t));\\  
             \vz(0)= \vz_{\text{in}};    \\
            \vz_{\text{out}} = \vz(T). \\            
        \end{cases}
        \label{eq:ode-tiv}
    \end{equation}

Let $\vz_1(t)$ be a solution of (\ref{eq:ode-tiv}) on $[0,\infty)$ and $\epsilon>0$ be a small positive value. We define the set $\sM_1=\{(\vz_1(t), t)|t\in [0,T], \|\vz_1(t)-\vz_1(0)\|\leq \epsilon\}$. This set  contains all points on the curve of $\vz_1(t)$ during $[0,T]$ that are also inside the $\epsilon$-neighborhood of $\vz_1(0)$. For some element $(\vz_1(T'),T')\in \sM_1$, let 
$\widetilde{\vz}_1(t)$ be the solution of (\ref{eq:ode-tiv}) which starts from $\widetilde{\vz}_1(0)=\vz_1(T')$. Then we have
    \begin{equation}
        \widetilde{\vz}_1(t) = \vz_1(t+T')
        \label{eq:tiv-time-delay}
    \end{equation}
    for all $t$ in  $[0, \infty)$. The property shown in equation~(\ref{eq:tiv-time-delay}) is known as the  \emph{time-invariant property}. It  indicates that the integral curve $\widetilde{\vz}_1(t)$ is the $-T'$ shift of $\vz_1(t)$ (Figure~\ref{fig:TisODE}).

We can regard $\widetilde{\vz}_1(0)$ as a slightly perturbed version of $\vz_1(0)$, and we are interested in how large the difference between $\widetilde{\vz}_1(T)$ and $\vz_1(T)$ is. In a robust model, the difference should be small. By equation (\ref{eq:tiv-time-delay}), we have $\|\widetilde{\vz}_1(T)-\vz_1(T)\| = \|\vz_1(T+T')-\vz_1(T)\|$. Since $T'\in[0,T]$, the difference between $\vz_1(T)$ and $\widetilde{\vz}_1(T)$ can be bounded as follows, 
\begin{align}
\|\widetilde{\vz}_1(T)\!-\!\vz_1(T)\| \! & =\! \left\|\int_{T}^{T+T'}f_{\theta}(\vz_1(t))\,\mathrm{d}t\right\|		\nonumber \\
&	\!\leq\! \left\|\int_{T}^{T+T'}|f_{\theta}(\vz_1(t))|\,\mathrm{d}t\right\|		\nonumber \\
&	\! \leq \!\left\|\int_{T}^{2T}|f_{\theta}(\vz_1(t))|\,\mathrm{d}t \right\|,
\label{eq:TisODE-terminal-diff}
\end{align}
where all norms are $\ell_2$ norms and $|f_{\theta}|$ denotes the element-wise absolute operation of a vector-valued  function $f_\theta$. That is to say, the difference between $\widetilde{\vz}_1(T)$ and $\vz_1(T)$ can be bounded  by only using the information of the curve $\vz_1(t)$. For any $t'\in [0,T]$ and element $(\vz_1(t'),t') \in \sM_1$, consider the integral curve that starts from $\vz_1(t')$. The difference between the output state of this curve and $\vz_1(T)$ satisfies inequality~(\ref{eq:TisODE-terminal-diff}).

Therefore, we  propose to add an additional term $L_{\mathrm{ss}} $ to the loss function when training the time-invariant neural ODE:
\begin{equation}
    L_{\mathrm{ss}} = \sum_{i=1}^{N} \left\|\int_{T}^{2T}|f_{\theta}(\vz_{i}(t))|\,\mathrm{d}t \right\|,
\end{equation}
where $N$ is the number of samples in the training set and $\vz_{i}(t)$ is the solution whose initial state equals to the feature of the $i^{\mathrm{th}}$ sample. The regularization term $L_{\mathrm{ss}}$ is termed as the steady-state loss. This terminology ``steady state'' is borrowed from the dynamical systems literature. In a stable dynamical system, the states stabilize around a fixed point, known as the {\em steady-state}, as time tends to infinity. If we can ensure that $L_{\mathrm{ss}}$ is small, for each sample, the outputs of all the points in $\sM_i$ will stabilize around $\vz_i(T)$. Consequently, the model is  robust. This modification of the neural ODE is dubbed \emph{Time-invariant steady neural ODE}.

\subsection{Evaluating robustness of TisODE-based classifiers}
\label{sec:TisODE}

\begin{table}[ht]
    \centering
    \caption[Robustness improvement via TisODE.]{Classification accuracy (mean $/$ std in \%) on perturbed images from MNIST, SVHN and ImgNet10. To evaluate the robustness of classifiers, we use three types of perturbations, namely zero-mean Gaussian noise with standard deviation $\sigma$, FGSM attack and PGD attack. From the results, the proposed TisODE effectively improve the robustness of the vanilla neural ODE.}
    \scalebox{0.9}{
    \begin{tabular}{c|c|cccc}
    \toprule
    & \multicolumn{1}{c|}{ Gaussian noise } & \multicolumn{4}{c}{ Adversarial attack} \\
    \midrule
    MNIST & $\sigma= \nicefrac{100}{255}$ &FGSM-0.3 & FGSM-0.5 & PGD-0.2 & PGD-0.3 \\
    \hline
    CNN & 98.7$/$0.1 & 54.2$/$1.1 & 15.8$/$1.3 & 32.9$/$3.7 & 0.0$/$0.0\\
    ODEnet & 99.4$/$0.1 &  71.5$/$1.1 & 19.9$/$1.2 & 64.7$/$1.8 & 13.0$/$0.2\\
    TisODE & \textbf{99.6}$/$0.0 &  \textbf{75.7}$/$1.4 & \textbf{26.5}$/$3.8 & \textbf{67.4}$/$1.5&\textbf{13.2}$/$1.0\\
    \midrule
    \midrule
    SVHN & $\sigma= \nicefrac{35}{255}$ & FGSM-5/255 & FGSM-8/255 & PGD-3/255 & PGD-5/255 \\
    \hline
    CNN & 90.6$/$0.2 & 25.3$/$0.6 & 12.3$/$0.7 & 32.4$/$0.4 & 14.0$/$0.5\\
    ODEnet & \textbf{95.1}$/$0.1 & 49.4$/$1.0 & 34.7$/$0.5 & 50.9$/$1.3 & 27.2$/$1.4\\
    TisODE & 94.9$/$0.1 & \textbf{51.6}$/$1.2 & \textbf{38.2}$/$1.9 & \textbf{52.0}$/$0.9 & \textbf{28.2}$/$0.3 \\
    \midrule
    \midrule
    ImgNet10 & $\sigma= \nicefrac{25}{255}$ & FGSM-5/255 & FGSM-8/255 & PGD-3/255 & PGD-5/255 \\
    \hline
    CNN & 92.6$/$0.6 & 40.9$/$1.8 & 26.7$/$1.7 & 28.6$/$1.5 & 11.2$/$1.2\\
    ODEnet & 92.6$/$0.5 & 42.0$/$0.4 & 29.0$/$1.0 & 29.8$/$0.4 & 12.3$/$0.6 \\
    TisODE & \textbf{92.8}$/$0.4 & \textbf{44.3}$/$0.7 & \textbf{31.4}$/$1.1 & \textbf{31.1}$/$1.2 & \textbf{14.5}$/$1.1 \\
    \bottomrule
    \end{tabular}
    }
    \label{tab:TisODE}
\end{table}{}
Here, we conduct experiments to evaluate the robustness of our proposed TisODE, and compare TisODE-based models with the vanilla ODEnets. We train all models with original non-perturbed images together with their Gaussian perturbed versions. The regularization parameter for the steady-state loss $L_{\mathrm{ss}}$ is set to be $0.1$. All other hyperparameters are exactly the same as those in Section~\ref{sec:gaus-pert}. We ensure that the well-trained CNNs, ODEnets, and the TisODE-based classifiers all have similar performances on the orginal non-perturbed images, i.e., around 99.5\% for MNIST, 95.0\% for SVHN, and 80.0\% for ImgNet10.

From the results in Table \ref{tab:TisODE}, we can see that our proposed TisODE-based models are clearly more robust compared to vanilla ODEnets. On the MNIST dataset, when combating FGSM-0.3 attacks, the TisODE-based models outperform vanilla ODEnets by more than $4$ percentage points. For the FGSM-0.5 adversarial examples, the accuracy of the TisODE-based model is $6$ percentage points better. On the SVHN dataset, the TisODE-based models perform better in terms of all forms of adversarial examples. On the ImgNet10 dataset, the TisODE-based models also outperform vanilla ODE-based models on all types of perturbations. In the presence of FGSM and PGD-5/255 examples, the accuracies are enhanced by more than $2$ percentage points.

\subsection{TisODE - A generally applicable drop-in technique for improving the robustness of deep networks}

In view of the excellent robustness of the TisODE, we claim that the proposed TisODE can be used as a general drop-in module for improving the robustness of deep networks. We support this claim by showing the TisODE can work in conjunction with other state-of-the-art techniques and further boost the models' robustness. These techniques include the feature denoising (FDn) method \citep{xie_feature_2019} and the input randomization (IR) method \citep{xie_mitigating_2018}.  We conduct experiments on the MNIST and SVHN datasets. All models are trained with original non-perturbed images together with their Gaussian perturbed versions. We show that models using the FDn/IRd technique becomes much more robust when equipped with the TisODE.  In the FDn experiments, the dot-product non-local denoising layer \citep{xie_feature_2019} is added to the head of the fully-connected classifier.

\begin{table}[ht]
    \centering
    \caption[Robustness improvement via TisODE in conjunction with other techniques.]{Classification accuracy (mean  $/$ std in  \%) on perturbed images from MNIST and SVHN. We evaluate against three types of perturbations, namely zero-mean Gaussian noise with standard deviation $\sigma$, FGSM attack and PGD attack. From the results, upon the CNNs modified with FDn and IRd, using TisODE can further improve the robustness.}
    \scalebox{0.9}{
    \begin{tabular}{c|c|cccc}
    \toprule
    & \multicolumn{1}{c|}{ Gaussian noise } & \multicolumn{4}{c}{ Adversarial attack} \\
    \midrule
    MNIST & $\sigma= \nicefrac{100}{255}$ & FGSM-0.3 & FGSM-0.5 & PGD-0.2 & PGD-0.3 \\
    \hline
    CNN               & 98.7$/$0.1 & 54.2$/$1.1  & 15.8$/$1.3  & 32.9$/$3.7    & 0.0$/$0.0\\
    \hline
    CNN-FDn       & 99.0$/$0.1  & 74.0$/$4.1 & 32.6$/$5.3  &     58.9$/$4.0 &8.2$/$2.6 \\
    TisODE-FDn    &  \textbf{99.4}$/$0.0   & 80.6$/$2.3 & 40.4$/$5.7 & 72.6$/$2.4 &28.2$/$3.6\\
    \hline
    CNN-IRd       & 95.3$/$0.9  & 78.1$/$2.2 & 36.7$/$2.1 & 79.6$/$1.9    & 55.5$/$2.9 \\
    TisODE-IRd      & 97.6$/$0.1 &  \textbf{86.8}$/$2.3 & \textbf{49.1}$/$0.2 & \textbf{88.8}$/$0.9 & \textbf{66.0}$/$0.9 \\
    \midrule
    \midrule
    SVHN & $\sigma= \nicefrac{35}{255}$ & FGSM-5/255 & FGSM-8/255 & PGD-3/255 & PGD-5/255 \\
    \hline
    CNN & 90.6$/$0.2 & 25.3$/$0.6 & 12.3$/$0.7 & 32.4$/$0.4 & 14.0$/$0.5\\
    \hline
    CNN-FDn    & 92.4$/$0.1 & 43.8$/$1.4 & 31.5$/$3.0 & 40.0$/$2.6 & 19.6$/$3.4   \\
    TisODE-FDn   & \textbf{95.2}$/$0.1 & 57.8$/$1.7 & 48.2$/$2.0 & 53.4$/$2.9 & 32.3$/$1.0\\
    \hline
    CNN-IRd    & 84.9$/$1.2 & 65.8$/$0.4 & 54.7$/$1.2 & 74.0$/$0.5 & 64.5$/$0.8   \\
    TisODE-IRd  & 91.7$/$0.5 & \textbf{74.4}$/$1.2 & \textbf{61.9}$/$1.8 & \textbf{81.6}$/$0.8 & \textbf{71.0}$/$0.5 \\
    \bottomrule
    \end{tabular}
    }
    \label{tab:drop-in}
\end{table}{} 

From Table \ref{tab:drop-in}, we observe that both FDn and IRd can effectively improve the adversarial robustness of vanilla CNN models (CNN-FDn, CNN-IRd). Furthermore, combining our proposed TisODE with FDn or IRd (TisODE-FDn, TisODE-IRd), the adversarial robustness of the resultant model is significantly enhanced. For example, on the MNIST dataset, the additional use of our TisODE increases the accuracies on the PGD-0.3 examples by at least 10 percentage points for both FDn (8.2\% to 28.2\%) and IRd (55.5\% to 66.0\%). However, on both MNIST and SVHN datasets, the IRd technique improves the robustness against adversarial examples, but its performance is worse  on random Gaussian noise. With the help of the TisODE, the degradation in the  robustness against random Gaussian noise  can be effectively ameliorated.

\section{Related works}
In this section, we briefly review related works on the neural ODE and works concerning improving the robustness of deep neural networks.

\textbf{Neural ODE:} The neural ODE~\citep{chen2018neural} method models the input and output as two states of a continuous-time dynamical system by approximating the dynamics of this system with trainable layers. Before the proposal of neural ODE, the idea of modeling nonlinear mappings using continuous-time dynamical systems was proposed in \citet{weinan2017proposal}. \citet{lu2017beyond} also showed that several popular network architectures could be interpreted as the discretization of a continuous-time ODE. For example, the ResNet \citep{he_deep_2016} and PolyNet \citep{zhang2017polynet} are associated with the Euler scheme and the FractalNet \citep{larsson2016fractalnet} is related to the Runge-Kutta scheme. In contrast to these discretization models, neural ODEs are endowed with an intrinsic invertibility property, which yields a family of invertible models for solving inverse problems \citep{ardizzone2018analyzing}, such as the FFJORD \citep{grathwohl2018ffjord}.

Recently, many researchers have conducted studies on neural ODEs from the perspectives of optimization techniques, approximation capabilities, and generalization. Concerning  the optimization of neural ODEs, the auto-differentiation techniques can effectively train ODEnets, but the training procedure is computationally and memory inefficient. To address this problem, \citet{chen2018neural} proposed to compute gradients using the adjoint sensitivity method \citep{pontryagin2018mathematical}, in which there is no need to store any intermediate quantities of the forward pass. Also in \citet{quaglino2019accelerating}, the authors proposed the SNet which accelerates the neural ODEs by expressing their dynamics as truncated series of Legendre polynomials. Concerning the approximation capability, \citet{dupont2019augmented} pointed out the limitations in approximation capabilities of neural ODEs because of the preserving of input topology. The authors proposed an augmented neural ODE which increases the dimension of states by concatenating zeros so that complex mappings can be learned with simple flow. The most relevant work to ours concerns strategies to improve the generalization of neural ODEs. In \citet{liu2019neural}, the authors proposed the neural stochastic differential equation (SDE) by injecting random noise to the dynamics function and showed that the generalization and robustness of vanilla neural ODEs could be improved.  However, our improvement on the neural ODEs is explored from a different perspective by introducing constraints on the flow. We empirically found that our proposal and the neural SDE can work in tandem to further boost the robustness of neural ODEs.

\textbf{Robust Improvement:}
A straightforward way of improving the robustness of a model is to smooth the loss surface by controlling the spectral norm of the Jacobian matrix of the loss function \citep{sokolic2017robust}. In terms of adversarial examples~\citep{chen2017zoo,carlini_towards_2017}, researchers have proposed adversarial training strategies~\citep{elsayed2018adversarial,tramer2017ensemble,madry_towards_2018} in which the model is fine-tuned with adversarial examples generated in real-time. However, generating adversarial examples is not computationally efficient, and there exists a trade-off between the adversarial robustness and the performance on original non-perturbed images~\citep{yan2018deep,tsipras_robustness_2018}. In \citet{wang2019resnets}, the authors model the ResNet as a transport equation, in which the adversarial vulnerability can be interpreted as the irregularity of the decision boundary. Consequently, a diffusion term is introduced to enhance the robustness of the neural nets. Besides, there are also some works that propose novel architectural defense mechanisms against adversarial examples. For example, \citet{xie_mitigating_2018} utilized random resizing and random padding to destroy the specific structure of adversarial perturbations; \citet{wang2018adversarial} and \citet{wang2018deep} improved the robustness of neural networks by replacing the output layers with novel interpolating functions; In \citet{xie_feature_2019}, the authors designed a feature denoising filter that can remove the perturbation's pattern from feature maps. In this work, we explore the intrinsic robustness of a specific novel architecture (neural ODE), and show that the proposed TisODE can improve the robustness of deep networks and can also work in tandem with these state-of-the-art methods to achieve further improvements.

\section{Chapter Summary}
In this chapter, we first empirically study the robustness of neural ODEs. Our studies reveal that neural ODE-based models are superior in terms of robustness compared to CNN models. We  then explore how to further boost the robustness of vanilla neural ODEs and propose the TisODE. Finally, we show that the proposed TisODE outperforms the vanilla neural ODE and also can work in conjunction with other state-of-the-art techniques to further improve the robustness of deep networks. Thus, the TisODE method is an effective drop-in module for building   robust deep models.




\section{Appendices}
\subsection{Networks used on the MNIST, the SVHN, and the ImgNet10 datasets}

\begin{table}[ht]
    \centering
    \caption{The architectures of the ODEnets on different datasets.} 
    \scalebox{0.9}{
    \begin{tabular}{c|c|c}
    \toprule
    MNIST  & Repetition & Layer \\
    \hline
    \multirow{2}{*}{FE} & $\times$1 & Conv(1, 64, 3, 1) + GroupNorm + ReLU \\

                        & $\times$1 & Conv(64, 64, 4, 2) + GroupNorm + ReLU \\
    \midrule
    RM & $\times$2            & Conv(64, 64, 3, 1) + GroupNorm + ReLU \\
    \midrule
    FCC & $\times$1 & AdaptiveAvgPool2d + Linear(64,10)\\
    \bottomrule
    \multicolumn{3}{c}{} \\
    \toprule
    SVHN  & Repetition & Layer \\
    \hline
    \multirow{2}{*}{FE} & $\times$1 & Conv(3, 64, 3, 1) + GroupNorm + ReLU \\

                        & $\times$1 & Conv(64, 64, 4, 2) + GroupNorm + ReLU \\
    \midrule
    RM & $\times$2            & Conv(64, 64, 3, 1) + GroupNorm + ReLU \\
    \midrule
    FCC & $\times$1 & AdaptiveAvgPool2d + Linear(64,10)\\
    \bottomrule
    \multicolumn{3}{c}{} \\
    \toprule
    ImgNet10 & Repetition & Layer \\
    \hline
    \multirow{4}{*}{FE} & $\times$1 & Conv(3, 32, 5, 2) + GroupNorm \\
                        & $\times$1    & MaxPooling(2) \\
                        & $\times$1 & BaiscBlock(32, 64, 2) \\
                        & $\times$1    & MaxPooling(2) \\
    \midrule
    \multirow{1}{*}{RM} & $\times$3        & BaiscBlock(64, 64, 1) \\
    \midrule
    FCC & $\times$1 & AdaptiveAvgPool2d + Linear(64,10)\\
    \bottomrule
    \end{tabular}
	}
    \label{tab:nets}
\end{table}

In Table \ref{tab:nets}, the four arguments of the Conv layer represent the input channel, output channel, kernel size, and the stride. The two arguments of the Linear layer represents the input dimension and the output dimension of this fully-connected layer. In the network on the ImgNet10, the BasicBlock refers to the standard architecture in \citep{he_deep_2016}, the three arguments of the BasicBlock represent the input channel, output channel and the stride of the Conv layers inside the block. Note that we replace the BatchNorm layers in BasicBlocks as the GroupNorm to guarantee that the dynamics of each datum is independent of other data in the same mini-batch.

\subsection{The Construction of ImgNet10 dataset} 
\begin{table}[ht]
    \centering
    \caption{The corresponding indexes to each class in the original ImageNet dataset}
    \scalebox{0.9}{
    \begin{tabular}{c|ccc}
    \toprule
    Class & \multicolumn{3}{c}{Indexing}\\
    \hline
    dog & n02090721, & n02091032, & n02088094 \\
    bird & n01532829, & n01558993, & n01534433 \\
    car & n02814533, & n03930630, & n03100240 \\
    fish & n01484850, & n01491361, & n01494475 \\
    monkey & n02483708, & n02484975, & n02486261 \\
    turtle & n01664065, & n01665541, & n01667114 \\
    lizard & n01677366, & n01682714, & n01685808 \\
    bridge & n03933933, & n04366367, & n04311004 \\
    cow & n02403003, & n02408429, & n02410509 \\
    crab & n01980166, & n01978455, & n01981276 \\
    \bottomrule
    \end{tabular}
    }
    \label{tab:ImgNet10}
\end{table}

\subsection{Gronwall's Inequality}
We formally state the Gronwall's Inequality here, following the version in \citep{howard1998gronwall}.

\begin{theorem}
    \label{thm:gronwal}
    Let $U \subset \sR^d$ be an open set. Let $f: U \times [0, T] \rightarrow \sR^d$ be a continuous function and let $\vz_1$, $\vz_2$: $[0,T]\rightarrow U$ satisfy the initial value problems:
    \begin{align}
        \frac{\mathrm{d} \vz_1(t)}{\mathrm{d} t} & = f(\vz_1(t), t), \quad \vz_1(t)=\vx_1  \nonumber \\
        \frac{\mathrm{d} \vz_2(t)}{\mathrm{d}t} & = f(\vz_2(t), t), \quad \vz_2(t)=\vx_2  \nonumber
    \end{align}
    Assume there is a constant $C\geq 0$ such that, for all $t\in [0,T]$,
    $$\|f(\vz_2(t),t) - f(\vz_1(t),t))\| \leq C\|\vz_2(t)-\vz_1(t)\|$$
    Then, for any $t\in [0,T]$,
    $$\|\vz_1(t) - \vz_2(t)\| \leq \|\vx_2-\vx_1\| \cdot e^{Ct}.$$
\end{theorem}

\subsection{More experimental results}

\subsubsection{Comparison in the setting of Adversarial training}
We implement the adversarial training of the models on the MNIST dataset, and the adversarial examples for training are generated in real-time via the FGSM method  (epsilon=0.3) during each epoch \citep{madry_towards_2018}. The results of the adversarially trained models are shown in Table \ref{tab:review1}. We can observe that the neural ODE-based models are consistently more robust than CNN models. The proposed TisODE also outperforms the vanilla neural ODE.

\begin{table}[ht]
    \centering
    \caption[Robustness comparison of different models under adversarial training.]{Classification accuracy (\%) on perturbed images from MNIST. To evaluate the robustness of classifiers, we use three types of perturbations, namely zero-mean Gaussian noise with standard deviation $\sigma$, FGSM attack and PGD attack.}
    \scalebox{0.9}{
    \begin{tabular}{c|c|ccc}
    \toprule
    & \multicolumn{1}{c|}{ Gaussian noise } & \multicolumn{3}{c}{ Adversarial attack} \\
    \midrule
    MNIST & $\sigma= \nicefrac{100}{255}$ &FGSM-0.3 & FGSM-0.5 & PGD-0.3 \\
    \hline
    CNN & 58.0 & 98.4 & 21.1 & 5.3\\
    ODEnet & 84.2 & 99.1 & 36.0 & 12.3\\
    TisODE & \textbf{87.9} & \textbf{99.1} & \textbf{66.5} & \textbf{78.9}\\
    \bottomrule
    \end{tabular}
    }
    \label{tab:review1}
\end{table}{}

\subsubsection{Experiments on the CIFAR10 dataset}
We conduct experiments on CIFAR10 to compare the robustness of CNN and neural ODE-based models. We train all the models only with original non-perturbed images and evaluate the robustness of models against random Gaussian noise and FGSM adversarial attacks. The results are shown in Table \ref{tab:review2}. We can observe that the ONENet is more robust than the CNN model in terms of both the random noise and the FGSM attack. Besides, our proposal, TisODE, can improve the robustness of the vanilla neural ODE.

\begin{table}[ht]
    \centering
    \caption[Robustness improvement on CIFAR10.]{Classification accuracy (\%) on perturbed images from CIFAR10. To evaluate the robustness of classifiers, we use two types of perturbations, namely zero-mean Gaussian noise with standard deviation $\sigma$ and FGSM attack.}
    \scalebox{0.9}{
    \begin{tabular}{c|cc|cc}
    \toprule
    & \multicolumn{2}{c|}{ Gaussian noise } & \multicolumn{2}{c}{ Adversarial attack} \\
    \midrule
    CIFAR10 & $\sigma= \nicefrac{15}{255}$ &$\sigma= \nicefrac{20}{255}$ & FGSM-8/255 & FGSM-10/255 \\
    \hline
    CNN & 70.2 & 57.6 & 24.3 & 18.4\\
    ODEnet & 72.6 & 60.6 & 31.2 & 26.0\\
    TisODE & \textbf{74.3} & \textbf{62.0} & \textbf{33.6} & \textbf{26.8}\\
    \bottomrule
    \end{tabular}
    }
    \label{tab:review2}
\end{table}{}

Here, we control the number of parameters to be the same for all kinds of models. We use a small network, which consists of five convolutional layers and one linear layer.

\begin{table}[ht]
    \centering
    \caption{The architecture of the ODEnet on CIFAR10.} 
    \scalebox{0.9}{
    \begin{tabular}{c|c|c}
    \toprule
      & Repetition & Layer \\
    \hline
    \multirow{2}{*}{FE} & $\times$1 & Conv(3, 16, 3, 1) + GroupNorm + ReLU \\

                        & $\times$1 & Conv(16, 32, 3, 2) + GroupNorm + ReLU \\
                        & $\times$1 & Conv(32, 64, 3, 2) + GroupNorm + ReLU \\
    \midrule
    RM & $\times$2             & Conv(64, 64, 3, 1) + GroupNorm + ReLU \\
    \midrule
    FCC & $\times$1 & AdaptiveAvgPool2d + Linear(64,10)\\
    \bottomrule
    \end{tabular}
	}
    \label{tab:review3}
\end{table}

\subsubsection{An extension on the comparison between CNNs and ODEnets}
Here, we compare CNN and neural ODE-based models by controlling both the number of parameters and the number of function evaluations. We conduct experiments on the MNIST dataset, and all the models are trained only with original non-perturbed images.

For the neural ODE-based models, the time range is set from 0 to 1. We use the Euler method, and the step size is set to be 0.05. Thus the number of evaluations is $1/0.05=20$. For the CNN models (specifically ResNet), we repeatedly concatenate the residual block for 20 times, and these 20 blocks share the same weights.  Our experiments show that, in this condition, the neural ODE-based models still outperform the CNN models (FGSM-0.15: 87.5\% vs. 81.9\%, FGSM-0.3: 53.4\% vs. 49.7\%, PGD-0.2: 11.8\% vs. 4.8\%). 

%% file: Chapters/icml2021.tex
\chapter{Improving Adversarial Robustness of CNNs via Channel-wise Importance-based Feature Selection}
\label{chp:icml2021}

\nocite{yan_cifs_2021}

We investigate the adversarial robustness of CNNs from the perspective of channel-wise activations.  By comparing normally trained and adversarially trained models, we observe that adversarial training (AT) robustifies CNNs by aligning the channel-wise activations of adversarial data with those of their natural counterparts. However, the channels that are \textit{negatively-relevant} (NR) to predictions are still over-activated when processing adversarial data. Besides, we also observe that AT does not result in similar robustness for all classes. For the robust classes, channels with larger activation magnitudes are usually more \textit{positively-relevant} (PR) to predictions, but this alignment does not hold for the non-robust classes. Given these observations, we hypothesize that suppressing NR channels and aligning PR ones with their relevances further enhances the robustness of CNNs under AT. To examine this hypothesis, we introduce a novel mechanism, \textit{i.e.}, \underline{C}hannel-wise \underline{I}mportance-based \underline{F}eature \underline{S}election (CIFS). The CIFS manipulates channels' activations of certain layers by generating non-negative multipliers to these channels based on their relevances to predictions. Extensive experiments on benchmark datasets including CIFAR10 and SVHN clearly verify the hypothesis and CIFS's effectiveness of robustifying CNNs.

\section{Introduction }\label{introduction}

Convolutional neural networks (CNNs) have achieved tremendous successes in real-world applications, such as  autonomous vehicles \citep{grigorescu2020survey, hu2020panda} and computer-aided medical diagnoses \citep{trebeschi2017deep, shu2020enhancing}. However, CNNs have be shown vulnerable to well-crafted (and even minute) adversarial perturbations to inputs \citep{szegedy_intriguing_2014, goodfellow_explaining_2015, ilyas_adversarial_2019}. This has become hazardous in high-stakes applications such as medical diagnoses and autonomous vehicles.

Recently, many empirical defense methods have been proposed to secure CNNs against these adversarial perturbations, such as adversarial training (AT) \citep{madry_towards_2018}, input/feature denoising \citep{xie_feature_2019, du_rain_2020} and defensive distillation \citep{papernot_distillation_2016}. AT~\citep{madry_towards_2018}, which generates adversarial data on the fly for training CNNs, has emerged as one of the most successful methods. AT effectively robustifies CNNs but leads to a clear drop in the accuracies for natural data~\citep{tsipras_robustness_2018} and suffers from the problem of overfitting to adversarial data used for training \citep{rice_overfitting_2020,zhang_geometry-aware_2020,chen2021robust}. To ameliorate these problems, researchers have proposed variants of AT, including TRADES \citep{zhang_theoretically_2019} and Friendly-Adversarial-Training (FAT) \citep{zhang_attacks_2020}. 
To further robustify CNNs under AT, many works attempt to propose novel defense mechanism to mitigate the effects of adversarial data on features \citep{xie_feature_2019, du_rain_2020, xu_interpreting_2019}. For example, \citet{xie_feature_2019} found that adversarial data result in abnormal activations in the feature maps and performed feature denoising to remove the adversarial effects. Most of these works improved robustness by identifying and suppressing abnormalities \tit{at certain positions} across channels~(commonly referred to as feature maps in CNNs), whereas the other direction, namely, the connection between robustness and irregular activations of certain \tit{entire channels}, has received scant attention.

Since channels of CNNs' deeper layers are capable of extracting semantic characteristic features \citep{zeiler_visualizing_2014}, the process of making predictions usually relies heavily on aggregating information from various channels \citep{bach2015pixel}. As such, anomalous activations of certain channels may result in incorrect predictions. Thus, it is imperative to explore 
which channels are \tit{entirely} irregularly activated by adversarial data and which channels' activations benefit or degrade robustness.
By utilizing this connection, we will be able to further enhance the robustness of CNNs via suppressing or promoting certain vulnerable or reliable channels respectively. 

In this work, we attempt to build such a connection by comparing the channel-wise activations of non-robust (normally trained) and robustified (adversarially trained) CNNs. The \tit{channel-wise activations} are defined as the average activation magnitudes of all features within channels \citep{bai_improving_2021}. To identify what types of channels appear to be abnormal under attacks, we regard \tit{channels' relevances} to prediction results (formally defined in Equation (\ref{eq:relevance-assessment}) as $\vg^l$) as the gradients of the corresponding logits \tit{w.r.t} channel-wise activations. The channels, whose relevances to prediction results are positive or negative ($\vg^l_{[i]}>0$ or $\vg^l_{[i]}\leq 0$), are called \textit{positively-relevant} (PR) or \textit{negatively-relevant} (NR) channels.

On the one hand, we observe that, AT robustifies CNNs by aligning adversarial data's channel-wise activations with those of natural data. However, we find that the NR channels of adversarially trained CNNs are still over-activated by adversarial data (see Figure \ref{fig: w-vs-act_adv-auto}). Thus, we wonder: \textit{If we suppress NR channels during AT to facilitate the alignment of channels' activations, will it benefit CNNs' robustness?} 
On the other hand, we find that adversarially trained classification models do not enjoy similar robustness across all the classes (see Figure \ref{fig: w-vs-act_adv-auto} and \ref{fig: w-vs-act_adv-cat}). For classes with relatively good robustness, channels' activations usually align well with their relevances, i.e., channels with larger activations are more PR to labels. Given this phenomenon, a natural question arises: \textit{If we align channels' activations with their relevances during AT, will it improve the robustness of CNNs?} Regarding these two questions, we propose a unified hypothesis on robustness enhancement, denoted as $\mathcal H$: \ul{Suppressing NR channels and aligning channels' activations with their relevances to prediction results benefit the robustness of CNNs}.

To examine this hypothesis, we propose a novel mechanism, called \underline{C}hannel-wise \underline{I}mportance-based \underline{F}eature \underline{S}election (CIFS), which adjusts channels' activations with an \tit{importance mask} generated from channels' relevances. For a certain layer, CIFS first takes as input the representation of a data point at this layer and makes a \textit{raw prediction} for the data point by a \textit{probe network}. The probe serves as the surrogate for the subsequent classifier (the composition of subsequent layers) in the backbone and is jointly trained with the backbone under supervision of true labels. Then, CIFS computes the gradients of the sum of the top-$k$ logits \textit{w.r.t.} the channels' activations. We can obtain the \tit{relevance} of each channel to the top-$k$ prediction results by accumulating the gradients within the channel. Finally, CIFS generates a mask of importance scores for each channel by mapping channels' relevances monotonically to non-negative values. 
Through extensive experiments, we answer the two questions in the affirmative and confirm hypothesis $\mathcal H$. Indeed, our results show that CIFS clearly enhances the adversarial robustness of CNNs. 

We comprehensively evaluate the robustness of CIFS-modified CNNs on benchmark datasets against various attacks. On the CIFAR10 dataset, CIFS improves the robustness of the ResNet-18 by $4$ percentage points against the PGD-100 attack. We also observe that CIFS ameliorates the overfitting during AT. In particular, the robustness at the last epoch is close to that at the best epoch. Finally, we conduct an ablation study to further understand how various elements of CIFS affect the robustness enhancement, such as the top-$k$ feedback and architectures of the probe network.

\def \SubFigWidth {0.35} 
\def \SubImgWidth {.99}
\begin{figure}[!ht]
	\centering
	\begin{subfigure}{\SubFigWidth\linewidth}
		\includegraphics[width=\SubImgWidth \linewidth]{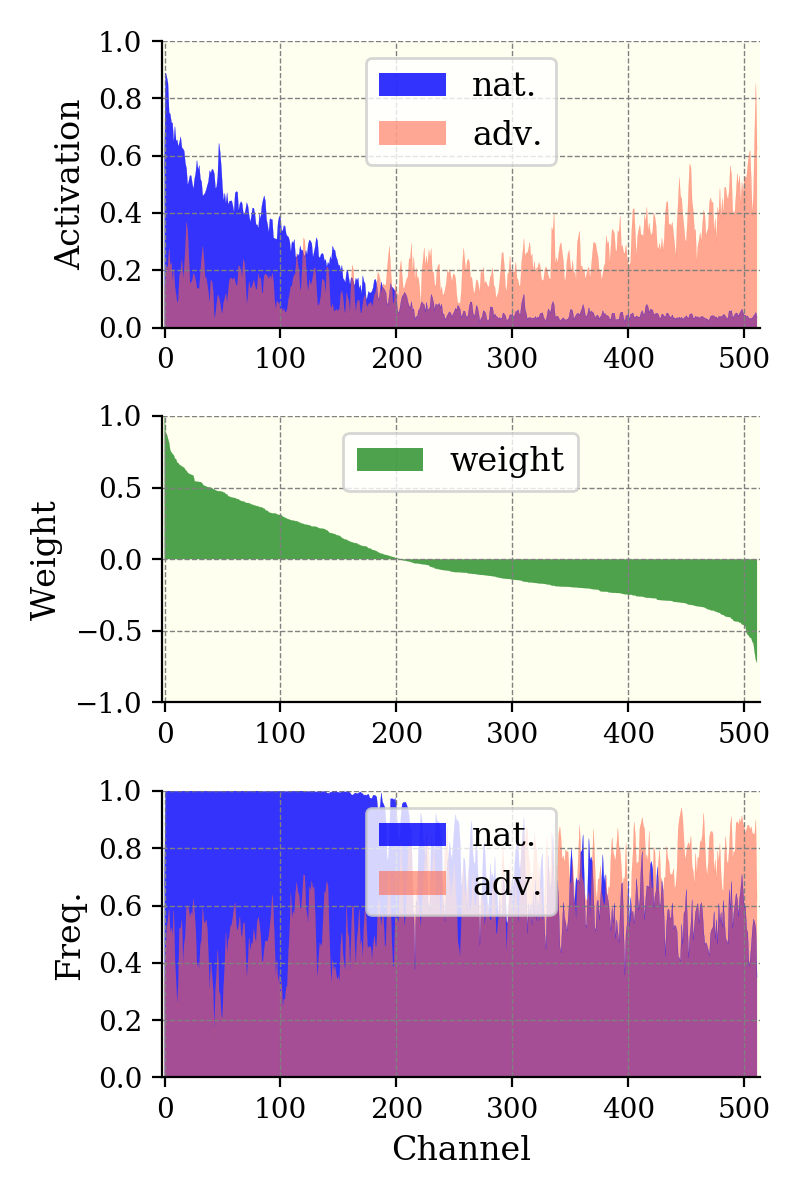}
		\caption{{  Normal / ``automobile''}}
		\label{fig: w-vs-act_normal-auto}
	\end{subfigure}
	\begin{subfigure}{\SubFigWidth\linewidth}
		\includegraphics[width=\SubImgWidth \linewidth]{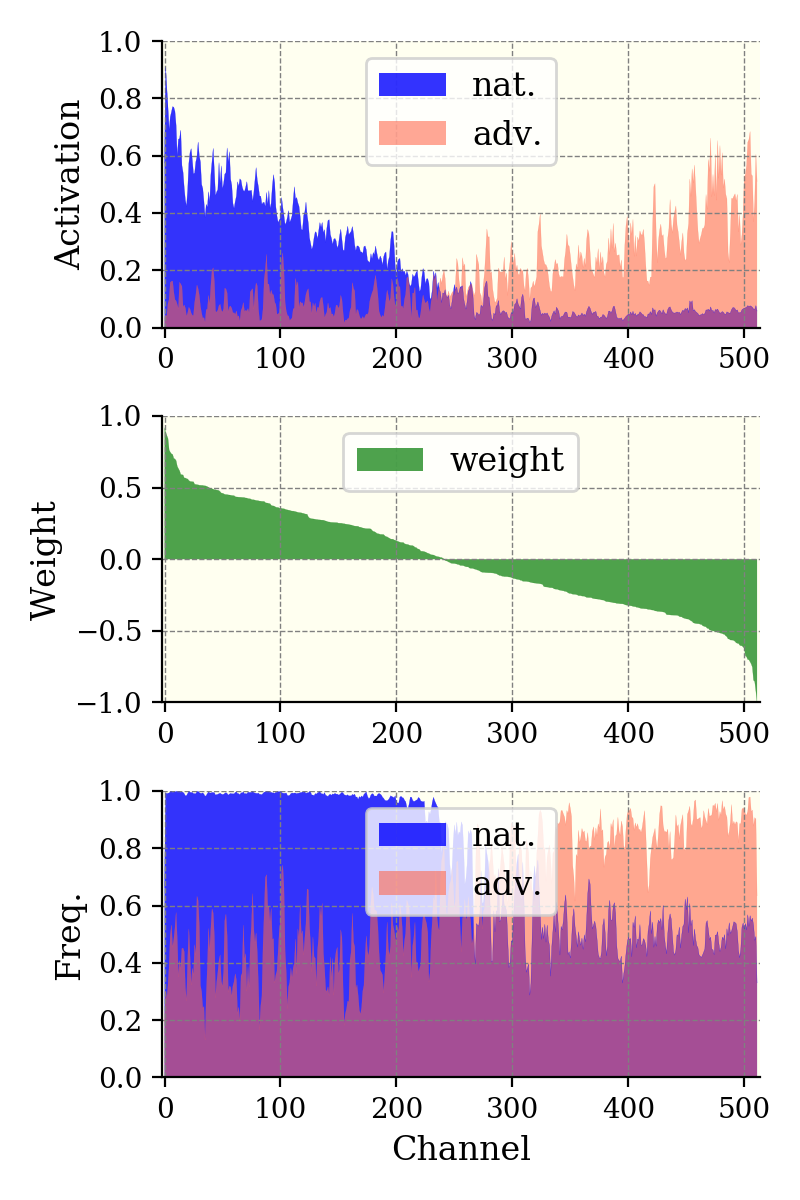}
		\caption{  Normal / ``cat''}
		\label{fig: w-vs-act_normal-cat}
	\end{subfigure}

	\begin{subfigure}{\SubFigWidth\linewidth}
		\includegraphics[width=\SubImgWidth \linewidth]{./Figs/icml2021/hyp/PAT_Res18_cifar10___sorted_weight_vs_act_automobile}
		\caption{  Adv. / ``automobile''} 
		\label{fig: w-vs-act_adv-auto}
	\end{subfigure}
	\begin{subfigure}{\SubFigWidth\linewidth}
		\includegraphics[width=\SubImgWidth \linewidth]{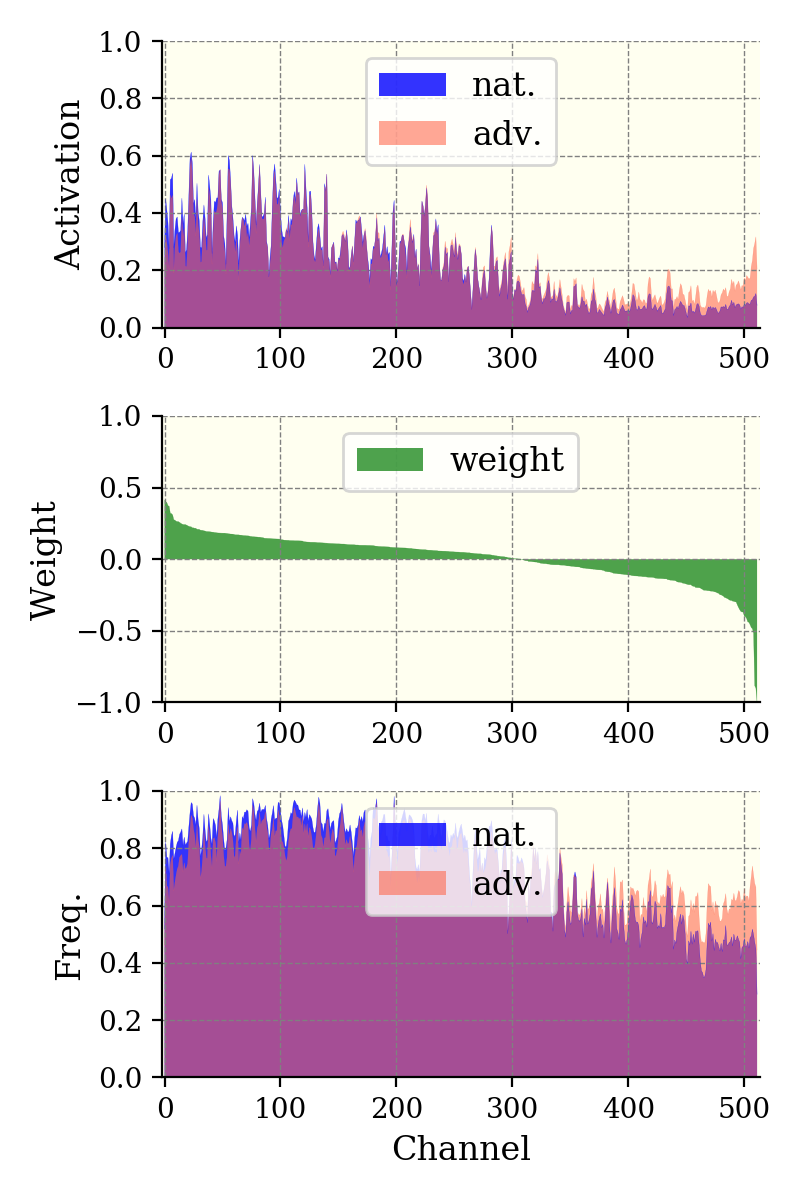}
		\caption{  Adv. / ``cat''}
		\label{fig: w-vs-act_adv-cat}
	\end{subfigure}
	\caption[Comparison of channel activations of adversarial data between normally and adversarially trained models.]{The magnitudes of channel-wise activations (top) at the penultimate layer, their activated frequencies (bottom), and the weights of the last linear layer (middle) \textit{vs.} channel indices. \#/$*$ means the CNN is trained in the \# way and we plot the average activations of data from the class $*$.
		\tit{The weights corresponding to class $*$ are sorted in the descending order and can indicate channels' relevances to class} $*$.
		The activation magnitudes and the activated frequencies of natural and adversarial (PGD-20) data are plotted according to the indices of the sorted weights.
		The robust accuracies on the whole CIFAR10 are 0\% for the normally trained ResNet-18 model and 46.6\% for the adversarially trained one (69.0\% for the ``automobile'' class and 16.7\% for the ``cat'' class). }
	\label{fig: w-vs-act}
\end{figure}

\section{Related Works}
This section briefly reviews relevant adversarial defense methods from two perspectives: adversarial training (AT)-based defense and robust network architecture design.

\paragraph{AT-based Defense} Adversarial training (AT) defends against adversarial attacks by utilizing adversarially generated data in model training \citep{goodfellow_explaining_2015}, formulated as a minimax optimization problem. Recently variants of AT \citep{cai_curriculum_2018,Wang_Xingjun_MA_FOSC_DAT,wang_improving_2020,wu2020adversarial,zhang_geometry-aware_2020} have been proposed. For example, the Misclassification-Aware-AdveRsarial-Training \citep{wang_improving_2020} modifies the process of generating adversarial data by simultaneously applying the misclassified natural data, together with the adversarial data for model training. Recent works have shown AT robustifies CNNs but degrades the natural accuracy \citep{tsipras_robustness_2018, zhang_theoretically_2019, lamb_interpolated_2019}. To achieve a better trade-off, \citet{zhang_theoretically_2019} decomposed the adversarial prediction error into the natural error and boundary error and proposed TRADES to control both terms at the same time. Besides, inspired by curriculum learning \citep{cai_curriculum_2018, bengio_curriculum_2009}, \citet{zhang_attacks_2020} proposed FAT to train models with increasingly adversarial data, which enhances generalization without sacrificing robustness. 


\paragraph{Robust Network Design} Other than robust training strategies, some works explored robust network architectures \citep{yan_robustness_2020, hsieh_robustness_2019}. For instance, the work by \citet{yan_robustness_2020} showed neural ODE-based models are inherently more robust than conventional CNN models;  \citet{guo_sparse_2018} demonstrated that appropriately designed higher model sparsity implies better robustness of nonlinear networks. Another line of works defended against adversarial attacks via gradient obfuscation, such as random or non-differentiable image/feature transformations \citep{xie_mitigating_2018, du_rain_2020, dhillon_stochastic_2018, xiao_enhancing_2019}. However, they have been shown to be insecure to adaptive attacks \citep{athalye_obfuscated_2018, tramer_adaptive_2020}. Recently, many researchers have attempted to develop novel mechanisms for robustness enhancement. \citet{xie_feature_2019} performed feature denoising to remove the adversarial effects on feature maps. \citet{zoran_towards_2020} utilized the spatial attention mechanism to identify highlight important regions of feature maps. Most of these works manipulated CNNs' intermediate representations in the \tit{spatial domain}, whereas our work studies the adversarial robustness from the channel-wise activation perspective.

Channel-wise Activation Suppressing (CAS) \citep{bai_improving_2021}, the most relevant work to ours, also studied the channel-wise activations of adversarial data. It showed channels are activated more uniformly by adversarial data compared to the natural ones, and AT improves the robustness by attempting to align the distributions of channels' activations of natural and adversarial data. However, there are still some channels that are over-activated by adversarial data. To suppress these channels, the authors proposed CAS to adjust channels' activations based on their importance. Although CAS empirically suppresses certain channels, the authors did not show that the suppressed channels correspond to the target ones; this means the primary objective of CAS may not have been met. Thus, there is no guarantee CAS can enhance the robustness of CNNs (see Section \ref{sec:exp-eva} for further evidence on this). \tit{In contrast}, our work first builds a connection between robustness and channels' activations via their relevances to predictions. Then, the proposed CIFS can explicitly control channels' activations based on their relevances. Finally, experiments demonstrate the effectiveness of CIFS on robustness enhancement.

\section{Channel-wise Importance-based Feature Selection }

In this section, we first study the adversarial robustness by comparing channels' activations of non-robust (normally trained) and robustified (adversarially trained) CNNs. Based on our observations of AT's effects, we propose a hypothesis on robustness enhancement via the adjustment of channels' activations (Section \ref{sec:cifs-adv-effects}). To examine this hypothesis, we then develop a novel mechanism, CIFS (Section \ref{sec:cifs-mechanism}), to manipulate channels' activation levels according to their relevances to predictions . Finally, we verify the proposed hypothesis through extensive experiments (Section \ref{sec:cifs-verify}). 

\subsection{Non-robust CNNs vs. Robustified CNNs: A Channel-wise Activation Perspective} \label{sec:cifs-adv-effects}


We compare a non-robust ResNet-18 \citep{he_deep_2016} model with an AT-robustified one on the CIFAR10 dataset \citep{krizhevsky2009learning}. 
In ResNet-18, the representations of penultimate layer are spatially averaged   for each channel, then fed into the last linear layer for making predictions. Thus, the weights of the last linear layer indicate channels' relevances to predictions (according to the definition of channels' relevances in Introduction).
We visualize the channel-wise activation magnitudes, the activated frequencies (counted via a threshold of 1\% of the largest magnitude among all channels) in the penultimate layer for both natural and adversarial data, as well as the weights of the last linear layer in Figure~\ref{fig: w-vs-act}. The details of implementation are provided in Appendix~\ref{apdx:visualization-normal-at}.

From Figures~\ref{fig: w-vs-act_normal-auto} and \ref{fig: w-vs-act_normal-cat}, we observe, for a non-robust ResNet-18, the activation distribution of the adversarial data is obviously mismatched with that of the natural data: natural data activate channels that are PR to predictions with high values and high frequency, while adversarial data tend to amplify the NR ones.
From Figures~\ref{fig: w-vs-act_adv-auto} and \ref{fig: w-vs-act_adv-cat}, we observe that AT robustifies the model by aligning the activation distribution of adversarial data with that of natural data. Specifically, when dealing with adversarial data, AT boosts the activation magnitudes of PR channels while suppressing the activations of NR ones. However, we observe that, for many NR channels (e.g., around 150 channels from $350^{\text{th}}$ to $512^{\text{th}}$), the activations of adversarial data are much higher than those of natural data. These over-activations decrease the prediction scores corresponding to their true categories. Given this observation, we wonder \textcolor{OrangeRed}{(Q1)}: \textit{if we suppress these NR channels to regularize the freedom of adversarial perturbations, will it further improve the model's robustness upon AT}?

Besides, an adversarially trained model does not enjoy similar robustness for all classes, i.e., the robust accuracy of a certain class may be much higher than another (e.g., ``automobile'' with 69.0\% vs. ``cat'' with 16.7\% against PGD-20). Comparing the activations of these two classes (Figures~\ref{fig: w-vs-act_adv-auto} and \ref{fig: w-vs-act_adv-cat}), we observe that, for the class with strong robustness (e.g., ``automobile''), channels' activations align better with their relevances to labels, i.e., the channel with a greater extent of activation usually corresponds to a larger weight in the linear layer. In contrast, this alignment does not hold for the class with relatively poor robustness (e.g., for class ``cat'', the most activated channels, lying between the $26^{\text{th}}$ and $125^{\text{th}}$, are sub-PR to predictions).
Given this phenomenon, we may ask another question \textcolor{OrangeRed}{(Q2)}: \textit{{If we scale channels' activations based on their relevances to predictions, will it improve the model's robustness?}}

Considering the two questions above, we propose the unified hypothesis $\mathcal H$, as stated in the Introduction.

\subsection{Importance-based Channel Adjustment} \label{sec:cifs-mechanism}

To examine the hypothesis $\mathcal H$, one needs a systematic approach to manipulate the channels, viz. selecting channels via suppressing NR ones but promoting PR ones. To this end, we introduce a mechanism, dubbed as Channel-wise Importance-based Feature Selection (CIFS). CIFS modifies layers of CNNs by adjusting channels' activations with importance scores that are generated from the channels' relevances to predictions. 

\begin{figure}[!ht]
	\centering
	\includegraphics[width=0.7\linewidth]{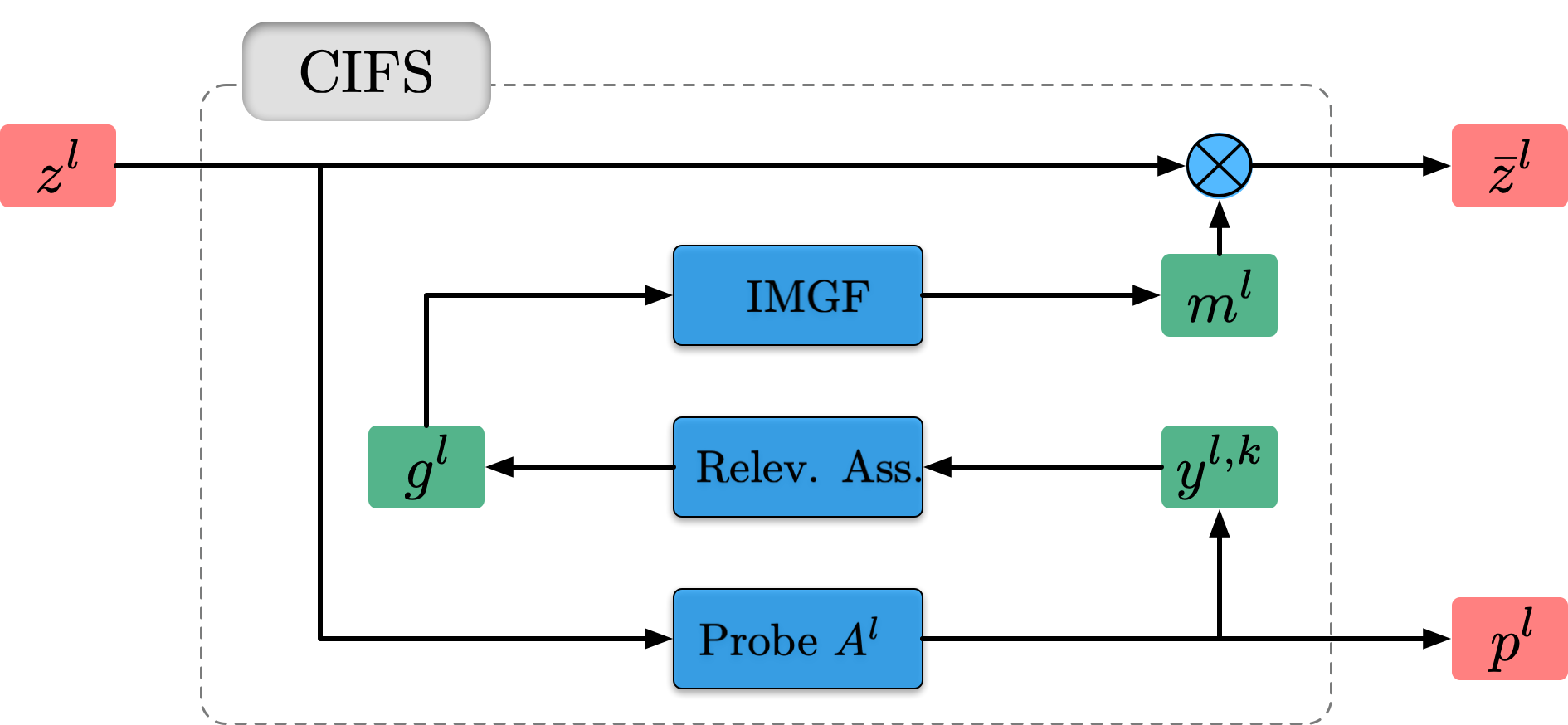}
	\vspace{.5em}
	\caption[Illustration of CIFS.]{CIFS: \tit{1)}~Probe Network $A^l$ first makes a raw prediction $\vp^l$ for $\vz^l$. \tit{2)}~Channels' relevances $\vg^l$ are assessed (Relev.~Ass.) based on the gradients of the top-$k$ prediction results $y^{l,k}$. \tit{3)}~The IMGF generates an importance mask $\vm^l$ from $\vg^l$ for channel adjustment.
	}
	\label{fig:CIFS}
\end{figure}

For clearly state CIFS, we first introduce some notations: For a $K$-category classification problem, let $(\rvx,\rvy) \sim P_{\rvx \rvy} $ denote the pair of the random input and its label, where $\rvx \in \sX \subset \sR^{n_X}$ and $\rvy \in \sY = \{0,1,\dots,K-1\}$. We design an $L$-layer CNN-based classification model to make accurate predictions for data sampled from $P_{\rvx \rvy}$. The $l^{\text{th}}$ layer is denoted by $f^{l}(\cdot)$ and parametrized by $\theta^{l} \in \Theta^l$; the mapping from the input to the $l^{\text{th}}$ layer's output is denoted by $f^{[l]}=f^{l}\circ f^{l-1}\circ \cdots \circ f^{1}$ and the combination of all the first $l$ layers' parameters is denoted by $\theta^{[l]}$, i.e., $\theta^{[l]}=(\theta^1, \dots, \theta^l)$.
Let us examine the $l^{\text{th}}$ layer where an input $\vx$ is transformed into a high-dimensional representation $\vz^l=f^{[l]}(\vx) \in \sR^{n^l_{\text{C}} \times n^l_{\text{F}}}$; $\vz^l$ has $n^l_{\text{C}}$ \underline{c}hannels and each channel is a \underline{f}eature vector of length $n^l_{\text{F}}$. With these notations, we elaborate the details of CIFS in \tit{\tbf{{three}}} steps (as shown in Figure \ref{fig:CIFS}).

\textcolor{Blue}{1) Surrogate Raw Prediction:}~To assess channels' relevances to predictions, 
a naive strategy is to compute the gradients of the final prediction with respect to $\vz^l$, i.e., $\nabla_{\vz^l} f^{[l+1:L]}(\vz^l)$, where $f^{[l+1:L]}=f^L \circ \cdots \circ f^{l+1}$. Since we need to adjust $\vz^l$ with importance scores generated from $\nabla_{\vz^l} f^{[l+1:L]}(\vz^l)$ and send the adjusted feature $\bar \vz^l$ to $f^{[l+1:L]}$ again for making the final prediction, it will result in computing the second-order derivatives during the training phase. Moreover, in practice, we may apply CIFS into multiple layers,
the forward pass will involve at least the second-order gradients (the latter CIFS-modified layer is recursively called). Thus, the back-propagation has to deal with at least the third-order gradients during training. This will aggravate the problem of training instability. 

Instead, inspired by the design of auxiliary classifiers in CAS \citep{bai_improving_2021}, CIFS builds a probe network $A^l$ as the surrogate of $f^{[l+1:L]}$ for a making raw prediction $\vp^l = A^l(\vz^l)$, so that we can use the gradients of $\vp^l$ to approximately assess the channels' relevances to the final prediction. The assessment does not involve other CIFS-modified layers. Thus, we can avoid the problem of back-propagation through high-order derivatives. The probe network $A^l$ is parameterized by $\theta^l_A$ and $\vp^l\in \sR^K$ represents the vector of prediction scores/logits. We can jointly optimize $A^l$ with the backbone network during the training phase under the supervision of true labels. 

\textcolor{Blue}{2) Relevance Assessment:}~
With the prediction $\vp^l$, we can compute the gradients of logits in $\vp^l$ \tit{w.r.t.} $\vz^l$ to assess the feature's relevances to each class.
We consider the top-$k$ prediction results ($k\geq 2$) for the assessment of channels' relevances. As data from two semantically similar classes (e.g., ``dog'' and ``cat'') usually share common features, the prediction for an input often assigns large scores to the classes similar to the true one and the top-$k$ results may include several of these similar classes \citep{jia_certified_2020}. In case the top-$1$ prediction is wrong, considering the top-$k$ results may help us reliably extract some common relevant features (see Section \ref{sec:exp-ablation} for more evidence). 

Let $y^{l,k}$ denote indices of the $k$ largest logits of prediction $\vp^l$. 
Let $\vdelta \in \sR^{n^l_{\text{C}}}$ be the channel-wise perturbation added to $\vz^l$, giving the perturbed representation $\vz^l_{\vdelta} = \vz^l + \vdelta \cdot \mathbf{1^{\top}}$. Here $\mathbf{1} \in \sR^{n^l_{\text{F}}}$ is the column vector with all elements as one, i.e., the features in the same channel are perturbed by a common value. We calculate the gradients of the sum of the top-$k$ logits \textit{w.r.t.} the channel-wise perturbation $\vdelta$:
\begin{equation}
	\small
	\vg^l 
	= \left.\nabla_{\vdelta} \sum_{i \in y^{l,k}} \vp^l(\vdelta)_{[i]} \right\vert_{\vdelta=\mathbf{0}} 
	= \left.\nabla_{\vz^l_{\vdelta}} \sum_{i \in y^{l,k}} A^l(\vz^l_{\vdelta})_{[i]} \right\vert_{z^l_{\vdelta} =z^l} \cdot \mathbf{1},
	\label{eq:relevance-assessment}
\end{equation}
where $\vg^l=(\vg^l_{[0]},...,\vg^l_{[n^l_{\text{C}}-1]})$
represents the vector of channels' relevances to the top-$k$ logits. During the training phase, since the true label of $\vz^l$ is given,  we replace the top-$1$  prediction in $y^{l,k}$ with the true label $y$ and keep other prediction results untouched. 

\textcolor{Blue}{3) Importance Mask Generation:}~
As we want to suppress or promote channels based on their relevances, we need to design proper Importance Mask Generating Functions (IMGFs), which monotonically map relevances to \textit{non-negative} importance scores; of particular importance is to map negative relevances to values close to zero.

Here, we provide several feasible options: To answer the first question on whether suppressing NR channels enhances robustness, one can use the sigmoid function as the IMGF. With a large value of $\alpha$, the sigmoid function serves as a switch by mapping negative relevances to importance scores  close to zero and the positive close to one. 
To answer the second question concerning aligning channel activations with their relevances, we can use the softplus or softmax function as the IMGF. Both of them can map negative relevacnes to values close zero and map positive relevances monotonically to positive values.
These three functions are stated here for ease of reference:
\begin{itemize}
	\item sigmoid: $\vm^l_{[i]} = \frac{1}{1+\exp(-\alpha \cdot \vg^l_{[i]})}, \quad \alpha>0$. 
	\item softplus: $\vm^l_{[i]} = \frac{1}{\alpha} \cdot \log(1+\exp (\alpha \cdot \vg^l_{[i]})), \quad \alpha>0$.
	\item softmax: $\vm^l_{[i]} = \frac{\exp(\vg^l_{[i]}/T)}{\sum_{j} \exp(\vg^l_{[j]}/T)}, \quad T>0$.
\end{itemize}
The usage of these functions will be discussed in detail in Section~\ref{sec:cifs-verify}. CIFS selects channels by multiplying the importance mask $\vm^l$ with $\vz^l$ as follows:
\begin{equation}
	\small \bar \vz^l = \vz^l \otimes \text{repmat}(\vm^l, 1, n^l_L),
\end{equation} 
where the ``$\text{repmat}$'' operation replicates the column vector $\vm^l$ along the second axis $n^l_L$ times.

\paragraph{Training of CIFS} In practice, we may apply the CIFS mechanism into several layers of a CNN. Let $I$ denote the set of indices of these layers, and $\theta_A^{I}$ denote the parameters of all the probes in the CIFS-modified layers. For each input $\vx$, the modified model $\bar f^{[L]}$ outputs $|I|$ raw predictions and one final prediction $\vp=\bar f^{[L]}(\vx)$. Given this, we use an adaptive loss function \citep{bai_improving_2021} for training the model: 
\begin{equation}
	\small
	\ell_{\beta}(\vx,\vy) = \frac{1}{1+\beta} \cdot \ell_{\mathrm{ce}}(\vp,\vy) + \frac{\beta}{(1+\beta)|I|} \cdot {\sum\limits_{l\in I} \ell_{\mathrm{ce}}(\vp^l, \vy)},
	\label{eq:adaptive-loss}
\end{equation}
where $\ell_{\text{ce}}(\cdot, \cdot)$ denotes the cross-entropy loss and the coefficient $\beta>0$ balances the accuracy of raw predictions by CIFS and the final prediction. Since the subsequent decisions closely depend on the channels of features selected by the previous CIFS-modified layers, we should choose a proper value of $\beta$ to make sure that the raw predictions made by CIFS are reliable. In practice, we set $\beta$ to be $|I|$, and the effect of $\beta$ is discussed in the ablation study (see Appendix \ref{apdx:ablation}). To robustify the CNN model against malicious attacks, we can train $\bar f^{[L]}$ in an adversarial manner with a perturbation budget $\epsilon$. Namely, we solve the following optimization problem:  
\begin{equation}
	\small
	\min_{\theta^{[L]}, \theta^{I}_A} \E_{(\rvx, \rvy)\sim P_{\rvx \rvy}} \left[ \max_{\rvx'\in \mathcal{B}(\rvx, \epsilon, l_{\infty})} \ell_{\beta}(\rvx',\rvy) \right],
	\label{eq:adv_cifs}
\end{equation}
where $\mathcal{B}(\vx, \epsilon, l_{\infty})=\{\vx' ~|~ \|\vx'-\vx\|_{l_{\infty}} \leq \epsilon \}$. 

\def \SubFigWidth {0.35} 
\def \SubImgWidth {.99}
\begin{figure*}[!ht]
    \centering
    \begin{subfigure}{\SubFigWidth\linewidth}
        \centering
        \includegraphics[width=\SubImgWidth \linewidth]{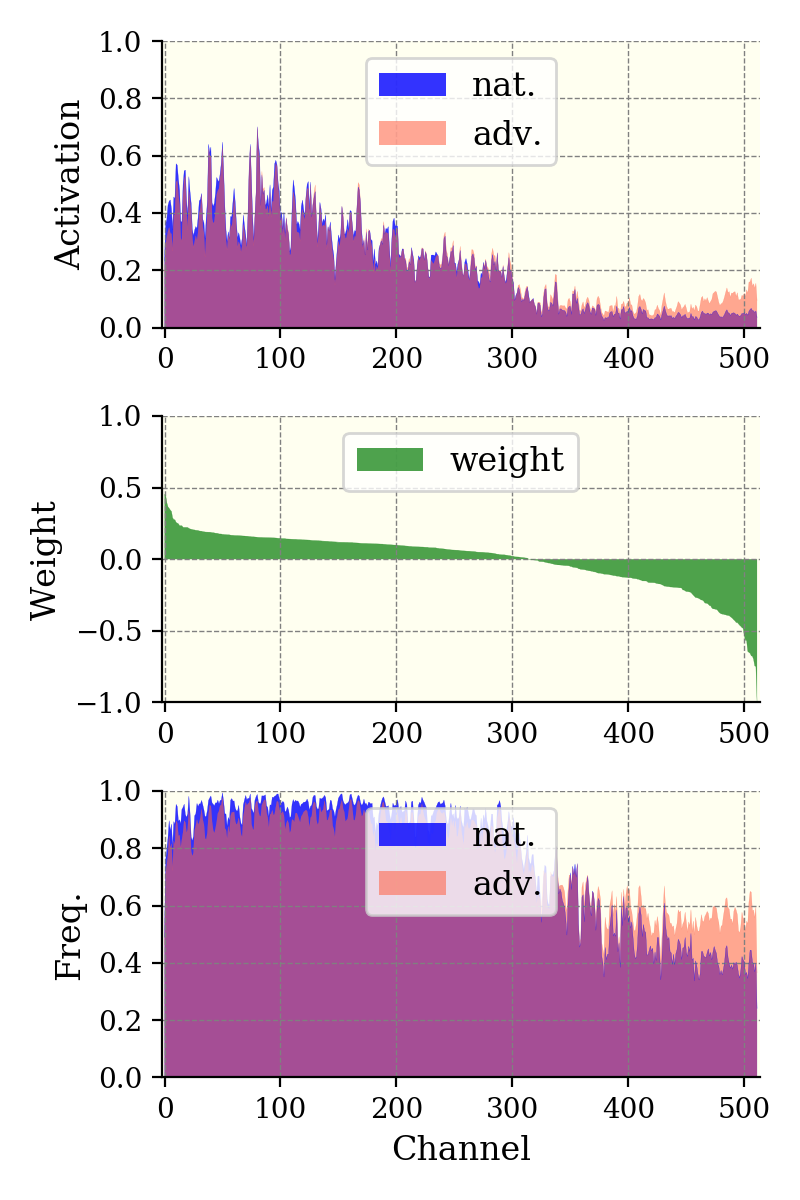}
        \caption{  non-CIFS}
        \label{fig:hyp-1-non-CIFS}
    \end{subfigure}
    \begin{subfigure}{\SubFigWidth\linewidth}
        \centering
        \includegraphics[width=\SubImgWidth \linewidth]{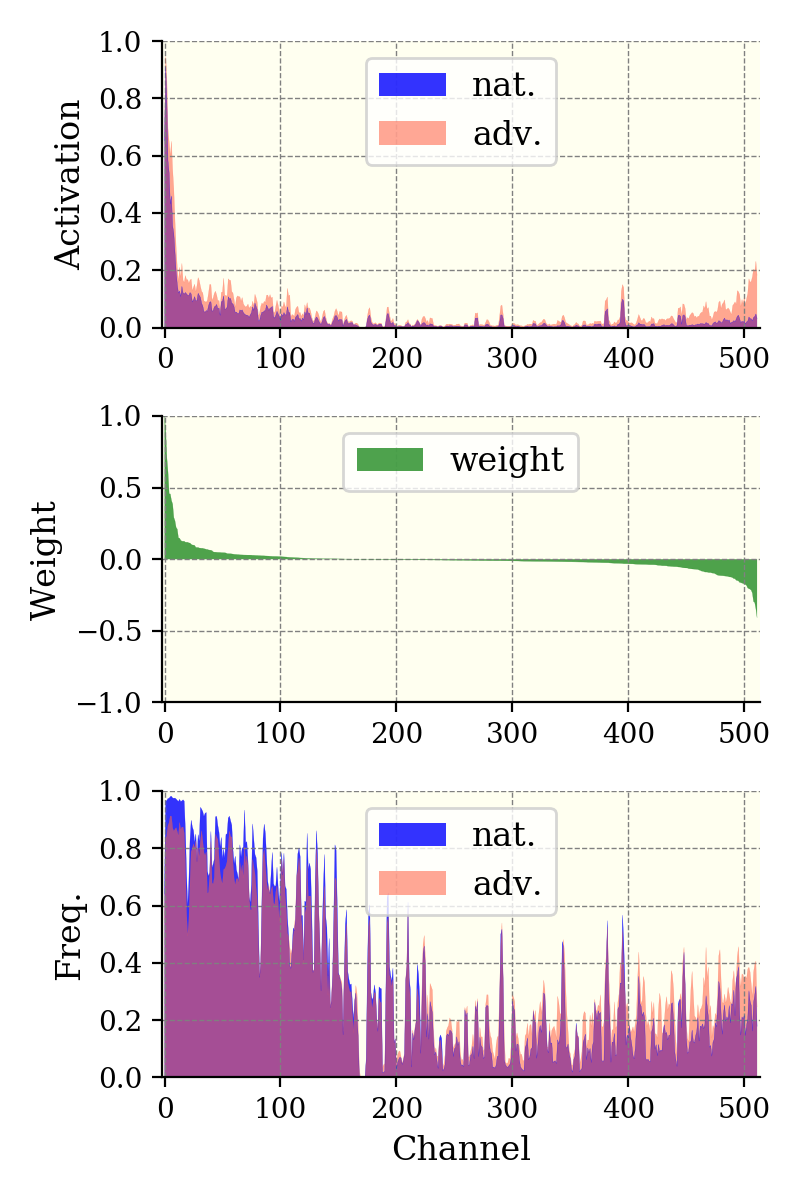}
        \caption{  CIFS-sigmoid}
        \label{fig:hyp-1-sigmoid}
    \end{subfigure}

    \begin{subfigure}{\SubFigWidth\linewidth}
        \centering
        \includegraphics[width=\SubImgWidth \linewidth]{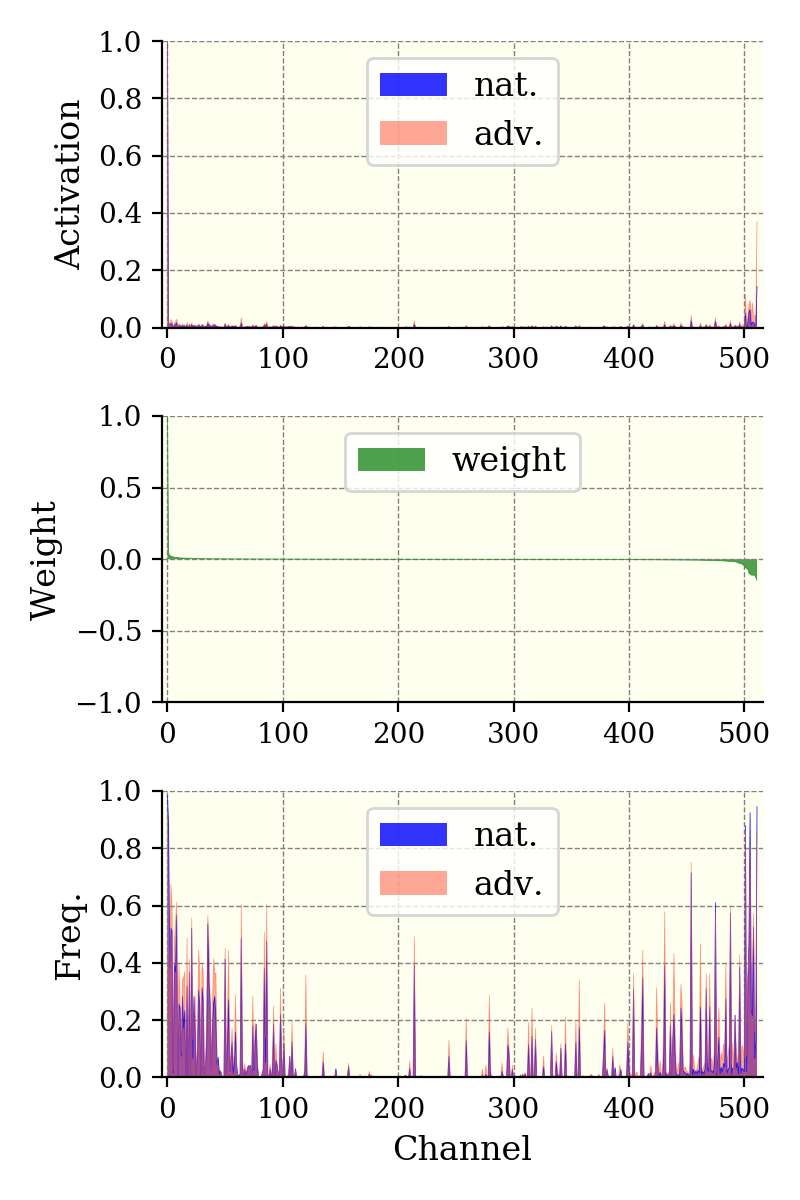}
        \caption{  CIFS-softplus}
        \label{fig:hyp-2-softplus}
    \end{subfigure}
    \begin{subfigure}{\SubFigWidth\linewidth}
        \centering
        \includegraphics[width=\SubImgWidth \linewidth]{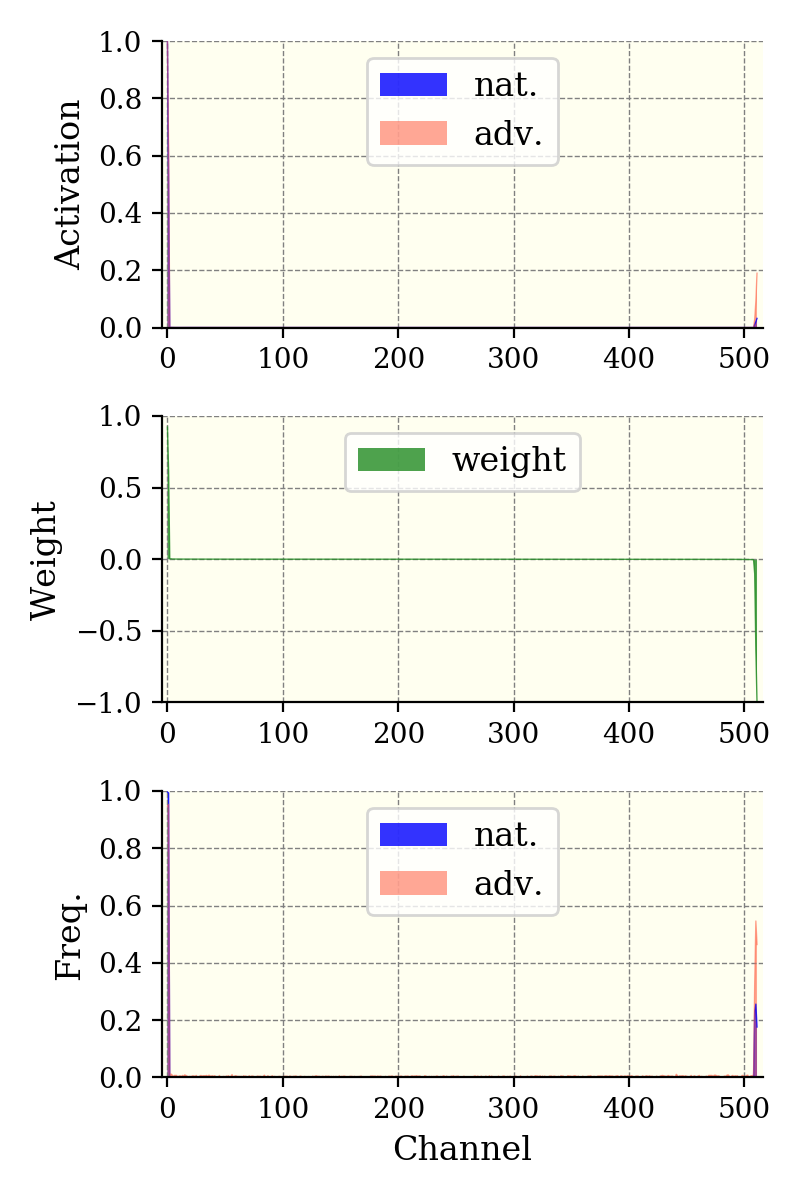}
        \caption{  CIFS-softmax}
        \label{fig:hyp-2-softmax}
    \end{subfigure}
    \caption[Comparison of channel activations of adversarial data between non-CIFS and CIFS modified models.]{The magnitudes of channel-wise activations (top) at the penultimate layer, their activated frequency (bottom), and the weights of the last linear layer (middle) \textit{vs.} channel indices. The activations of natural data and their PGD-20 examples are averaged over all test samples in the ``airplane'' category. The robust accuracies against PGD-20 (on the whole dataset) are 46.64\% for non-CIFS, 49.87\% for the CIFS-sigmoid, 50.38\% for the CIFS-softplus, and 51.23\% for the CIFS-softmax respectively}
    \label{fig:hyp}
\end{figure*}

\subsection{Verification of Hypothesis  \texorpdfstring{$\mathcal H$}{hyp-H} on Robustness Enhancement} \label{sec:cifs-verify}
We verify the hypothesis $\mathcal H$ by answering the two questions, \textcolor{OrangeRed}{Q1} and \textcolor{OrangeRed}{Q2}, in Section~\ref{sec:cifs-adv-effects} respectively.


To answer \textcolor{OrangeRed}{Q1}, we applied the sigmoid function to generate the mask $\vm^l$ from $\vg^l$. Setting $\alpha$ to be large enough (here, $\alpha=10$), we can generate importance scores close to zeros for NR channels and scores close to one for the PR ones, so that we approximately annihilate the NR channels but leave the PR ones as unchanged. We adversarially trained a ResNet-18 model and its CIFS-modified version. As shown in Figure~\ref{fig:hyp-1-non-CIFS}, the NR channels (Channel 300-512) of the vanilla ResNet-18 model are clearly activated (the average activation magnitudes of these channels are larger than 0.1; the activation frequencies are over 0.4). In contrast, the ResNet-18 with CIFS-sigmoid effectively suppresses the activation of NR channels (Channel $100^{\text{th}}$--$512^{\text{th}}$ in Figure~\ref{fig:hyp-1-sigmoid}). Most of their mean activation magnitudes are smaller than 0.05, and their activation frequencies have clearly decreased. The experimental results show that, under AT, the vanilla ResNet-18 model results in a \underline{46.64\%} defense rate against the PGD-20 attack while its CIFS-sigmoid modified version achieves \underline{49.87\%}. More results can be found in Appendix \ref{apdx:visualization-cifs-modification}. Thus, we conclude that suppressing NR channels enhances the robustness of CNNs. 

To answer \textcolor{OrangeRed}{Q2}, we applied the softplus and softmax functions as IMGFs to generate the mask $\vm^l$ from $\vg^l$ respectively. Here the coefficient $\alpha$ in the softplus is set to be $5$ and the temperature $T$ in the softmax is set to be $1$. 
From Figures~\ref{fig:hyp-2-softplus} and \ref{fig:hyp-2-softmax}, we observe that, by generating importance scores positively correlated with the relevances, the model tends to completely focus on few relevant (positive and negative) channels. The channel of the greatest weight (most PR to predictions) is activated with the highest magnitude.
Most channels become irrelevant (small absolute values of weights) to the predictions and are activated at a low level. 
Using softplus as the IMGF (Figure~\ref{fig:hyp-2-softplus}), 
the irrelevant channels are sparsely activated, and the activation magnitudes are smaller than 5\% of the most important channel. 
In Figure~\ref{fig:hyp-2-softmax}, this phenomenon is enhanced by using softmax as IMGF: most channels become irrelevant to predictions and are usually deactivated. We evaluated the robustness of these two CIFS-modified CNNs against PGD-20 attack. Both of them outperformed the CIFS-sigmoid and the vanilla ResNet-18 classifiers (robust accuracies of CIFS-softplus and CIFS-softmax are \underline{50.38\%} and \underline{51.23\%} respectively, \tit{vs.},  \underline{49.87\%} for CIFS-sigmoid and \underline{46.64\%} for the vanilla ResNet-18). We also found CIFS can ameliorate the class-wise imbalance of adversarial robustness (\tit{e.g.}, CIFS-softmax increases the PGD-20 accuracy from 16.7\% to 22.3\% for class ``cat''). More details are provided in Appendix \ref{apdx:visualization-cifs-modification}. Thus, we conclude that aligning channels activations with their relevances to predictions can further robustify CNNs upon suppressing NR ones.

Given these empirical results, we verified the hypothesis $\mathcal H$ and justified that CIFS is an effective mechanism to improve the adversarial robustness of CNNs. In the following section, we conduct extensive experiments to evaluate the robustness enhancement through CIFS and study CIFS in an ablation manner.

\section{Experiments} \label{sec:exp}

\subsection{Robustness Evaluation} \label{sec:exp-eva}
We utilize the CIFS to modify CNNs in different architectures to perform classification tasks on benchmark datasets, namely a ResNet-18 and a WideResNet-28-10 on the CIFAR10 \citep{krizhevsky2009learning} dataset, a ResNet-18 on the SVHN \citep{netzer2011reading} dataset and 
a ResNet-10 on the Fashion-MNIST \citep{xiao2017_online} dataset.  We train the models with the standard PGD adversarial training (AT) \citep{madry_towards_2018} and its variants, such as FAT \citep{zhang_attacks_2020}, to show that CIFS can work under various AT-strategies.  We compare CIFS-modified CNNs with the vanilla versions as well as the CAS-modifications, where CAS \citep{bai_improving_2021} also modifies CNNs by adjusting channels' activations. Here, we report the results on CIFAR10 and SVHN. Results on FMNIST are presented in Appendix \ref{apdx:fmnist}.

\paragraph{Adaptive Attacks} As mentioned in Section \ref{sec:cifs-mechanism}, each CIFS-modified layer of CNNs outputs a raw prediction. To generate adversarial examples that are as strong as possible, we follow the strategy used in CAS and attack CNNs via the adaptive loss function $\ell_{\beta}$ in Equation (\ref{eq:adaptive-loss}) that considers all the raw and final predictions. We let the value of $\beta$ be chosen by the attacker, i.e., the attacker can try various values of $\beta_{\text{atk}}$ and select one that results in the most harmful perturbations. Our setting is more challenging for defense than CAS where the \tit{same} value of $\beta$ is used for training and attack. Here, for each adversarial attack, we evaluate the robustness by choosing $\beta_{\text{atk}}$ from $\{0, 0.1, 1, 2, 10, 100, \infty\}$ and report the worst robust accuracy\footnote{Results of various values of $\beta$ are present in Appendix \ref{apdx:cifar10-at}. We observe that CAS can improve the robustness of CNNs in most cases but \tit{fail} when attackers completely focus on CAS modules.}. Setting $\beta_{\text{atk}}=\infty$ means the attacks completely focus on the CIFS-modified layers \footnote{For $\beta_{\text{atk}}=\infty$, we consider the cases of attacking both CIFS-modified layers simultaneously and attacking each  separately.} and only consider the second term in Equation (\ref{eq:adaptive-loss}).

\subsubsection{Robustness Enhancement of CIFS under AT}
We adversarially train ResNet-18 and WRN-28-10 models with PGD-10 ($\epsilon=\nicefrac{8}{255}$) adversarial data.
CIFS is applied to the last two residual blocks of each model. The probes for the last and penultimate blocks are a linear layer and a multi-layer perceptron (MLP) respectively. Channels' relevances are assessed based on top-$2$ results and we use the softmax function with $T=1$ as the IMGF.
Other training details are provided in Appendix \ref{apdx:cifar10} and \ref{apdx:svhn}. 

\paragraph{Defense Results} We evaluate the robustness of CNNs against four types of white-box attacks: FGSM \citep{szegedy_intriguing_2014}, PGD-20 \citep{madry_towards_2018}, C\&W \citep{carlini_towards_2017}, and PGD-100. The $l_{\infty}$-norm of the perturbations are bounded by the value of $\epsilon=\nicefrac{8}{255}$. Here, we report the robustness evaluated at the last epoch for each model. Detailed attack settings and more defense results~( AutoAttack\footnote{AutoAttack consists of both white and black-box attacks.}~\citep{croce_reliable_2020} and the best epochs' results), are present in Appendix \ref{apdx:cifar10} and \ref{apdx:svhn}.

The defense results on CIFAR10 are reported in Table \ref{tab:white-box-cifar}. We observe that, for both of the ResNet-18 and WRN-28-10 architectures, CIFS consistently outperforms the counterparts against various types of adversarial attacks. For example, the CIFS-modified WRN-28-10 can defend the PGD100 attack with a success rate of $48.74\%$, which exceeds the second best by more than 4 percentage points. In contrast, under the strong adaptive attack, we see that the baseline CAS cannot improve and even worsens the robustness of CNNs. The defense results on SVHN are reported in Table \ref{tab:white-box-svhn}. The results also verify the  effectiveness of CIFS on improving robustness. 

\begin{table}[h]
    \centering
    \caption[Robustness comparison of defense methods on CIFAR10.]{Robustness comparison of defense methods on CIFAR10. We report the accuracies (\%) for adversarial and natural data. For each model, the results of the strongest attack are marked with an underline.
    }
    \scalebox{.9}{
    \begin{tabular}{cccccc}
    \toprule
    \tbf{\tit{ResNet-18}} & Natural & FGSM & PGD-20 & C\&W & PGD-100 \\
    \hline
    Vanilla & 84.56 & 55.11 & 46.62 & 45.95 & \underline{44.72} \\
    CAS     & 86.73 & 55.99 & 45.29 & 44.18 & \underline{43.22} \\
    CIFS    & 83.86 & \tbf{58.86} & \tbf{51.23} & \tbf{50.16} & \underline{\tbf{48.70}} \\
    \bottomrule
	\toprule 
    \tbf{\tit{WRN-28-10}} & Natural & FGSM & PGD-20 & C\&W & PGD-100 \\
    \hline
    Vanilla & 87.29 & 58.50 & 49.17 & 48.68 & \underline{47.08} \\
    CAS     & 88.05 & 57.94 & 49.03 & 47.97 & \underline{47.25} \\
    CIFS    & 85.56 & \tbf{61.34} & \tbf{53.74} & \tbf{53.20} & \underline{\tbf{51.51}} \\
    \bottomrule
    \end{tabular}
    }
    \label{tab:white-box-cifar}
\end{table}{}

\begin{table}[h]
    \centering
    \caption[Robustness comparison of defense methods on SVHN.]{Robustness comparison of defense methods on SVHN. The accuracies (\%) for natural and adversarial data are reported. 
    }
    \scalebox{.9}{
    \begin{tabular}{cccccc}
    \toprule
    \tbf{\tit{ResNet-18}} & Natural & FGSM & PGD-20 & C\&W & PGD-100 \\
    \hline
    Vanilla & 93.72 & \tbf{65.87} & 50.35 & 47.89 & \underline{45.81} \\
    CAS     & 94.08 & 65.24 & 48.47 & 46.15 & \underline{43.75} \\
    CIFS    & 93.94 & 66.24 & \tbf{52.02} & \tbf{50.13} & \underline{\tbf{47.49}} \\
    \bottomrule
    \end{tabular}
    }
    \label{tab:white-box-svhn}
\end{table}{}

\begin{figure}[h!]
	\centering
	\includegraphics[width=0.7\linewidth]{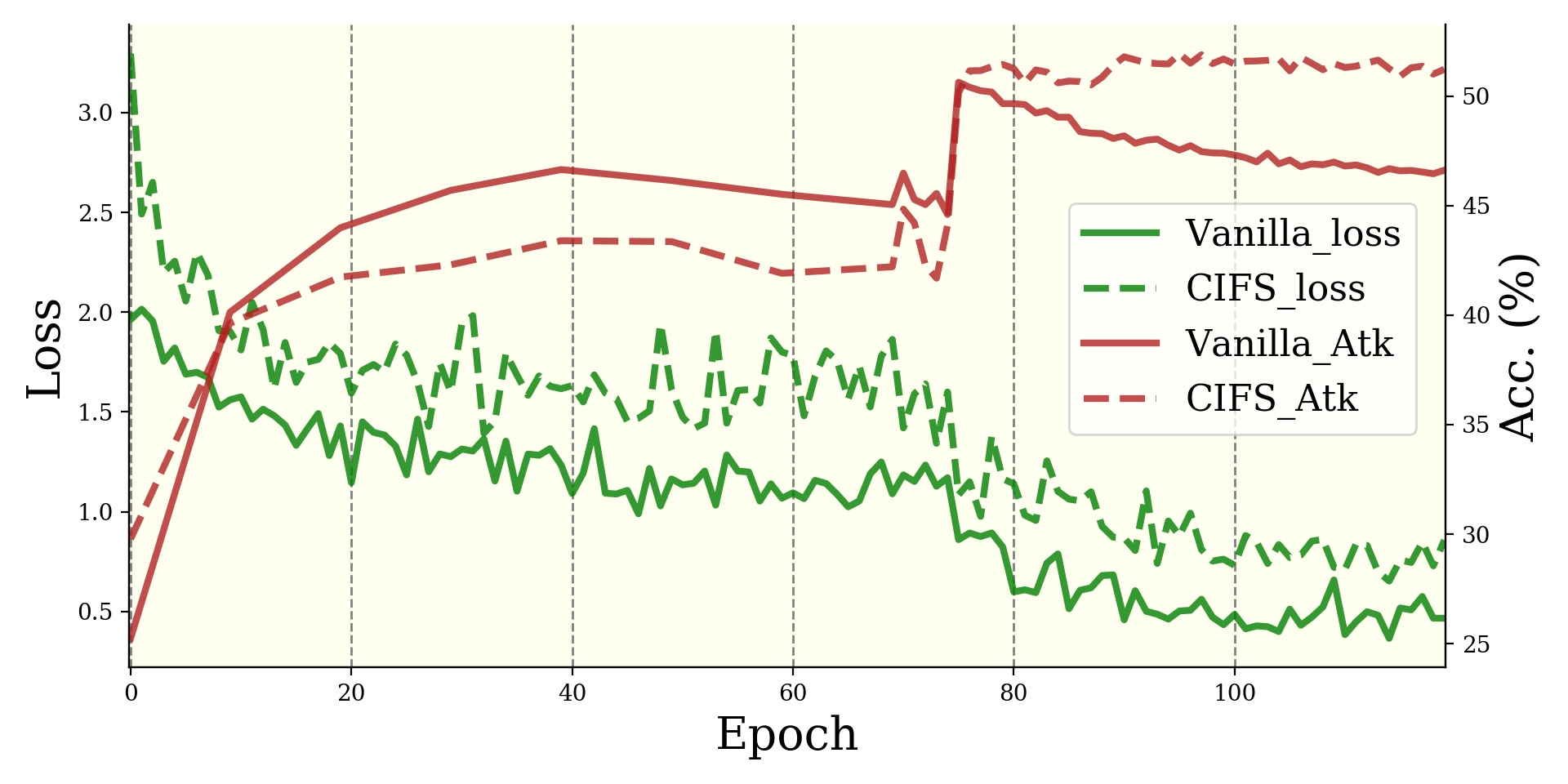}
	\caption[Comparison of training dynamics of non-CIFS and CIFS modified ResNet-18 models.]{Comparison on the training of a vanilla ResNet-18 and its CIFS-modified version. Training losses and accuracies against PGD-20 attack are plotted.}
    \label{fig:tr_procedure}
\end{figure}

\paragraph{Training Procedure} We train CNN classifiers in an adversarial manner for $120$ epochs and adjust the learning rate with a multiplier $0.1$ at epoch $75$ and epoch $90$. We summarize the training procedure by plotting the curves of training losses and the PGD-20 accuracies \tit{w.r.t.} epochs in Figure \ref{fig:tr_procedure}. We observe the best adversarial robustness of the vanilla ResNet-18 ($50.64\%$ defense rate) appears around the $75^{\text{th}}$ epoch. After epoch $75$, the model starts to overfit to training data, i.e., the training loss continues decreasing, but the robust accuracy drops as well. In contrast, the overfitting problem is ameliorated by the application of CIFS. We can see that the best robust accuracy appears around the $90^{\text{th}}$ epoch; After the best epoch, the training loss continues decreasing, but the robustness is maintained around the peak. This phenomenon may result from the fact that CIFS suppresses redundant channels. The model redundancy can be controlled by selecting few highly relevant channels and deactivating others. In this way, the overfitting in training is ameliorated.

On the computation overhead, we report the training time and the evaluation time of a ResNet-18 classifier on the CIFAR10 dataset for reference. For PGD-10 adversarial training, the vanilla CNN takes 166s for each epoch while the CIFS-modified model takes 172s. For the PGD-20 evaluation, the vanilla CNN takes 53s for the CIFAR10 test set; the CIFS-modified model takes 56s instead. In short, the proposed CIFS does not result in too much extra computation. 

\subsubsection{CIFS Working in Conjunction with Variants of AT}
CIFS improves the adversarial robustness by adjusting channels' activations, which is orthogonal to defense training strategies. Here, we train the CIFS-modified CNNs with variants of AT to examine whether CIFS can work in conjunction with other state-of-the-art training-based defense techniques. We consider the FAT \citep{zhang_attacks_2020} and TRADES \citep{zhang_theoretically_2019} strategies. 
We report the defense results of FAT in Table \ref{tab:fat} and provide the results of TRADES in Appendix \ref{apdx:cifar10-trades}. We observe that FAT training strategy improves the natural accuracy and robustness of CNNs (compared to the results in Table \ref{tab:white-box-cifar}). Under the FAT strategy, CIFS also improves the adversarial robustness of the vanilla CNNs in both ResNet-18 and WRN-28-10 architectures.

\begin{table}[ht]
    \centering
    \caption[Robustness comparison of defense methods trained with FAT on CIFAR10.]{Robustness comparison of defense methods trained with FAT on CIFAR10. Comparing the accuracies (\%) against the strongest attacks, we observe that CIFS clearly robustifies CNNs.}
    \scalebox{.9}{
    \begin{tabular}{cccccc}
    \toprule
    \tbf{\tit{ResNet-18}} & Natural & FGSM & PGD-20 & C\&W & PGD-100 \\
    \hline
    Vanilla & 87.16 & 56.43 & 47.64 & 46.01 & \underline{45.35} \\
    CIFS     & 86.35 & \tbf{59.47} & \tbf{51.68} & \tbf{51.84} & \underline{\tbf{49.52}} \\
    \bottomrule
    \toprule
    \tbf{\tit{WRN-28-10}} & Natural & FGSM & PGD-20 & C\&W & PGD-100 \\
    \hline
    Vanilla & 88.37 & 58.81 & 49.62 & 48.49 & \underline{47.58}\\
    CIFS  & 86.74 & \tbf{60.67} & \tbf{51.99} & \tbf{52.34} & \underline{\tbf{49.87}}\\
    \bottomrule
    \end{tabular}
    }
    \label{tab:fat}
\end{table}{}

\subsection{Ablation Study}  \label{sec:exp-ablation}
Here, we conduct an ablation study to further understand the robustness properties of CIFS. Specifically, we investigate the effects of the feedback from the top-$k$ predictions. The ablation experiments are conducted on CIFAR10 based on the ResNet-18 model. Besides, in Appendix \ref{apdx:ablation}, we also study cases in which the CIFS is applied to different layers, the probe networks are in different architectures, various values of $\beta$ are used for training. 

\paragraph{Feedback from Top-$k$ Prediction Results}
As is well-known, the top-$k$ classification accuracy for $k>1$ is always not worse than the top-$1$. 
For example, in Table \ref{tab:topk}, we can see that the top-$2$ accuracy of an adversarially trained ResNet-18 against the PGD-20 attack exceeds the top-$1$ accuracy by $25$ percentage points. This implies that, although adversarial data can usually fool the classifier (i.e., low top-$1$ accuracy), the prediction confidence of the true class is still high and the corresponding score highly likely lies among the top-$2$ or $3$ logits.

\begin{table}[h]
    \centering
    \caption{Top-$k$ accuracies (\%) against adversarial attacks on CIFAR10 of an adversarially trained ResNet-18.}
    \scalebox{.9}{
    \begin{tabular}{cccccc}
    \toprule
    \tbf{\tit{ResNet-18}} & top-$1$ & top-$2$ & top-$3$\\
    \hline
    FGSM & 55.11 & 76.22 & 85.20 \\
    PGD-20 & 46.62 & 71.71 & 81.60 \\
    \bottomrule
    \end{tabular}
    }
    \label{tab:topk}
\end{table}{}

CIFS generates the importance mask from the raw prediction and uses it to suppress or promote channels at the current layer. The final prediction made by subsequent layers strongly depends on the channels selected by CIFS. To ensure the accuracy of final predictions, the logits used for generating importance scores should include the true label's logit for each input so that the truly important channels will be highlighted. According to Table \ref{tab:topk}, if we use the top-$1$ logit to assess the importance of channels for PGD-20 adversarial data, the probability of incorrect assessment is over $50$\%. Instead, if we use the feedback from top-$2$ or $3$ logits, the truly important channels can highly likely be promoted. The following table presents more experiments that justify this argument. 

\begin{table}[h]
    \centering
    \caption[Robustness comparison (\%) of importance assessment based on top-$k$ results.]{Robustness comparison (\%) of importance assessment based on top-$k$ results against the PGD-20 attack. The column header $*/\#$ represents the attack point and the output prediction. For example, CIFS/Final means the model outputs the final prediction and the attacker solely focuses on the CIFS's raw prediction.}
    \scalebox{.9}{
    \begin{tabular}{cccccc}
    \toprule
    \tbf{\tit{ResNet-18}} & Natural/Final & CIFS/CIFS & CFIS/Final & Adap/Final\\
    \hline
    Vanilla & 84.56 & - & - & 46.62 \\
    top-$1$ & 87.63 & 47.24 & 47.24 & {47.24} \\
    top-$2$ & 83.86 & 48.72 & 54.96 & {\tbf{51.23}} \\
    top-$3$ & 83.49 & 47.59 & 55.39 & {49.91} \\
    \bottomrule
    \end{tabular}
    }
    \label{tab:purify}
\end{table}{}

From Table \ref{tab:purify}, we observe that, for the top-$1$ case, the defense rate of `CIFS/CIFS' is the same as that of `CIFS/Final' and that of `Adap/Final.' This implies that, once an adversarial example successfully fools the raw prediction of CIFS, the final prediction also will be incorrect. Thus, the attacker only needs to focus on the CIFS's raw predictions to break the model. In contrast, for the top-$2$ case, the defense rate of `CIFS/Final' exceeds the `CIFS/CIFS' by 6 percentage points. This means that nearly 6\% adversarial data mislead the CIFS's raw predictions. However, through the channel adjustment via CIFS, these adversarial data are ``purified'', and more relevant characteristic features are thus transmitted to subsequent layers of CNNs. As such, these adversarial data are finally classified correctly. In this case, the attacker has to exhaustively search for an adaptive loss function to generate attacks, and the CIFS-modified CNNs are safer and more reliable. More discussion on why the top-$k$ assessment performs better and how to choose $k$ is provided in Appendix \ref{apdx:ablation}.

\section{Chapter Summary}

We developed the CIFS mechanism to verify the hypothesis that suppressing NR channels and aligning PR ones with their relevances to predictions benefits adversarial robustness. Empirical results demonstrate the effectiveness of CIFS on enhancing CNNs' robustness.

There are two limitations of our current work: 1) We empirically verify the hypothesis $\mathcal H$, but it is still difficult to explicitly, not intuitively, explain why the adjustment of channels improves robustness. 2) Although CIFS ameliorates the overfitting during AT and improves the robustness, it sometimes leads to a bit drop in natural accuracies on certain datasets. In the future, we will attempt to address these two limitations.

\section{Appendices}

\subsection{Details on Visualizing Channel-wise Activations} \label{apdx:visualization}

\subsubsection{Non-robust CNNs vs. Robustified CNNs} \label{apdx:visualization-normal-at}
We train ResNet-18 models to perform the classification task on the CIFAR10 dataset. The models are trained normally and adversarially. We use adversarial data generated by PGD-10 attack ($\epsilon=\nicefrac{8}{255}$, step size $\nicefrac{\epsilon}{4}$, and random initialization) for adversarial training.

The ResNet-18 network consists of one convolutional layer, eight residual blocks, and one linear fully-connected (FC) layer connected successively. Each residual block contains two convolutional layers for the residual mapping. We visualize the features of the penultimate layer (the output of the eighth residual block) and the weights of the last linear layer in Figure \textcolor{Blue}{1}. Specifically, the weights of the last FC layer for a certain class are sorted and plotted in descending order. We process the penultimate layer's features with the global average pooling operation to obtain the channel-wise activations. For a certain class, we calculate each channel's mean activation magnitude over all the test samples in this category. We normalize the mean channel-wise activations by dividing them by their absolute maximum. The mean channel-wise activations are plotted according to the indices of the sorted weights. We also record the activated frequency of each channel. Here, the channel is regarded to be activated if its activation magnitude is larger than a threshold (1\% of the maximum of all channels' activations). 

\subsubsection{Channel-wise Activations of CIFS-modifed CNNs}
\label{apdx:visualization-cifs-modification}
We train CIFS-modified CNNs normally and adversarially by using the adaptive loss in Equation (\textcolor{Blue}{3}). We use the PGD-10 attack to generate adversarial data. We illustrate the channels' activations of CIFS-modified CNNs in Figure \textcolor{Blue}{3}. The implementation details are same as those in Appendix \ref{apdx:visualization-normal-at}.

In Figure \textcolor{Blue}{3}, we show the channels' activations of data in class ``airplane''. Here, we plot the channels activations of data in other classes. From Figure \ref{fig:apx-hyp}, we see that CIFS indeed suppresses negatively-relevant (NR) channels and promotes the positively-relevant (PR) ones.

\begin{figure}[!ht]
	\centering
	\includegraphics[width=.7\linewidth]{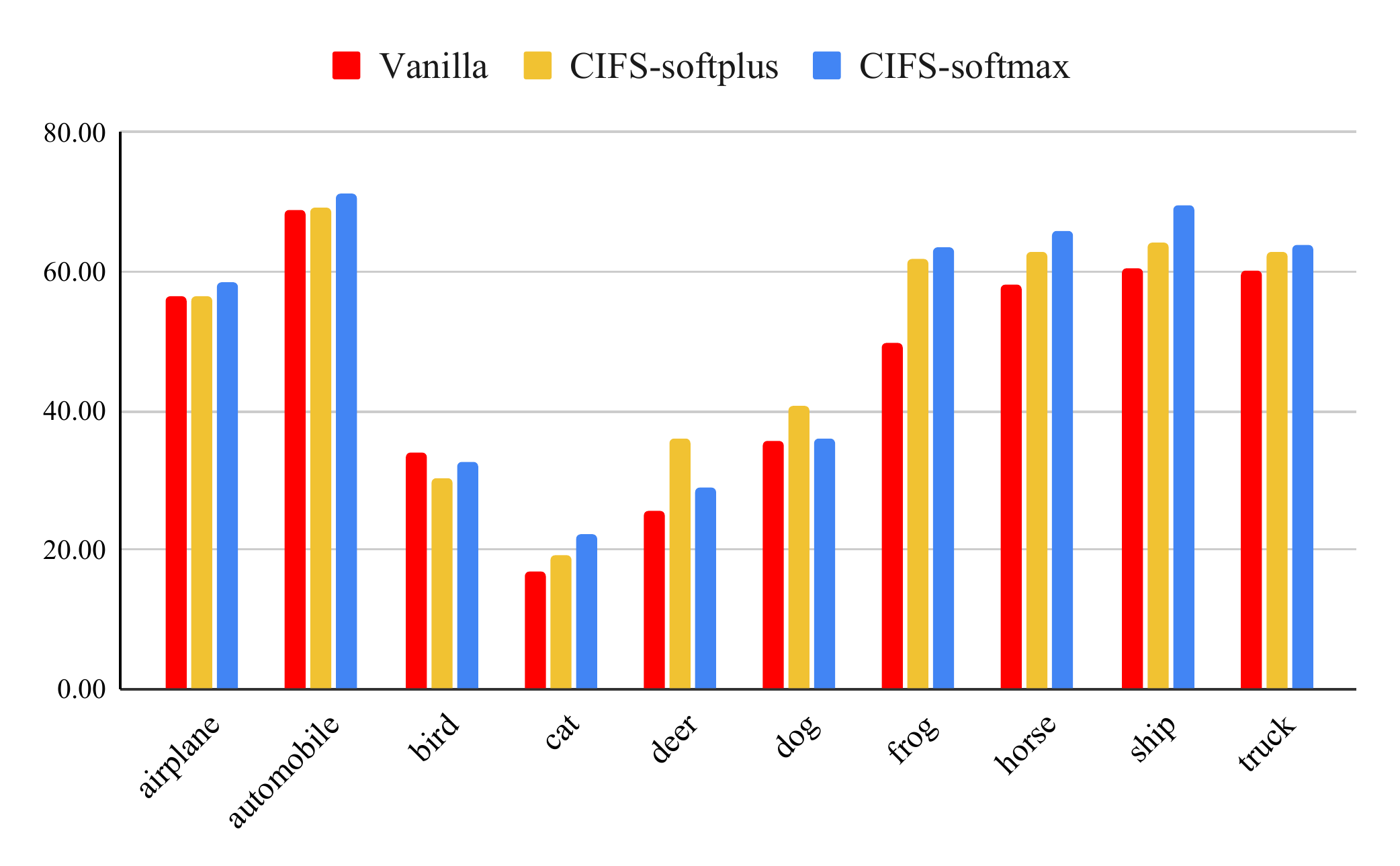}
	\caption{Robust accuracies (\%) of PGD-20 adversarial data for various classes of the CIFAR10 dataset.}
	\label{fig:apdx-imbal}
\end{figure}

Besides, we also observe that CIFS ameliorates the class-wise imbalance of robustness under AT. In Figure \ref{fig:apdx-imbal}, we can see that, for the data in class ``cat'' and class ``deer'', the robust accuracies of the vanilla ResNet-18 model are 16.70\% and 25.50\%. Modifying the vanilla model with CIFS-softmax, we can improve the robust accuracies by 5.6 and 3.3 percentage points, respectively.

\def \SubFigWidth {0.35} 
\def \SubImgWidth {.99}
\begin{figure}[!ht]
	\centering
	\begin{subfigure}{\SubFigWidth\linewidth}
		\centering
		\includegraphics[width=\SubImgWidth \linewidth]{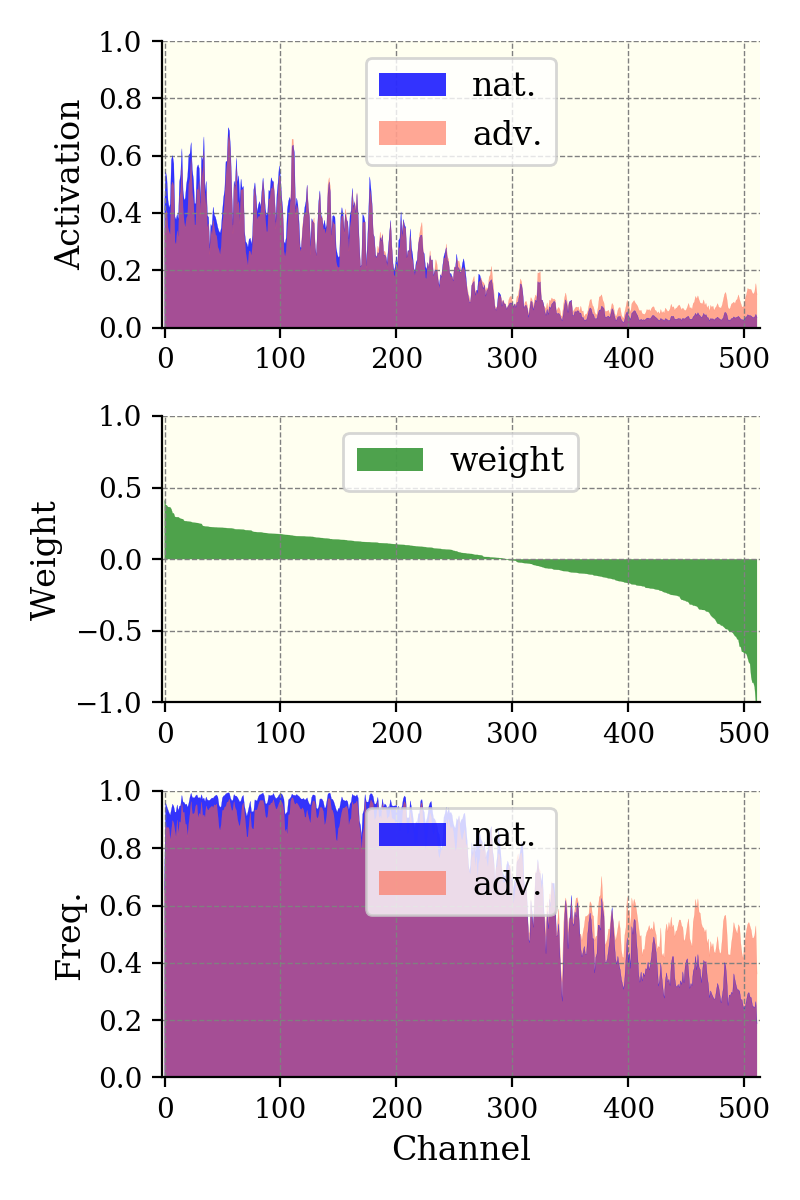}
		\caption{ ``ship'': non-CIFS}
	\end{subfigure}
	\begin{subfigure}{\SubFigWidth\linewidth}
		\centering
		\includegraphics[width=\SubImgWidth \linewidth]{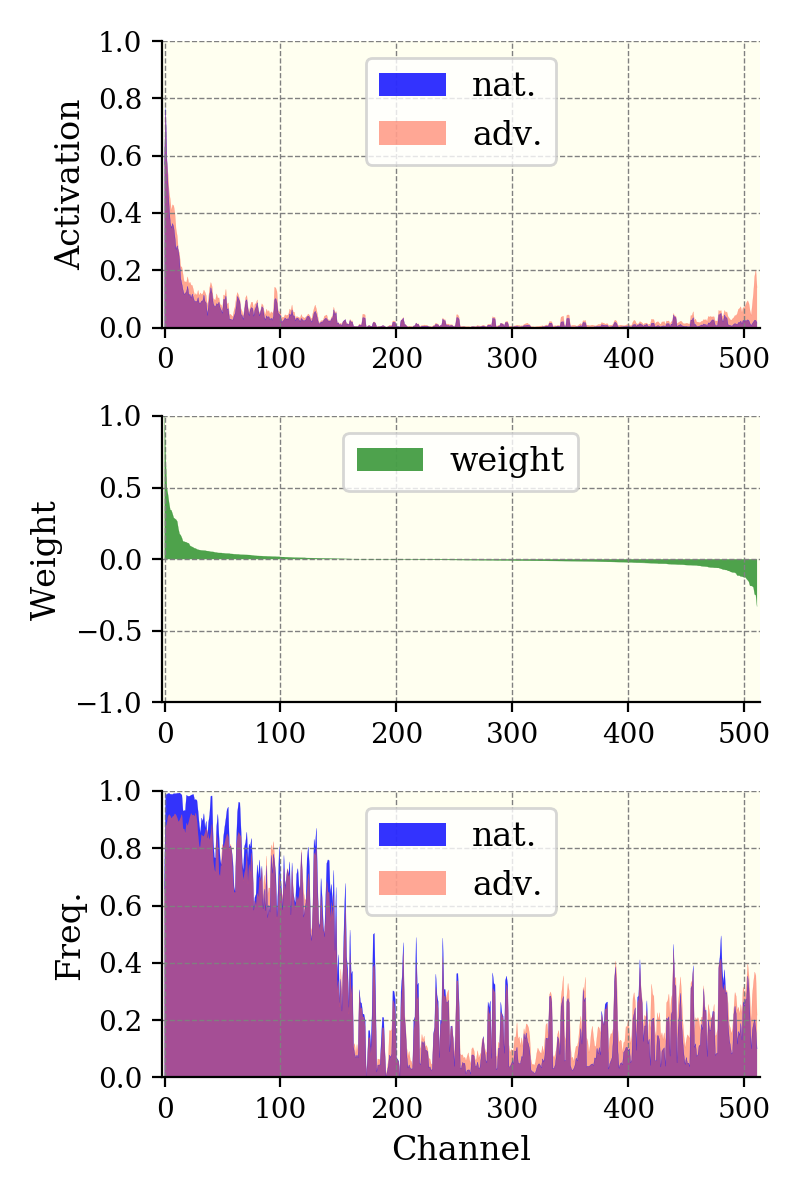}
		\caption{  ``ship'': CIFS-sigmoid}
	\end{subfigure}
	
	\begin{subfigure}{\SubFigWidth\linewidth}
		\centering
		\includegraphics[width=\SubImgWidth \linewidth]{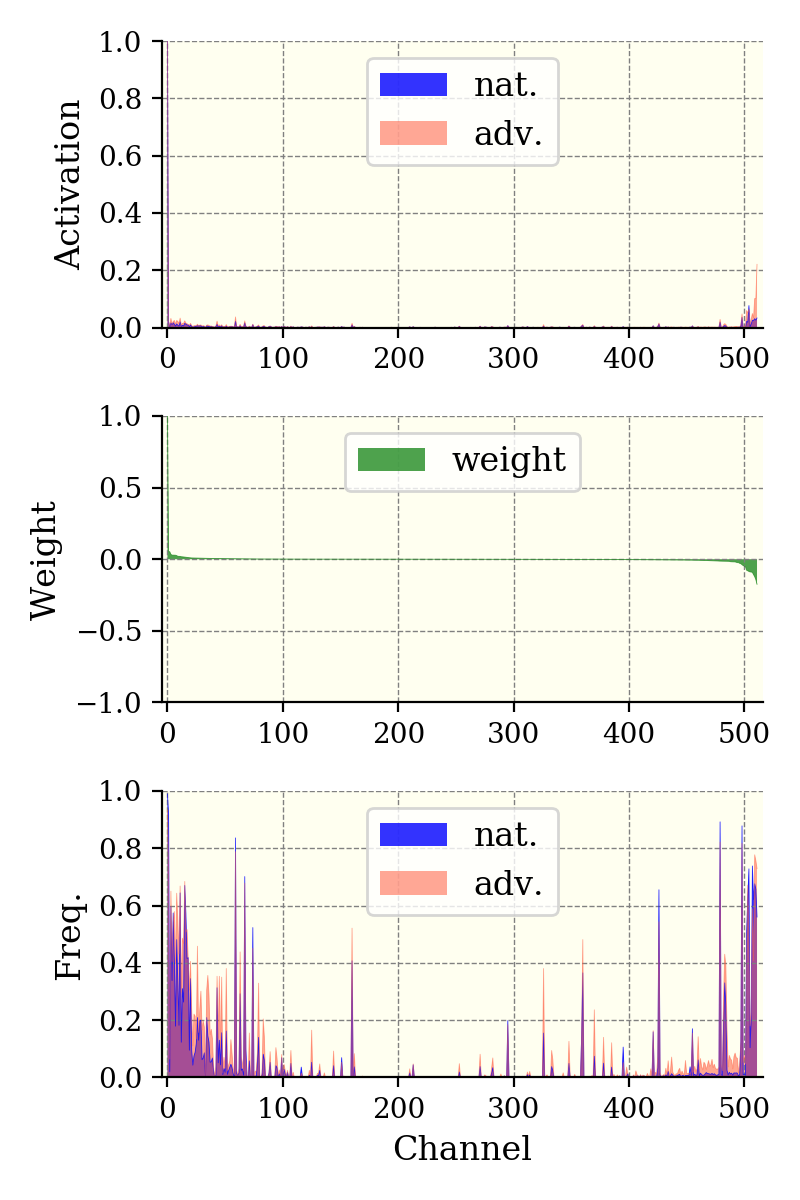}
		\caption{  ``ship'': CIFS-softplus}
	\end{subfigure}
	\begin{subfigure}{\SubFigWidth\linewidth}
		\centering
		\includegraphics[width=\SubImgWidth \linewidth]{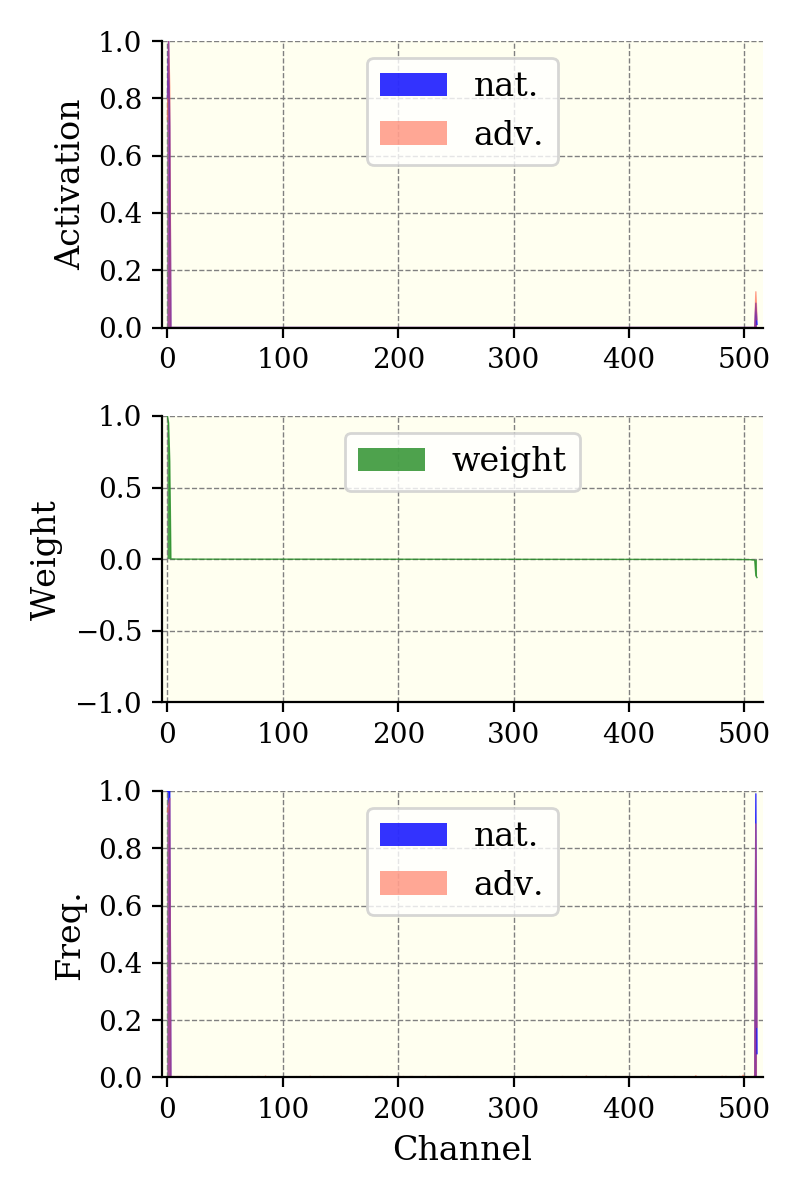}
		\caption{  ``ship'': CIFS-softmax}
	\end{subfigure}
	\caption[More comparison of channel activations of adversarial data between normally and adversarially trained models.]{The magnitudes of channel-wise activations (top) at the penultimate layer, their activated frequency (bottom), and the weights of the last linear layer (middle) \textit{vs.} channel indices. Data from class ``ship'' (a-d), ``automobile''(e-h), and ``frog''(i-l) , are used here. The robust accuracies against PGD-20 (on the whole dataset) are 46.64\% for non-CIFS, 49.87\% for the CIFS-sigmoid, 50.38\% for the CIFS-softplus, and 51.23\% for the CIFS-softmax respectively.}
	\label{fig:apx-hyp}
\end{figure}

\begin{figure}[!ht]\ContinuedFloat
	\centering
	\begin{subfigure}{\SubFigWidth\linewidth}
		\centering
		\includegraphics[width=\SubImgWidth \linewidth]{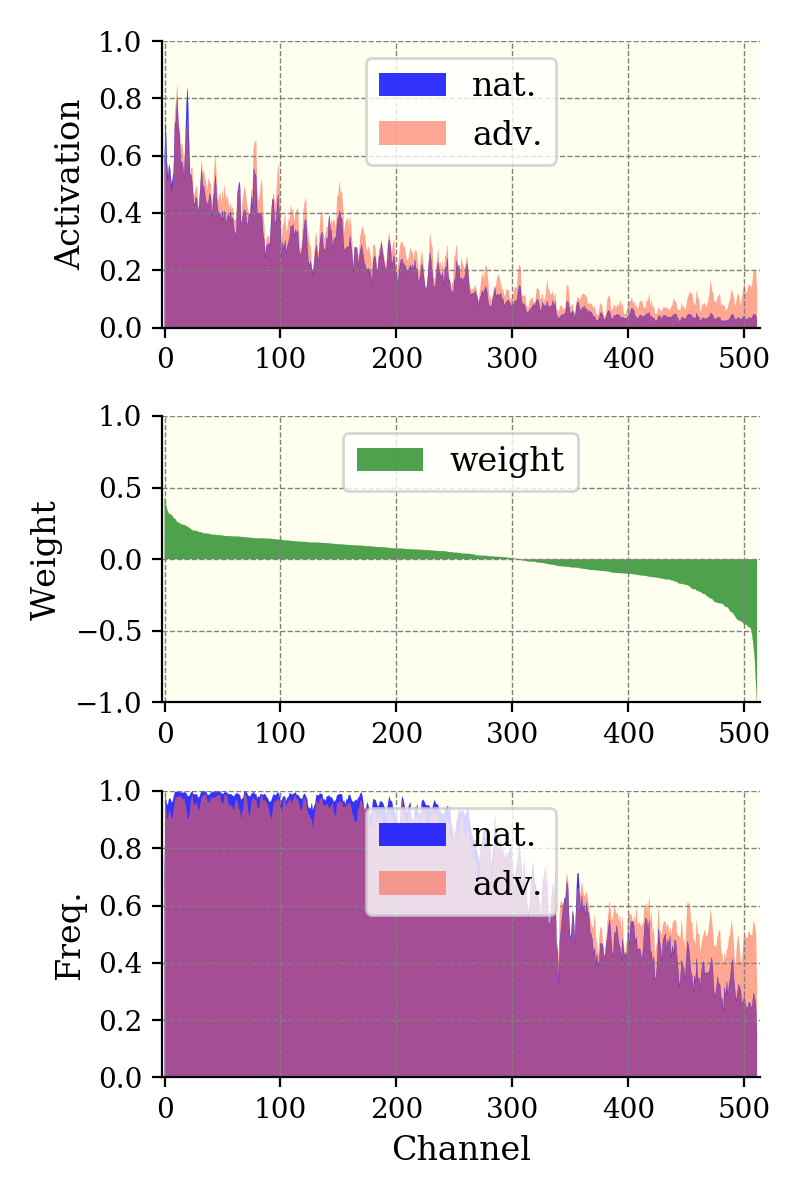}
		\caption{  ``autmobile'': non-CIFS}
	\end{subfigure}
	\begin{subfigure}{\SubFigWidth\linewidth}
		\centering
		\includegraphics[width=\SubImgWidth \linewidth]{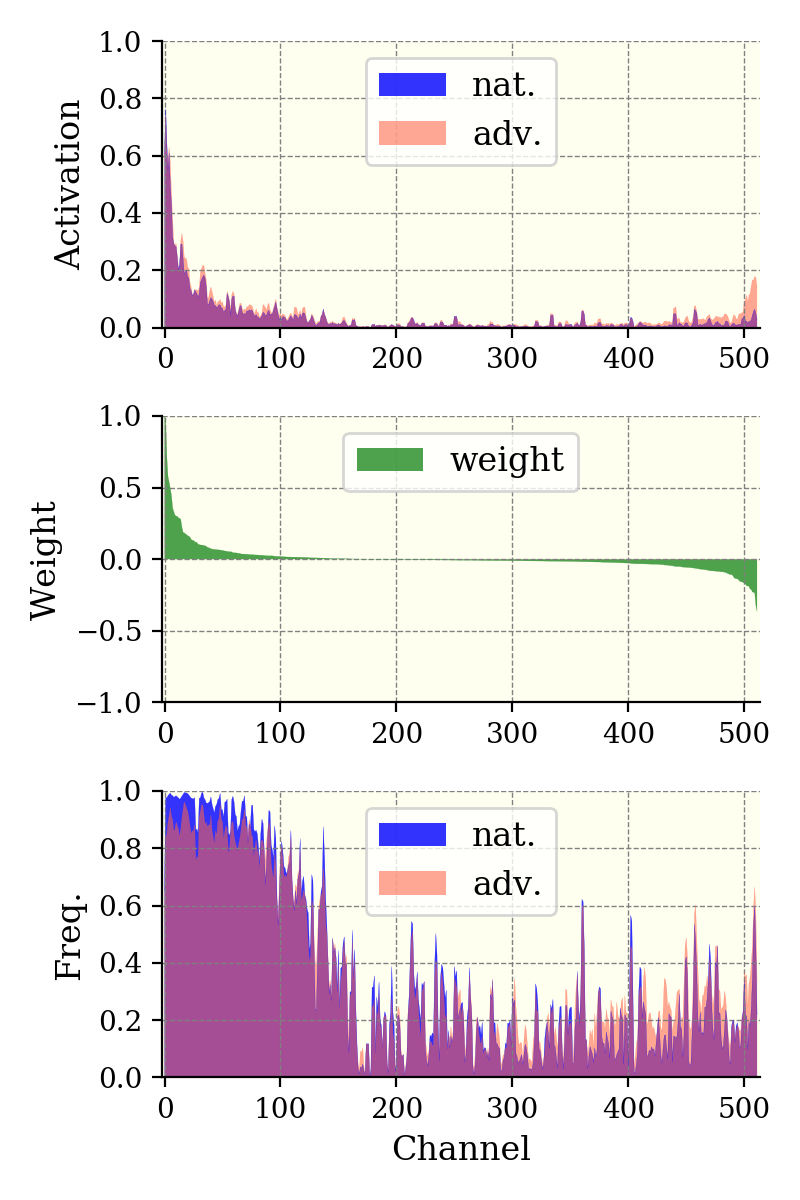}
		\caption{  ``autmobile'': CIFS-sigmoid}
	\end{subfigure}
	
	\begin{subfigure}{\SubFigWidth\linewidth}
		\centering
		\includegraphics[width=\SubImgWidth \linewidth]{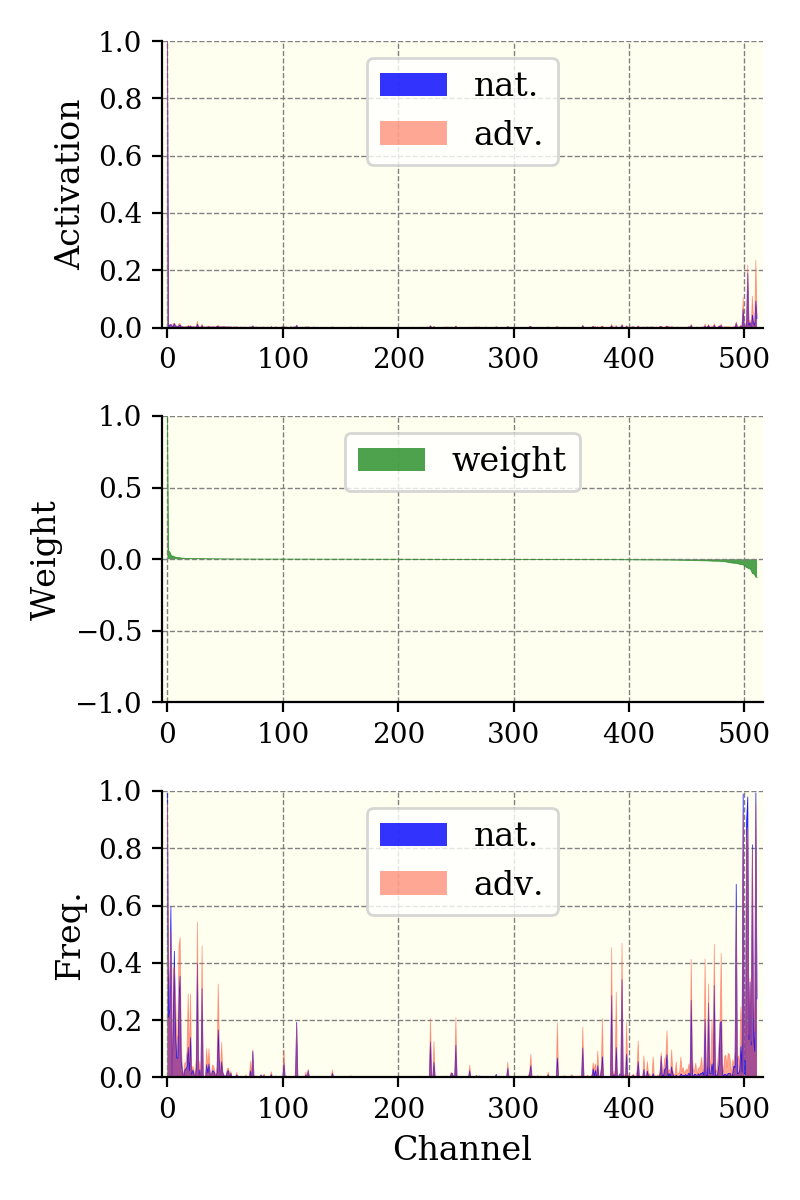}
		\caption{  ``autmobile'': CIFS-softplus}
	\end{subfigure}
	\begin{subfigure}{\SubFigWidth\linewidth}
		\centering
		\includegraphics[width=\SubImgWidth \linewidth]{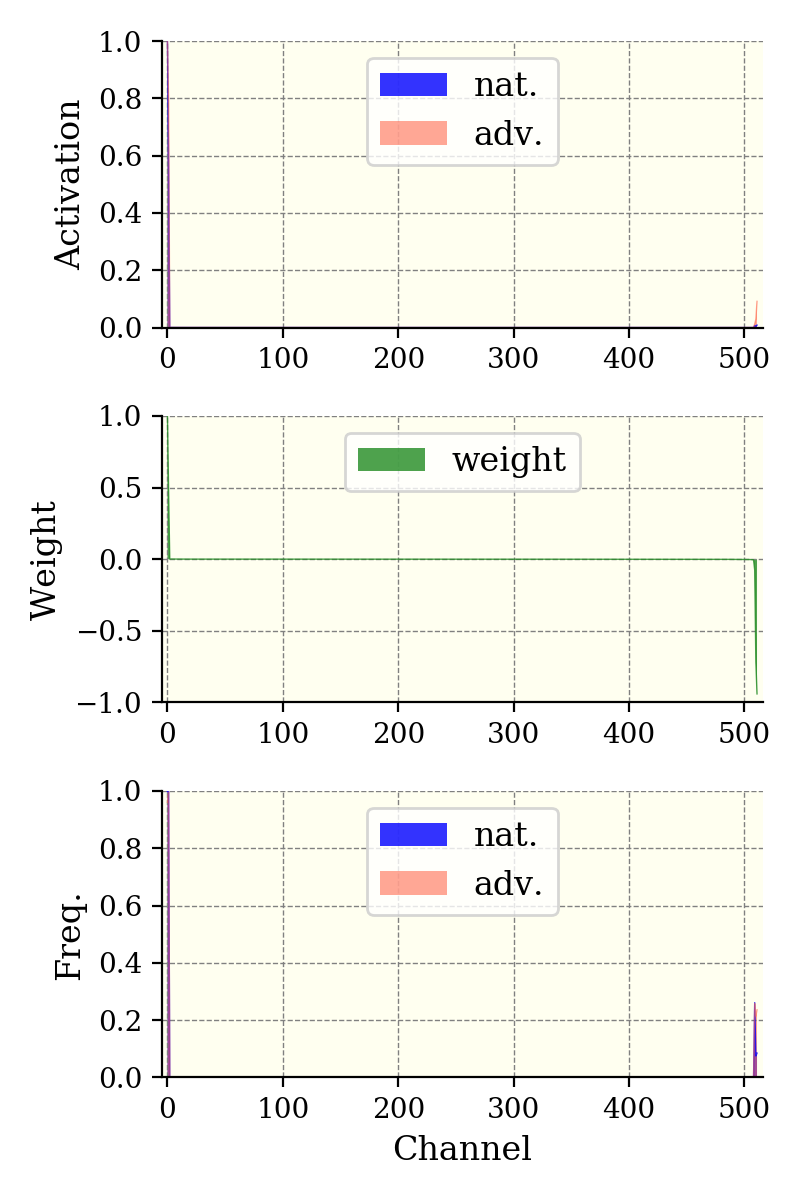}
		\caption{  ``autmobile'', CIFS-softmax}
	\end{subfigure}
	\caption[More comparison of channel activations of adversarial data between normally and adversarially trained models.]{The magnitudes of channel-wise activations (top) at the penultimate layer, their activated frequency (bottom), and the weights of the last linear layer (middle) \textit{vs.} channel indices. Data from class ``ship'' (a-d), ``automobile''(e-h), and ``frog''(i-l) , are used here. The robust accuracies against PGD-20 (on the whole dataset) are 46.64\% for non-CIFS, 49.87\% for the CIFS-sigmoid, 50.38\% for the CIFS-softplus, and 51.23\% for the CIFS-softmax respectively.}
	\label{fig:apx-hyp}
\end{figure}

\begin{figure}[!ht]\ContinuedFloat
	\centering
	\begin{subfigure}{\SubFigWidth\linewidth}
		\centering
		\includegraphics[width=\SubImgWidth \linewidth]{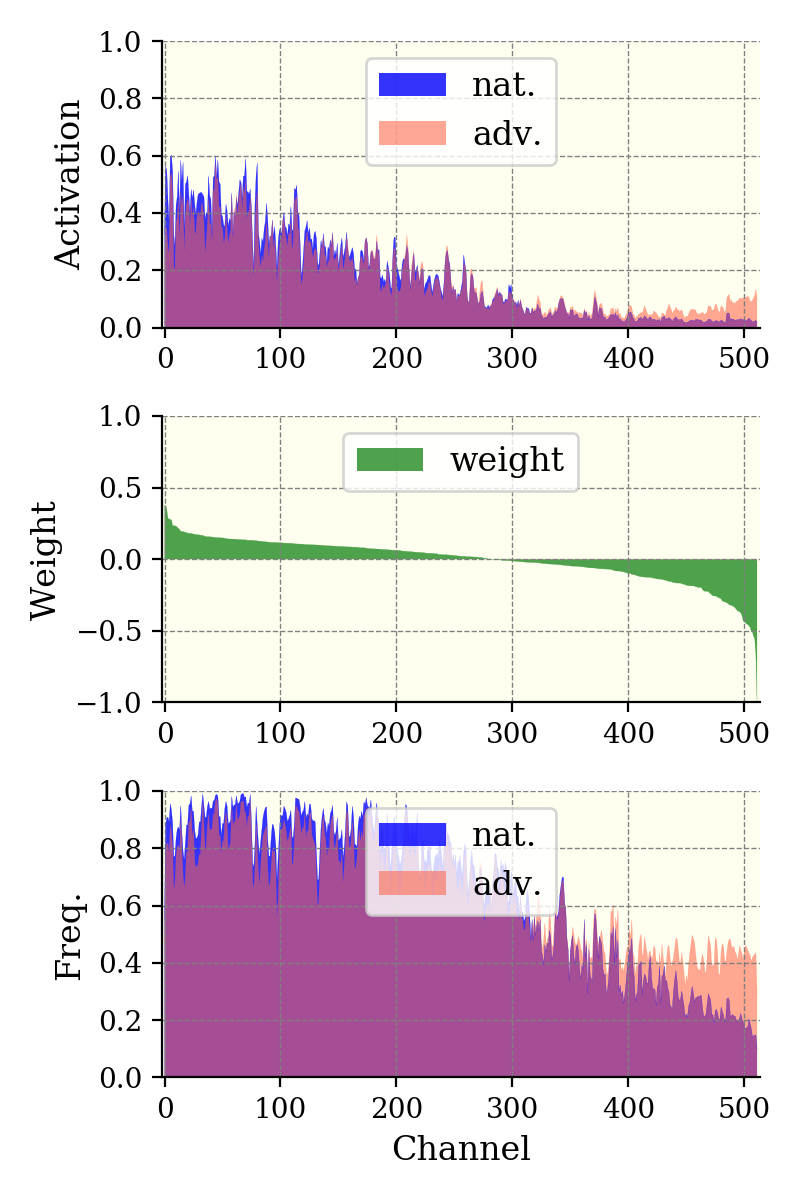}
		\caption{  ``frog'': non-CIFS}
	\end{subfigure}
	\begin{subfigure}{\SubFigWidth\linewidth}
		\centering
		\includegraphics[width=\SubImgWidth \linewidth]{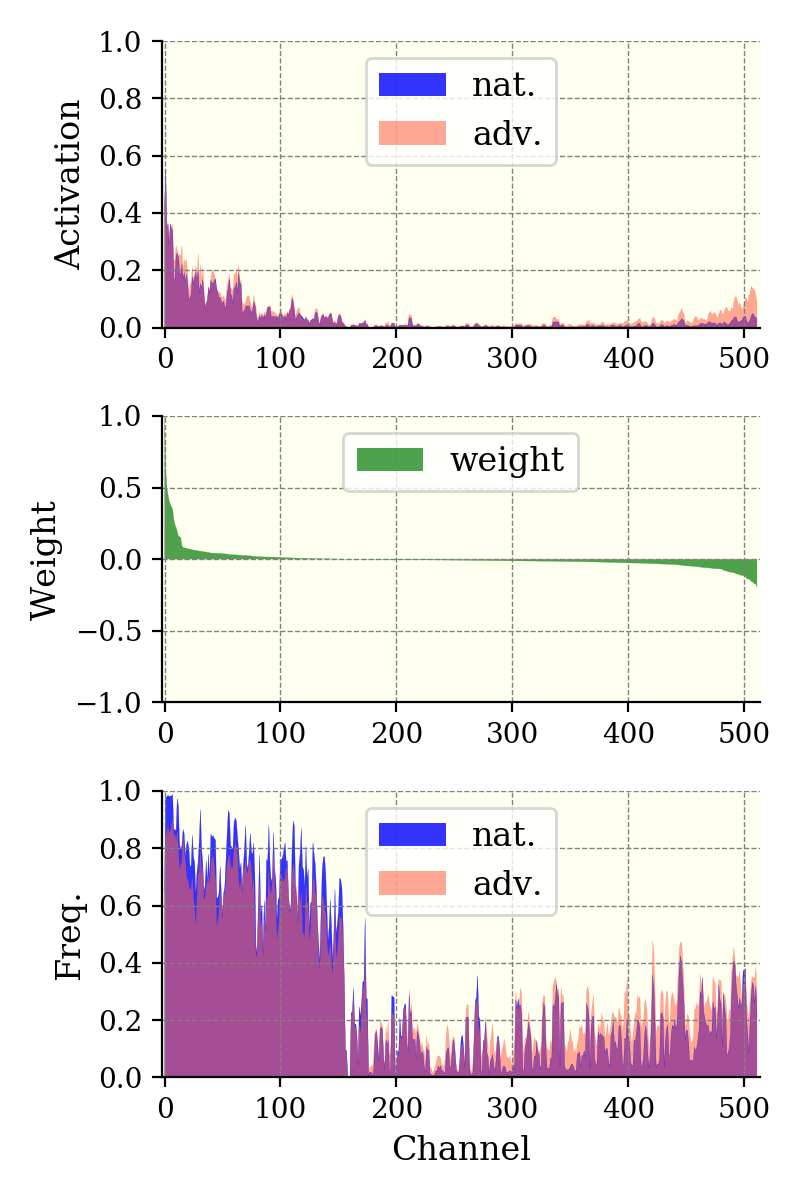}
		\caption{  ``frog'': CIFS-sigmoid}
	\end{subfigure}
	
	\begin{subfigure}{\SubFigWidth\linewidth}
		\centering
		\includegraphics[width=\SubImgWidth \linewidth]{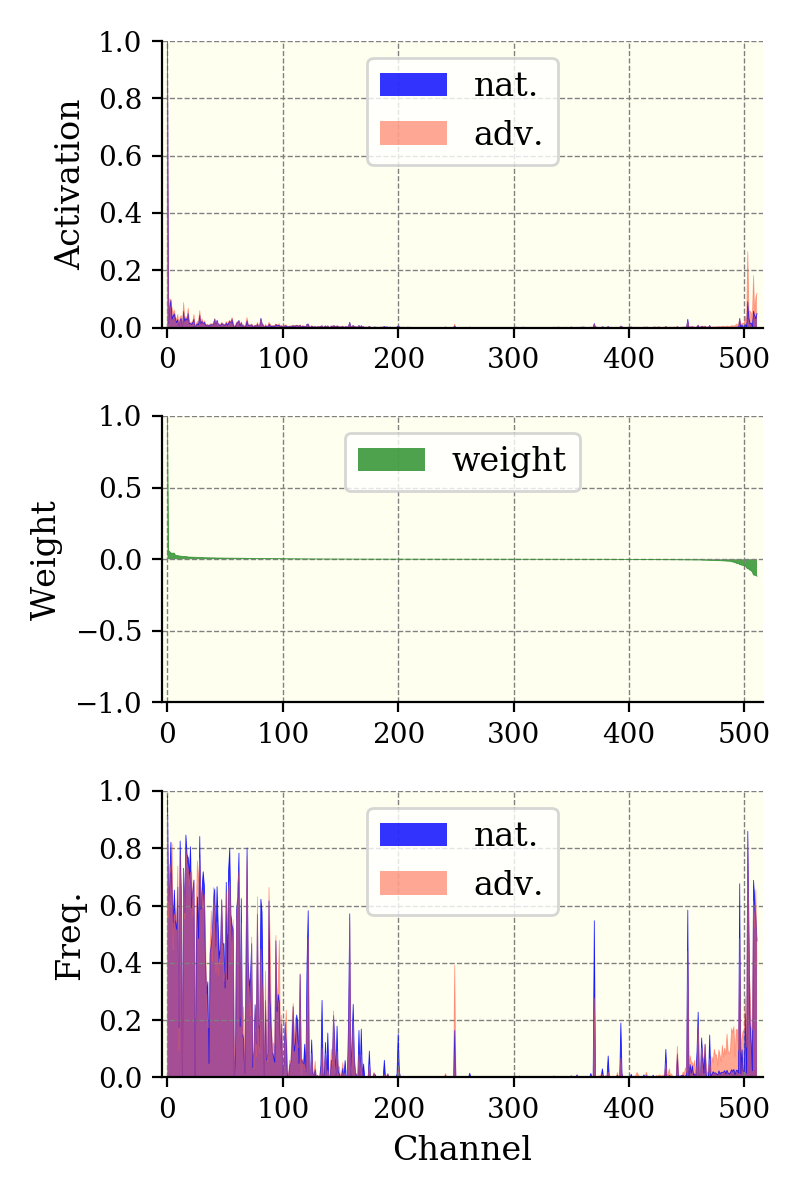}
		\caption{  ``frog'': CIFS-softplus}
	\end{subfigure}
	\begin{subfigure}{\SubFigWidth\linewidth}
		\centering
		\includegraphics[width=\SubImgWidth \linewidth]{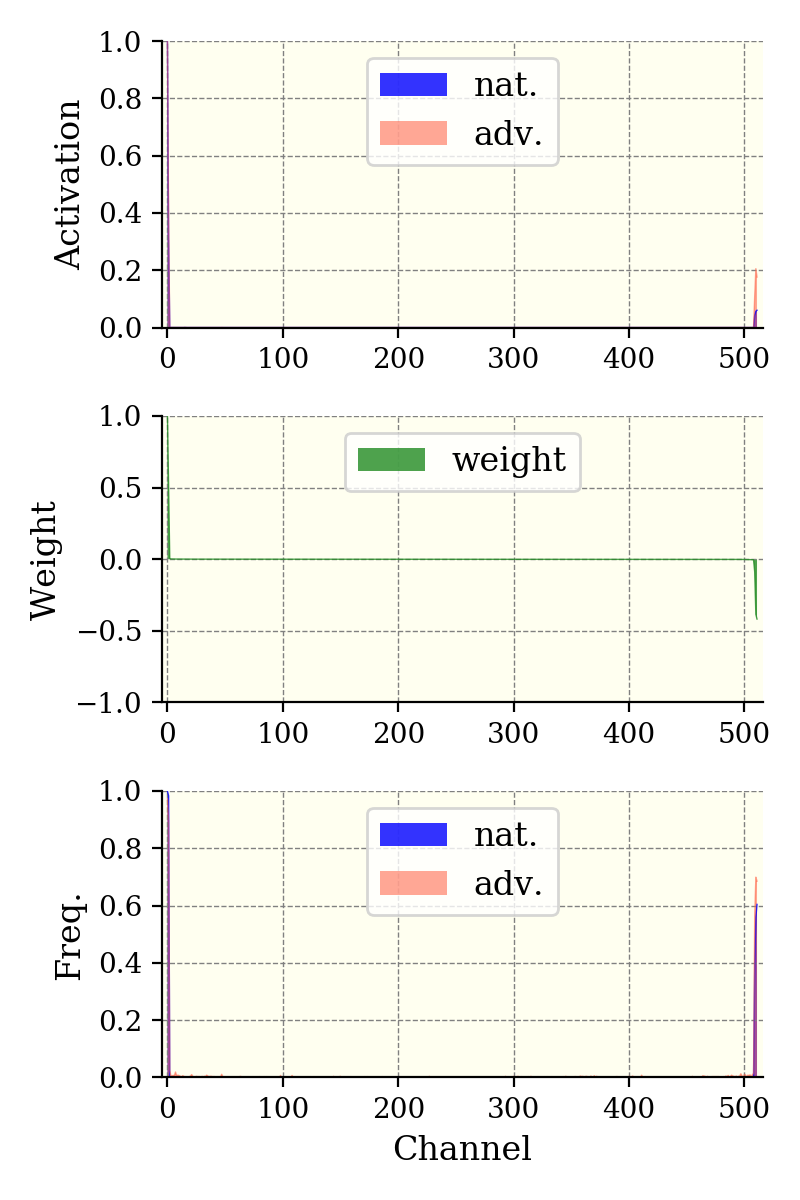}
		\caption{  ``frog'', CIFS-softmax}
	\end{subfigure}
	\caption[More comparison of channel activations of adversarial data between normally and adversarially trained models.]{The magnitudes of channel-wise activations (top) at the penultimate layer, their activated frequency (bottom), and the weights of the last linear layer (middle) \textit{vs.} channel indices. Data from class ``ship'' (a-d), ``automobile''(e-h), and ``frog''(i-l) , are used here. The robust accuracies against PGD-20 (on the whole dataset) are 46.64\% for non-CIFS, 49.87\% for the CIFS-sigmoid, 50.38\% for the CIFS-softplus, and 51.23\% for the CIFS-softmax respectively.}
	\label{fig:apx-hyp}
\end{figure}

\subsection{Robustness Evaluation on CIFAR10} \label{apdx:cifar10}

\subsubsection{Robustness Enhancement of CIFS under AT} \label{apdx:cifar10-at}
\textbf{Training and Evaluation details}: On the CIFAR10 dataset, we train ResNet-18 and WRN-28-10 models with PGD-10 adversarial examples ($\epsilon=\nicefrac{8}{255}$, step size $\nicefrac{\epsilon}{4}$ with random initialization). The $\beta$ in CIFS is set to be $2$. For the ResNet-18 and its CIFS-modified version, we train models for $120$ epochs with the SGD optimizer (momentum $0.9$ and weight decay $0.0002$). The learning rate starts from $0.1$ and is multiplied with $0.1$ at epoch $75$ and epoch $90$. For the WRN-28-10, we train model for $110$ epochs with weight decay $0.0005$. 

In Section \textcolor{Blue}{4.1}, we evaluate the robustness of CNNs against four white-box attacks with a perturbation budget $\epsilon=\nicefrac{8}{255}$ in $l_{\infty}$ norm --- FGSM, PGD-20 (step size $\nicefrac{\epsilon}{10}$), C\&W (optimized by PGD for 30 steps with a step size $\nicefrac{\epsilon}{10}$) and PGD-100 (step size $\nicefrac{\epsilon}{10}$).

\textbf{Robustness Evaluation with AutoAttack}: 
Here, we also report the robust accuracies of defense methods against AutoAttack \citep{croce_reliable_2020}, which consists of both white-box and black-box attacks. AutoAttack regards models to be robust at a certain data point only if the models correctly classify all types of adversarial examples generated by AutoAttack of that data point. We consider the AutoAttack including one strong white-box attack (Auto-PGD \citep{croce_reliable_2020}) and one black-box attack (Square-Attack \citep{andriushchenko2020square}). Since the Square Attack requires many queries, we sample 2,000 images (200 per class) from the CIFAR10 for evaluation. The attack parameters are set according to the officially released AutoAttack\footnote{\url{https://github.com/fra31/auto-attack}}. From Table \ref{tab:apdx-aa-cifar}, we observe that CIFS enjoys better robustness against AutoAttack in comparison to the vanilla ResNet-18 model and its CAS-modified version. 

\begin{table}[t!]
    \centering
    \caption[Robustness comparison (AutoAttack) of defense methods on CIFAR10.]{Robustness comparison of defense methods on CIFAR10. We report the last-epoch robust accuracies (\%) against AutoAttack.}
    \label{tab:apdx-aa-cifar}
    \scalebox{.9}{
    \begin{tabular}{cccc}
    \toprule
    Model & Vanilla & CAS & CIFS  \\
    ResNet-18 & 44.00 & 42.70 & \tbf{46.20} \\
    WRN-28-10 & 47.20 & 46.55 & \tbf{49.75} \\
    \bottomrule
    \end{tabular}
    }
\end{table}{}

\textbf{Best-epoch robustness during training}: 
Due to the susceptibility of overtrained models to overfitting \citep{rice_overfitting_2020}, it seems reasonable to compare the results at the end of the training (and not for the  best epochs) \citep{madry_towards_2018, zhang_attacks_2020, rice_overfitting_2020}. In Section \textcolor{Blue}{4.1}, we report the robust accuracies of ResNet-18 and WRN-28-10 models at the last epochs. Here, we also provide the results at the \tbf{best} epochs for reference. 

From Table \ref{tab:apdx-cifar-best}, we see that, for the ResNet-18 architecture, the CIFS-modified model results in the similar best-epoch robustness (PGD-100) to that of the vanilla ResNet-18. For the WRN-28-10, the vanilla model has the better best-epoch robustness compared to the CIFS-modified version. This may be due to the fact that CIFS suppresses redundant channels and reduces the model capacity. 

By comparing results in Table \ref{tab:apdx-cifar-best} with those in Table \textcolor{Blue}{1}, we observe that CIFS indeed ameliorates the overfitting of AT. Specifically, the best-epoch robust accuracy of the vanilla WRN-28-10 (resp. ResNet-18) against PGD-100 attack is 54.17\% (resp.49.47\%), but the last-epoch accuracy drops to 47.08\% (resp. 44.72\%). In contrast, for the CIFS-modified versions, the last-epoch robust accuracies against PGD-100 attack are maintained around the best-epoch ones (for WRN-28-10, from 52.03\% to 51.51\%; for ResNet-18, from 49.76\% to 48.74\%). 

\begin{table}[h!]
    \centering
    \caption[Robustness comparison (best-epoch) of defense methods on CIFAR10]{Robustness comparison of defense methods on CIFAR10. We report the best robust accuracies (\%) during training. For each model, the results of the strongest attacks are marked with an underline.}
    \label{tab:apdx-cifar-best}
    \scalebox{.9}{
    \begin{tabular}{cccccc}
    \toprule
    \tbf{\tit{ResNet-18}} & Natural & FGSM & PGD-20 & C\&W & PGD-100 \\
    \hline
    Vanilla & 83.63 & 56.73 & 50.64 & 49.51 & \underline{49.47} \\
    CAS & 85.66 & 56.25 & 47.69 & 46.52 & \underline{45.69} \\
    CIFS & 82.46 & \tbf{58.98} & \tbf{51.94} & \tbf{51.25} & \tbf{\underline{49.76}} \\
    \bottomrule
    \toprule
    \tbf{\tit{WRN-28-10}} & Natural & FGSM & PGD-20 & C\&W & PGD-100\\
    \hline
    Vanilla & 86.53 & \tbf{61.43} & \tbf{55.69} & \tbf{54.45} & \tbf{\underline{54.17}} \\
    CAS & 87.51 & 58.54 & 52.06 & 51.27 & \underline{50.69} \\
    CIFS & 84.67 & 61.03 & 54.09 & 53.76 & \underline{52.03} \\
    \bottomrule
    \end{tabular}
    }
\end{table}{}


\textbf{Robust accuracies for various values of $\beta_{\text{atk}}$:} In Section \textcolor{Blue}{4.1}, we evaluate the robustness of CIFS-modified models by using the adaptive loss in Equation (\textcolor{Blue}{3}). For each type of attack, we assign various values to $\beta_{\text{atk}}$ and report the worst robust accuracies. Here, for reference, we provide the defense results of the ResNet-18 model on CIFAR10 for different values of $\beta_{\text{atk}}$ that are used in Section \textcolor{Blue}{4.1}. The results in Table \textcolor{Blue}{1} (ResNet-18) are collected from Table \ref{tab:apdx-beta-attack}. 

\begin{table}[t!]
    \centering
    \caption[Robust accuracies (\%) for values of $\beta_{\text{atk}}$ on CIFAR10.]{Robust accuracies (\%) for values of $\beta_{\text{atk}}$ on CIFAR10. The value ``$\infty$'' means the attack only considers the second term in Equation (\textcolor{Blue}{3}). The value ``$\infty$-1'' (resp. ``$\infty$-2'') means the attacker completely focuses on the first (resp. second) CIFS-modifed layer. The bracketed numbers are those reported in Table \textcolor{Blue}{1} (ResNet-18).}
    \scalebox{.9}{
    \begin{tabular}{ccccccc}
    \toprule
    \tbf{\tit{ResNet-18}} & $\beta_{\text{atk}}$ & Natural &  FGSM & PGD-20 & C\&W & PGD-100 \\
    \midrule
    Vanilla & - & [84.56] & [55.11] & [46.62] & [45.95] & [\underline{44.72}] \\
    \hline
    CAS     & 0& [86.73] & 83.17 & 88.45 & 88.52 & 88.24 \\
    ~ & 0.1 & - & 58.61 & 61.36 & 85.51 & 62.40\\
    ~ & 1 & - & 56.36 & 52.86 & 62.34 & 56.02\\
    ~ & 2 & - & 56.06 & 49.76 & 54.94 & 50.62\\
    ~ & 10 & - & 56.03 & 47.47 & 49.35 & 47.70\\
    ~ & 100 & -  & 56.02 & 47.04 & 48.36 & 46.74\\
    ~ & $\infty$ & - & 56.02 & 47.06 & 48.31 & 46.55\\
    ~ & $\infty$-1 & - & [55.99] & [45.29] & [44.18] & [\underline{43.22}] \\
    ~ & $\infty$-2 & - & 82.68 & 87.87 & 87.79 & 87.72\\
    \hline
    CIFS & 0 & [83.86] & 60.58 & 52.64 & 51.32 & 49.94 \\
    ~ & 0.1 & - & [\tbf{58.86}] & 51.40 & 50.88 & 49.42\\
    ~ & 1 & - & 59.20 & 51.28 & [\tbf{50.16}] & 48.74\\
    ~ & 2 & - & 59.24 & [\tbf{51.23}] & 50.28 & 48.79 \\
    ~ & 10 & - & 59.35 & 51.27 & 50.70 & [\underline{\tbf{48.70}}] \\
    ~ & 100 & - & 59.38 & 51.41 & 51.04 & 48.80\\
    ~ & $\infty$ & - & 59.43 & 51.45 & 51.08 & 48.82\\
    ~ & $\infty$-1 & - & 61.06 & 54.96 & 53.83 & 52.82\\
    ~ & $\infty$-2 & - & 60.03 & 52.30 & 50.92 & 50.03\\
    
    \bottomrule
    \end{tabular}
    }
    \label{tab:apdx-beta-attack}
\end{table}{}

\subsubsection{Robustness Enhancement under TRADES} \label{apdx:cifar10-trades}
To improve the robustness of CNNs, various training-based strategies have been proposed, including vanilla adversarial training (AT) \citep{madry_towards_2018}, friendly-adversarial training (FAT) \citep{zhang_attacks_2020}, and TRADES \citep{zhang_theoretically_2019}. In Section \textcolor{Blue}{4.1}, we show that CIFS can further enhance the robustness of CNNs under the vanilla AT and FAT. Here, we conduct more experiments to check whether TRADES is also suitable for CIFS.

\begin{table}[h!]
    \centering
    \caption[Robustness comparison of vanilla CNNs and their CIFS-modified version under various AT-based strategies.]{Robustness comparison of vanilla CNNs and their CIFS-modified version under various AT-based strategies. We report the robust accuracies (\%) on various types of adversarial data.}
    \label{tab:apdx-trades}
    \scalebox{.9}{
    \begin{tabular}{ccccc}
    \toprule
    \tbf{\tit{ResNet-18}} & Natural & FGSM & PGD-20 & PGD-100 \\
    \hline
    Vanilla-AT & 84.56 & 55.11 & 46.62 & \underline{44.72} \\
    Vanilla-TRADES & 83.96 & 57.09 & 50.27 & \underline{{48.83}}  \\
    Vanilla-FAT & 87.16 & 56.43 & 47.64  & \underline{45.35} \\
    \hline
    CIFS-AT     & 83.86 & {58.86} & {51.23} & \underline{{48.74}} \\
    CIFS-TRADES     & 85.20 & 54.76 & 46.13 & \underline{43.65}  \\
    CIFS-FAT & 86.35 & \tbf{59.47} & \tbf{51.68} & \underline{\tbf{49.52}}\\
    \bottomrule
    \end{tabular}
    }
\end{table}{}

From Table \ref{tab:apdx-trades}, we observe that, for the vanilla ResNet-18 model, TRADES effectively robustifies the network and outperforms its counterparts by a large margin (e.g., 48.83\% of TRADES \tit{vs.} 44.72\% of AT against PGD-100 attack). However, for the CIFS-modified models, TRADES performs worse than AT and FAT. In general, CIFS-modification in combination with the FAT training strategy achieves the best robustness against various attacks.

\subsection{Robustness Evaluation on SVHN} \label{apdx:svhn}
\textbf{Training and Evaluation details}: On the SVHN dataset, we train the ResNet-18 model and its CIFS-modified version with PGD-10 adversarial examples ($\epsilon=\nicefrac{8}{255}$, step size $\nicefrac{\epsilon}{4}$ with random initialization). We train models for $120$ epochs with the SGD optimizer (momentum $0.9$ and weight decay $0.0005$). The learning rate starts from $0.01$ and is multiplied with $0.1$ at epoch $75$ and epoch $90$.

In Section \textcolor{Blue}{4.1}, we evaluate the robustness of CNNs against four white-box attacks with a perturbation budget $\epsilon=\nicefrac{8}{255}$ in $l_{\infty}$ norm --- FGSM, PGD-20 (step size $\nicefrac{\epsilon}{10}$), C\&W (optimized by PGD for 30 steps with a step size $\nicefrac{\epsilon}{10}$) and PGD-100 (step size $\nicefrac{\epsilon}{10}$).

\textbf{Robustness Evaluation with AutoAttack}: 
Here, we also report the robust accuracies of defense methods against AutoAttack on SVHN (Table \ref{tab:apdx-aa-svhn}). The evaluation settings of AutoAttack follows those in Appendix \ref{apdx:cifar10-at}.

\begin{table}[h!]
    \centering
    \caption[Robustness comparison (AutoAttack) of defense methods on SVHN.]{Robustness comparison of defense methods on SVHN. We report the robust accuracies (\%) at the last epochs.}
    \label{tab:apdx-aa-svhn}
    \scalebox{.9}{
    \begin{tabular}{cccc}
    \toprule
    \tbf{\tit{ResNet-18}} & Vanilla & CAS & CIFS  \\
    AutoAttack & 40.60 & 39.30 & \tbf{42.10} \\
    \bottomrule
    \end{tabular}
    }
\end{table}{}

\tbf{Best-epoch robustness during training:} In Section \textcolor{Blue}{4.1}, we report the robust accuracies of ResNet-18 models at the last epochs during training. Here, we report the best-epoch robustness for reference (Table \ref{tab:apdx-svhn-best}). We see that CIFS modified version enjoys the better best-epoch robustness in comparison to the vanilla ResNet-18 model.
\begin{table}[h!]
    \centering
    \caption[Robustness comparison (best-epoch) of defense methods on SVHN.]{Robustness comparison of defense methods on SVHN. We report the best robust accuracies (\%) during training. For each model, the results of the strongest attack are marked with an underline.}
    \label{tab:apdx-svhn-best}
    \scalebox{.9}{
    \begin{tabular}{cccccc}
    \toprule
    \tbf{\tit{ResNet-18}} & Natural & FGSM & PGD-20 & C\&W & PGD-100 \\
    \hline
    Vanilla & 93.88 & 66.02 & 51.71 & 48.87 & \underline{47.59} \\
    CAS & 93.90 & 65.53 & 50.52 & 48.39 & \underline{46.39} \\
    CIFS & 93.27 & \tbf{67.36} & \tbf{52.67} & \tbf{50.20} & \tbf{\underline{48.36}} \\
    \bottomrule
    \end{tabular}
    }
\end{table}{}

\subsection{More Results on FMNIST} \label{apdx:fmnist}
\textbf{Training and Evaluation details:} On the FMNIST dataset, we train ResNet-10 with PGD-20 adversarial examples ($\epsilon=0.3$, step size $0.02$ with random initialization). The $\beta$ in CIFS is set to be $2$. We train models for $120$ epochs with the SGD optimizer (momentum $0.9$ and weight decay $0.0002$). The learning rate starts with $0.1$ and is multiplied with $0.1$ at epochs $45$, $75$ and $90$.

We evaluate the robustness of the ResNet-10 models against FGSM, PGD-20, and PGD-100 white-box attacks. The perturbation is bounded by $\epsilon=0.3$ in $l_{\infty}$ norm. The step size of PGD-20 is set to be $0.01$, and that of PGD-100 is set to be $0.02$. Here, we report both the last-epoch robust accuracies and the best-epoch robust accuracies in Table \ref{tab:apdx-fmnist-robustness}.

\begin{table}[h!]
    \centering
    \caption[Robustness comparison of defense methods on FMNIST.]{Robustness comparison of defense methods on FMNIST. For each model, the robust accuracies (\%) of the strongest attack are remarked with an underline. For each type of attack, the best defense results are highlighted in bold. Comparing the defense rates of the strongest attacks, we observe that CIFS outperforms other defenses by a large margin.}
    \label{tab:apdx-fmnist-robustness}
    \scalebox{.9}{
    \begin{tabular}{cccccc}
    \toprule
    \tbf{\tit{Last}} & Natural & FGSM & PGD-40 & PGD-100 \\
    \hline
    Vanilla & 85.19 & {80.52} & 66.47 & \underline{60.99} \\
    CAS & 86.59 & \tbf{82.45} & 65.58 & \underline{59.51} \\
    CIFS & 83.35     & 77.48 & \tbf{66.59} & \underline{\tbf{65.50}} \\
    \bottomrule
    \toprule
    \tbf{\tit{Best}} & Natural & FGSM & PGD-40 & PGD-100 \\
    \hline
    Vanilla & 85.19 & {81.21} & 67.63 & \underline{63.36} \\
    CAS & 86.63 & \tbf{83.59} & 68.73 & \underline{62.65} \\
    CIFS & 83.32     & 78.55 & \tbf{69.05} & \underline{\tbf{67.21}} \\
    \bottomrule
    \end{tabular}
    }
\end{table}

\subsection{More Results on Ablation Study} \label{apdx:ablation}

\subsubsection{Effects of $\beta$ in CIFS:} \label{apdx:ablation-beta}

Here, we train CIFS-modified ResNet-18 models on CIFAR10 with various values of $\beta$ in Equation (\textcolor{Blue}{3}). The coefficient $\beta$ balances the accuracies of raw predictions and the final prediction. From Table \ref{tab:apdx-cifar-beta}, we observe that $\beta$ values that are too small or too large values lead to drops in the accuracies of natural data and adversarial data. On the one hand, if the value of $\beta$ is too small, the raw predictions made by CIFS are not reliable. Thus, the channels selected by CIFS may not be the truly relevant ones with respect to the ground-truth class. On the other hand, if the value of $\beta$ is too large, the optimization procedure mostly considers the raw predictions, the final prediction (output) becomes unreliable. When $\beta=2$, we achieve the best robustness against various types of attack.

\begin{table}[h!]
    \centering
    \caption{Robustness accuracies (\%) on CIFAR10 for CIFS with various values of $\beta$.}
    \label{tab:apdx-cifar-beta}
    \scalebox{.9}{
    \begin{tabular}{ccccc}
    \toprule
    \tbf{\tit{ResNet-18}} & Natural & FGSM & PGD-20 & PGD-100 \\
    \hline
    Vanilla & 84.56 & 55.11 & 46.62 & \underline{44.72} \\
    $\beta=0.1$ & 75.22 & 53.41 & 48.10 & \underline{46.28} \\
    $\beta=1$ & 82.34 &58.15 & 50.50 & \underline{48.35}\\
    $\beta=2$ & 83.86 & \tbf{58.86} & \tbf{51.23} & \underline{\tbf{48.74}} \\
    $\beta=10$ & 82.97 & 57.62 & 49.34 & \underline{47.10}\\
    $\beta=100$ & 75.41 & 52.90 & 45.00 & \underline{43.12}\\
    
    \bottomrule
    \end{tabular}
    }
\end{table}{}

\subsubsection{Effects of the top-$k$ feature assessment}

In general, $k$ should be larger than 1 but not too large. 

If we use the top-$1$, once adv. data fool probe nets, the channels relevant to true labels will be missed, and this will lead to wrong predictions (Table \textcolor{Blue}{5}, line top-1). Instead, we use top-$2$ for reliable channel selection. The efficacy is attributed to \tit{two} aspects: {Firstly}, the top-$2$ accuracies of adv. data are usually high (see Table \textcolor{Blue}{4}), thus channels relevant to top-$2$ logits \tit{include those relevant to the true class}. 
{Secondly}, 
\citet{tian2021analysis} reports that CNNs' predictions of adv. data usually belong to the superclass that contains true labels. Classes (e.g., cat, dog) in the same superclass (e.g., animals) usually share similar semantic features. Thus, {most of the top-$2$ selected channels are useful} for predicting the true class. 

Although the top-$2$ selected channels may contain info about the other wrong class, the following layers (after CIFS) are capable of “purifying” features and make better predictions. This is verified by Table \textcolor{Blue}{5}, the results in the line top-2 (CIFS/CIFS 48.72\% vs. CIFS/Final 54.96\%) mean that around 6\% adv. data, which successfully fool probes, are still finally correctly classified. However, too large $k$ may degrade the relevance assessment due to too much noisy info (e.g., the effect of top-3 is worse than top-2 in Table \textcolor{Blue}{5}).

\subsubsection{Layers to be modified}

\tbf{Positions of CIFS modules}: Here, we try different combinations of the layers to be modified by CIFS. In CNNs, the features of deep layers are usually more characteristic in comparison to those in the shallower layers \citep{zeiler_visualizing_2014}, and each channel of the features captures a distinct view of the input. The predictions often depend only on the information of a few essential views of the inputs. CIFS improves adversarial robustness by adjusting channel-wise activations. Thus, we apply CIFS to the deeper layers instead of the shallower ones. Specifically, we modify the ResNet-18 by applying CIFS at the last (P1) and/or the second last (P2) residual blocks. The experimental results are reported in Table \ref{tab:positions}. We observe that simultaneously applying CIFS into P1 and P2 performs the best against various attacks. Intuitively, because the features can be progressively refined, applying CIFS at P1\&P2 better purifies the channels compared to applying it only at P1 or P2.

\begin{table}[h]
    \centering
    \caption{Robustness (\%) comparison of the positions where CIFS modules are placed.}
    \scalebox{.9}{
    \begin{tabular}{cccccc}
    \toprule
    \tbf{\tit{ResNet-18}} & Natural & FGSM & PGD-20 & PGD-100\\
    \hline
    Vanilla & 84.56 & 55.11 & 46.62 & \underline{44.72} \\
    P1 & 84.02 & 57.60 & 48.45 & \underline{45.95} \\
    P2 & 82.62 & 56.55 & 47.22 & \underline{44.81} \\
    P1-P2 & 83.86 & \tbf{58.86} & \tbf{51.23} & \underline{\tbf{48.74}} \\
    \bottomrule
    \end{tabular}
    }
    \label{tab:positions}
\end{table}{}

\begin{table}[h]
	\centering
	\caption{Robustness comparison (\%) of the probe architectures in CIFS modules (at P1-P2).}
	\scalebox{.9}{
		\begin{tabular}{cccccc}
			\toprule
			\tbf{\tit{ResNet-18}} & Natural & FGSM & PGD-20 & PGD-100\\
			\hline
			Vanilla & 84.56 & 55.11 & 46.62 & \underline{44.72} \\
			Linear-Linear & 81.52 & 58.33 & \tbf{51.32} & \underline{\tbf{49.07}} \\
			MLP-Linear & 83.86 & \tbf{58.86} & 51.23 & \underline{48.74} \\
			\bottomrule
		\end{tabular}
	}
	\label{tab:linear}
\end{table}{}

\subsubsection{Architecture of Probe Networks}

\tbf{Linear vs. Non-linear Probe}: For a certain layer modified by CIFS, the probe network in CIFS serves as the surrogate classifier of the subsequent layers in the backbone model. Thus, the probe networks should be powerful enough to make correct predictions based on the features of this layer. For the CIFS in the last residual block, we use a linear layer network as the probe, while for the CIFS in the second last residual block, we compare the cases of using a linear layer versus using a two-layer MLP network. From Table \ref{tab:linear}, we observe that the MLP-Linear combination shows a similar performance compared to the combination of two linear layers against adversarial attacks, but enjoys a clear advantage on the natural data ($83.86$\% vs. $81.52$\%). This is because the features in the second last residual block are not as characteristic as those in the last block and cannot be linearly separated. The MLP can thus classify the features better than a pure linear layer.

%% file: Chapters/conclusion.tex
\chapter{Conclusion}
\label{chp:conclusion}
\section{Summary of Contributions}
The overarching theme of this thesis is building adversarially robust deep vision systems, of which three aspects were studied, i.e., evaluating adversarial robustness, enhancing robustness, and the connection between adversarial robustness and adaptivity to unseen data.

In Chapter \ref{chp:ijcai2022}, we systematically investigate the robustness of deep image denoising models. In particular, we first propose a novel PGD-based adversarial attack method, ObsAtk, to evaluate the robustness of deep image denoisers. ObsAtk is the first attack method for image denoising that considers the zero-mean assumption of natural noise. Then, we develop an adversarial training strategy, HAT, that can effectively robustify deep image denoisers. 
Finally, we reveal that adversarial robustness benefits the generalization to unseen types of noise, i.e., deep denoisers trained with HAT by using only Gaussian synthetic data can effectively remove various types of real-world noise that are unseen during training.

In Chapter \ref{chp:iclr2020} and \ref{chp:icml2021}, we explore new families of models that can used to construct more robust image classifiers in comparison to vanilla convolutional layers. Specifically, in Chapter \ref{chp:iclr2020}, we, for the first time, study the robustness property of neural ordinary differential equations (NODEs). We empirically demonstrate that NODE-based classifiers outperform CNN-based ones against input perturbations. We further boost the robustness of NODEs by introducing the time-invariant property and imposing a steady-state constraint on the flow of ODEs. The resultant variant, TisODE, enjoys superior robustness compared to the vanilla version and can work in conjunction with other robustification techniques. 

In Chapter \ref{chp:icml2021}, we improve the robustness of vanilla CNNs by introducing a novel channel-wise feature selection mechanism, CIFS. Inspired by the observations that adversarial training robustifies CNNs by aligning the channels' activations with their relevances to true labels, we propose CIFS to modify vanilla convolutional layers by adjusting channels' activations with multipliers generated based on their relevances. Extensive results verify the effectiveness of CIFS on robustification. 

\section{Recommendations for Future Research}
The works introduced in the previous chapters  naturally lead to various avenues for further research. We mention several possible extensions in this section.

\paragraph{Robustness of Image Restoration Algorithms}
Existing research papers mainly focus on the robustness of semantic-level tasks, such as classification and detection. The robustness of image restoration models, including denoising, deblurring, super-resolution, and compressive sensing, is still under-explored. Image restoration lies at the heart of intelligent vision systems. The performance of downstream image understanding tasks highly depends on the quality of the reconstruction from degraded images. Thus, it is essential to design robust image restoration models that can reliably reconstruct high-quality images against malicious degradation. 

At first, we need to develop effective and efficient attack methods to evaluate the robustness of image restoration algorithms. These are open problems to solve because such attack methods have to consider the constraint about the intrinsic properties of noise/perturbations in the natural world. For example, the noise in images is usually assumed to be zero-mean; blurring kernels, which smoothen images, usually contain all non-negative elements and satisfy certain types of symmetry. Then, we can use the developed attack methods to generate adversarial examples for adversarial training. When designing adversarial training strategies, one should consider the performance on authentic data, the robustness issue, as well as the generalization to unseen types of degradation sources.

\paragraph{Robustifying Deep Models without Adversarial Training}
Adversarial training (AT) is a generally applicable framework for training robust models. However, training a large, robust model with PGD-AT consumes a massive amount of computational resources, around ten or even twenty times of those required by normal training. That is because AT needs to perform multiple times of back-propagation to generate adversarial examples for each batch of data on the fly. Although some research works attempt to accelerate AT, the computational overhead is still very high. This situation motivates us to explore novel architectures that can build robust models without adversarial training. 

In Chapter \ref{chp:iclr2020}, we study the robustness property of NODEs. Although the proposed TisODE enjoys superior robustness compared to convolutional layers, the prediction performances on adversarial data are still suboptimal. TisODE introduces a steady-state constraint on the flow of ODEs, but this constraint is not sufficient for the stability of states. For future research, one may consider a stronger regularization of the flow, i.e., the Lyapunov stability condition, to encourage the models to learn stable equilibrium points in feature spaces. The challenges here are how to ensure the Jacobian of the dynamics function to be symmetric and negative-definite. In practice, the other challenge is how to perform backpropagation through the eigenvalue decomposition of the Jacobian if one proposes to regularize the largest eigenvalue in the loss function.